%% file: main.tex
\documentclass[runningheads]{llncs}
\usepackage{eccv}

\usepackage{eccvabbrv}
\usepackage{comment}
\usepackage{graphicx}
\usepackage{booktabs}
\usepackage{subcaption}
\usepackage{xcolor}         % colors
\usepackage{colortbl}  %彩色表格需要加载的宏包
\usepackage{array}   %对表列和表格线的设置需要用到array宏包
\usepackage[accsupp]{axessibility}  % Improves PDF readability for those with disabilities.
\usepackage{algpseudocode} % 算法伪代码
\usepackage{amsmath} % 数学公式
\usepackage{amssymb} % 数学符号
\usepackage{multirow}
\usepackage{adjustbox}

\usepackage{hyperref}

\usepackage{url}            % simple URL typesetting
\usepackage{booktabs}       % professional-quality tables
\usepackage{amsfonts}       % blackboard math symbols
\usepackage{nicefrac}       % compact symbols for 1/2, etc.
\usepackage{microtype}      % microtypography
\usepackage{wrapfig}
\usepackage{bbm}
\usepackage{caption}
\usepackage{subcaption}
\usepackage[ruled]{algorithm2e}

\usepackage{float}
\usepackage{enumitem}
\usepackage[utf8]{inputenc}
\usepackage{amsmath}
\usepackage{algorithm2e}

\usepackage{hyperref}

\usepackage{orcidlink}

\usepackage[utf8]{inputenc}
\usepackage{amsmath}
\usepackage{algorithm2e}
\SetKwComment{Comment}{//}{}

\begin{document}

\title{Dual-level Adaptive Self-Labeling for Novel Class Discovery in Point Cloud Segmentation} 

\titlerunning{Dual-level Adaptive Self-Labeling for NCD in Point Cloud Segmentation}

\author{Ruijie Xu\inst{1,*} \and
Chuyu Zhang\inst{1,}\thanks{Both authors contributed equally. Code is available at \href{https://github.com/RikkiXu/NCD_PC}{Github}.} \and
Hui Ren\inst{1} \and Xuming He\inst{1,3}}

\authorrunning{Ruijie Xu, Chuyu Zhang et al.}

\institute{ShanghaiTech University, Shanghai, China \and
% Lingang Laboratory, Shanghai, China \and
Shanghai Engineering Research Center of Intelligent Vision and Imaging, Shanghai, China\\
\email{\{xurj2022,zhangchy2,renhui,hexm\}@shanghaitech.edu.cn}}

\maketitle
\input{section/abstract.tex}
\input{section/intro.tex}
\input{section/related_work.tex}
\input{section/method.tex}
\input{section/exp.tex}

\input{section/conclusion.tex}
%\input{section/suppl.tex}


\bibliographystyle{splncs04}
\bibliography{main}

\end{document}

\title{Dual-level Adaptive Self-Labeling for Novel Class Discovery in Point Cloud Segmentation} 

\titlerunning{Dual-level Adaptive Self-Labeling for NCD in Point Cloud Segmentation}

\author{Ruijie Xu\inst{1,*} \and
Chuyu Zhang\inst{1,}\thanks{Both authors contributed equally. Code is available at \href{https://github.com/RikkiXu/NCD_PC}{Github}.} \and
Hui Ren\inst{1} \and Xuming He\inst{1,3}}

\authorrunning{Ruijie Xu, Chuyu Zhang et al.}

\institute{ShanghaiTech University, Shanghai, China \and
% Lingang Laboratory, Shanghai, China \and
Shanghai Engineering Research Center of Intelligent Vision and Imaging, Shanghai, China\\
\email{\{xurj2022,zhangchy2,renhui,hexm\}@shanghaitech.edu.cn}}

\maketitle

\input{section/suppl.tex}


\newpage
\bibliographystyle{splncs04}
\bibliography{main}

%% file: section/abstract.tex
\begin{abstract}
We tackle the novel class discovery in point cloud segmentation, which discovers novel classes based on the semantic knowledge of seen classes.
Existing work proposes an online point-wise clustering method with a simplified equal class-size 
constraint on the novel classes to avoid degenerate solutions. However, the inherent imbalanced distribution of novel classes 
in point clouds typically violates the equal class-size constraint. Moreover, point-wise clustering ignores the rich spatial context information of objects, which results in less expressive representation for semantic segmentation.
To address the above challenges, we propose a novel self-labeling strategy that adaptively generates high-quality pseudo-labels for imbalanced classes during model training. In addition, we develop a dual-level representation that incorporates regional consistency into the point-level classifier learning, reducing noise in generated segmentation. Finally, we conduct extensive experiments on two widely used datasets, SemanticKITTI and SemanticPOSS, and the results show our method outperforms the state of the art by a large margin.
 \keywords{Novel class discovery \and Point clouds semantic segmentation \and Long-tailed learning}
\end{abstract}

% The mutual information is estimated by sampling multiple labeled and unlabeled data pairs. It is a tractable estimation and can be trained end-to-end. 

%% file: section/intro.tex
\section{Introduction}

Point cloud segmentation is a core problem in 3D perception~\cite{landrieu2018large} and potentially useful for a wide range of  
applications, such as autonomous driving and intelligent robotics~\cite{li2020deep,roriz2021automotive}. Recently, there has been tremendous progress in semantic segmentation of point clouds due to the utilization of deep learning techniques~\cite{hu2020randla,lai2022stratified}. However, current segmentation methods primarily focus on a closed-world setting where all the semantic classes are known beforehand. As such it has difficulty in coping with open-world scenarios where both known and novel classes coexist, which are commonly seen in real-world applications.

For open-world perception, a desirable capability is to automatically acquire new concepts based on existing knowledge~\cite{han2019learning}. While there has been much effort into addressing the problem of novel class discovery for 2D or RGBD images~\cite{nakajima2019incremental, han2021autonovel,fini2021unified, zhao2022novel}, few works have explored the corresponding task for 3D point clouds. Only recently, Riz et al.~\cite{riz2023novel}
propose an online point-wise clustering method for discovering novel classes in 3D point cloud segmentation. To avoid degenerate solutions, their method relies on an equal class-size constraint on the novel classes. Despite its promising results, such a simplified assumption faces two key challenges: First, the distribution of novel classes in point clouds is inherently imbalanced due to the different physical sizes of objects and the density of points. Imposing the equal-size constraint can be restrictive, causing the splitting of large classes or the merging of smaller ones. In addition, point-wise clustering tends to ignore the rich spatial context information of objects, which leads to less expressive representation for semantic segmentation.

To tackle the above challenges, we propose a dual-level adaptive self-labeling framework for novel class discovery in point cloud segmentation. The key idea of our approach is two-fold: 1) We design a novel self-labeling strategy that adaptively generates high-quality imbalanced pseudo-labels for model training, which facilitates clustering novel classes of varying sizes; 2) To incorporate semantic context, we develop a dual-level representation of 3D points by grouping points into regions and jointly learns the representations of novel classes at both the point and region levels. Such a dual-level representation imposes additional constraints on grouping the points likely belonging to the same category. This helps in mitigating the noise in the generated segmentation.

Specifically, our framework employs an encoder to extract point features for the input point cloud and average pooling to compute representations of pre-computed regions. Both types of features are fed into a prototype-based classifier to generate predictions across both known and novel categories for each point and region. To learn the feature encoder and class prototypes, we introduce a self-labeling-based learning procedure that iterates between pseudo-label generation for the novel classes and the full model training with cross-entropy losses on points and regions. Here the key step is to generate imbalanced pseudo labels, which is formulated as a semi-relaxed Optimal Transport (OT) problem with adaptive regularization on class distribution. Along with the training, we employ a data-dependent annealing scheme to adjust the regularization strength. Such a design prevents discovering degenerate solutions and meanwhile enhances the model flexibility in learning the imbalanced data distributions.

To demonstrate the effectiveness of our approach, we conduct extensive experiments on two widely-used datasets: SemanticKITTI~\cite{behley2019semantickitti} and SemanticPOSS~\cite{pan2020semanticposs}. 
The experimental results show that our method outperforms the state-of-the-art approaches by a large margin. Additionally, we conduct comprehensive ablation studies to evaluate the significance of the different components of our method.
The contributions of our method are summarized as follows:

\begin{enumerate}%[leftmargin=8mm]
%xx,能干嘛，结果%
\item We propose a novel adaptive self-labeling framework for novel class discovery in point cloud segmentation, better modeling imbalanced novel classes.

\item We develop a dual-level representation for learning novel classes in point cloud data, which incorporates semantic context via augmenting the point prediction with regional consistency.

\item Our method achieves significant performance improvement on the SemanticPOSS and SemanticKITTI datasets across nearly all the experiment settings.
\end{enumerate}

%% file: section/related_work.tex
\section{Related Work}

\paragraph*{Point cloud semantic segmentation.} 
%Various methods have been proposed for point cloud semantic segmentation. According to input format, those method can be categorized as range view~\cite{cortinhal2020salsanext,milioto2019rangenet++}, bird's eye view~\cite{zhang2020polarnet,li2022bevformer}, and voxel~\cite{choy20194d,zhu2021cylindrical}. Despite their great progress, they usually focus on fully supervised scenarios, which rely on massive annotation data. To reduce annotation cost, several semi-supervised point cloud semantic segmentation methods are proposed~\cite{kong2023lasermix,jiang2021guided}, which assume unlabeled data are known classes. Unlike them, we assume unlabeled data are novel classes and aim to discover novel classes by the knowledge of known classes.
Point cloud semantic segmentation has attracted much attention in recent years~\cite{choy20194d,zhang2020polarnet, zhu2021cylindrical,li2022bevformer}. 
%These methods can be categorized based on their input format, namely range view~\cite{cortinhal2020salsanext, milioto2019rangenet++}, bird's eye view~\cite{zhang2020polarnet, li2022bevformer}, and voxel-based~\cite{choy20194d, zhu2021cylindrical} approaches.
%While these methods have made significant advancements, their primary focus is on fully supervised scenarios that heavily rely on annotations for each class. 
% 相比之下，ncd对于fully来说具有的优势，我们的方法是使用一个空间的先验来xx，同时在unsupervised的方法中也有使用了这样的先验
While previous methods have made significant progress, their primary focus is on closed-world scenarios that heavily rely on annotations for each class and cannot address open-world challenges. In contrast, we aim to develop a model to discover novel classes in 3D open-world scenarios.
% novel class discovery leverages pre-labeled known classes to transfer knowledge to novel classes, enabling a better understanding of open-world scenarios.
In the context of point cloud representation learning, incorporating spatial context is pivotal for enhancing representation learning. Several works~\cite{Long:CVPR23,zhang2023growsp} introduce a hierarchical representation learning strategy that leverages regions as intermediaries to connect points and semantic clusters. 
% Specifically, they encourage points within the same region to converge and regions within the same cluster to coherent. 
Unlike them, we develop a dual-level learning strategy that concurrently learns to map points and regions to semantic classes. Thanks to the learning of region-level representation, our method is less sensitive to the local noises in point clouds. Moreover, we cluster regions into semantic classes by an imbalance-aware self-labeling algorithm instead of simple K-Means.

\paragraph*{Novel class discovery.} The majority of research on Novel Class Discovery (NCD) has focused on learning novel visual concepts in the 2D image domain via designing a variety of unsupervised losses on novel class data or regularization strategies~\cite{han2019learning,zhao2021novel,fini2021unified,yang2022divide,zhang2023promptcal,gu2023class}. 
% The initial formulation of this problem was carried out in the context of 2D image recognition by \cite{han2019learning}. Subsequent works have explored different approaches to tackle NCD, either employing pair-wise similarity loss~\cite{han2021autonovel, zhong2021neighborhood, zhong2021openmix} or self-labeling loss~\cite{fini2021unified, yang2022divide} to cluster novel classes.
% Furthermore, \cite{vaze2022generalized} extended NCD to a more practical setting, where unlabeled data comprises both known and novel classes. 
Among them, EUMS~\cite{zhao2022novel} addresses novel class discovery in semantic segmentation, employing a saliency model for clustering novel classes, along with entropy ranking and dynamic reassignment for clean pseudo labels. More relevantly, Zhang et al.~\cite{zhang2023novel} consider the NCD task in long-tailed classification scenarios, and develop a bi-level optimization strategy for model learning. It adopts a fixed regularization to prevent degeneracy, imposing strong restrictions on learned representations, and a complex dual-loop iterative optimization procedure. In contrast, we propose an adaptive regularization strategy, which is critical for the success of our self-labeling algorithm. Moreover, our formulation leads to a convex pseudo-label generation problem, efficiently solvable by a fast scaling algorithm~\cite{cuturi2013sinkhorn,chizat2018scaling} (see Appendix A
%\ref{appendix:srot} 
for detailed comparisons).
Perhaps most closely related to our work is~\cite{riz2023novel}, which 
%More recently, Riz et al.~\cite{riz2023novel} 
explored the NCD problem for the task of point cloud semantic segmentation. Assuming a uniform distribution of novel classes, they develop an optimal-transport-based self-labeling algorithm to cluster novel classes. However, the method neglects intrinsic class imbalance and spatial context in point cloud data, often leading to sub-optimal clustering results.

%formulate the pseudo labelling generation process in the long-tailed scenario as a bi-level optimization problem, which alternately estimates cluster distributions and generates pseudo labels by solving an optimal transport problem. The difference between our formulation with \cite{zhang2023novel} are that: 1) we propose a novel adaptive regularization strategy, which is vital important for the success of our self-labelling algorithm, instead of a constant regularization; 2) thanks to the optimization technique developed by \cite{cuturi2013sinkhorn,chizat2018scaling}, we optimize our formulation directly by light-speed scaling algorithm instead of bi-level optimization technique, which is time-consuming and need to additional hyperparameters for inner loop optimization (refer to Appendix \ref{appendix:srot} for detail comparison).

% Despite achieving state-of-the-art performance compared to existing methods, they overlook the intrinsic imbalanced nature and density prior property inherent in point cloud data, resulting in suboptimal clustering outcomes.

\paragraph*{Optimal transport for pseudo labeling.}  %aims to find the optimal couple matrix under the constraint of marginal distributions. 
Unlike naive pseudo labeling~\cite{lee2013pseudo}, Optimal Transport (OT)~\cite{villani2009optimal,phatak2023computing}-based methods allow us to incorporate prior class distribution into pseudo-labels generation. Therefore, it has been used as a pseudo-labels generation strategy for a wide range of machine learning tasks, including semi-supervised learning~\cite{lai2022sar,tai2021sinkhorn,taherkhani2020transporting}, clustering~\cite{asano2020self,caron2020unsupervised,zhang2023p}, and domain adaptation~\cite{flamary2016optimal,chang2022unified,Liu_2023_CVPR}. However, most of these works assume the prior class distribution is either known or simply the uniform distribution, which is restrictive for NCD. By contrast, we consider a more practical scenario, where the novel class distribution is unknown and imbalanced, and design a semi-relaxed OT formulation with a novel adaptive regularization. 

%% file: section/method.tex
\section{Method}

In this section, we first introduce the problem setup of novel class discovery for point cloud segmentation and an overview of our method in Sec.\ref{sec:ps}. We then describe our network architecture, including dual-level representation of point clouds in Sec.\ref{sec:rep}. Subsequently, we present in detail our adaptive self-labeling framework for model learning that discovers the novel classes in Sec.\ref{sec:selflabeing}. Finally, we introduce our strategy to estimating the number of novel classes in Sec.\ref{sec:estmatek_method}.

\subsection{Problem Setup and Overview}\label{sec:ps}

%For the task of point cloud segmentation, the novel class discovery problem aims to learn to classify 3D 
%points of a scene into known and novel semantic classes from a dataset consisting of annotated points 
%from the known classes and unlabeled points from novel ones. Formally, we consider a training set of 
%3D scenes $\mathcal{D}^{tr}=  \{(\mathbf{D}^s,\mathbf{X}^u)_i\}_{i=1}^{|\mathcal{D}^{tr}|}$, where each 
%scene comprises two parts: 1) an annotated part of the scene $\mathbf{D}^s=\{{(x^s_n, y^s_n)}\}^N_{n=1}$, 
%which belongs to the known classes $C^s$ and consists of original point clouds along with the corresponding 
%ground truth labels for each point; 2) a unknown part of the scene $\mathbf{X}^u=\{{(x^u_m)}\}^M_{m=1}$, 
%which belongs to the novel classes  $C^u$ and does not contain any label information. These two sets $C^s$ 
%and $C^u$ are mutually exclusive, i.e., $C^s \cap C^u = \emptyset$.  
%Our goal is to learn a point cloud segmentation network that can accurately segment new scenes in a test set, 
%denoted as $\mathcal{D}^{te}=\{\mathbf{X}^{te}_i\}_{i=1}^{|\mathcal{D}^{te}|}$, each of which includes both
%known and novel classes.
For the task of point cloud segmentation, the novel class discovery problem aims to learn to classify 3D points of a scene into known and novel semantic classes from a dataset consisting of annotated points from the known classes and unlabeled points from novel ones. 
% Formally, we consider a training set of 3D scenes $\mathcal{D}^{tr}=  \{(\mathbf{D}^s,\mathbf{X}^u)_i\}_{i=1}^{|\mathcal{D}^{tr}|}$, where each scene comprises two parts: 1) an annotated part of the scene $\mathbf{D}^s=\{{(x^s_n, y^s_n)}\}^N_{n=1}$, which belongs to the known classes $C^s$ and consists of original point clouds along with the corresponding ground truth labels for each point; 2) a unknown part of the scene $\mathbf{X}^u=\{{(x^u_m)}\}^M_{m=1}$, which belongs to the novel classes  $C^u$ and does not contain any label information. These two sets $C^s$ and $C^u$ are mutually exclusive, i.e., $C^s \cap C^u = \emptyset$.  
% Our goal is to learn a point cloud segmentation network that can accurately segment new scenes in a test set, denoted as $\mathcal{D}^{te}=\{\mathbf{X}^{te}_i\}_{i=1}^{|\mathcal{D}^{te}|}$, each of which includes both known and novel classes.

Formally, we consider a training set of 3D scenes, where each scene comprises two parts: 1) an annotated part of the scene $\{(x^s_n, y^s_n)\}^N_{n=1}$, which belongs to the known classes $C^s$ and consists of original point clouds along with the corresponding labels for each point; 2) an unknown part of the scene $\{(x^u_m)\}^M_{m=1}$, which belongs to the novel classes $C^u$ and does not contain any label information. These two sets $C^s$ and $C^u$ are mutually exclusive, i.e., $C^s \cap C^u = \emptyset$.  
Our goal is to learn a point cloud segmentation network that can accurately segment new scenes in a test set, each of which includes both known and novel classes.

\begin{figure}
    \centering
    \includegraphics[scale=0.3]{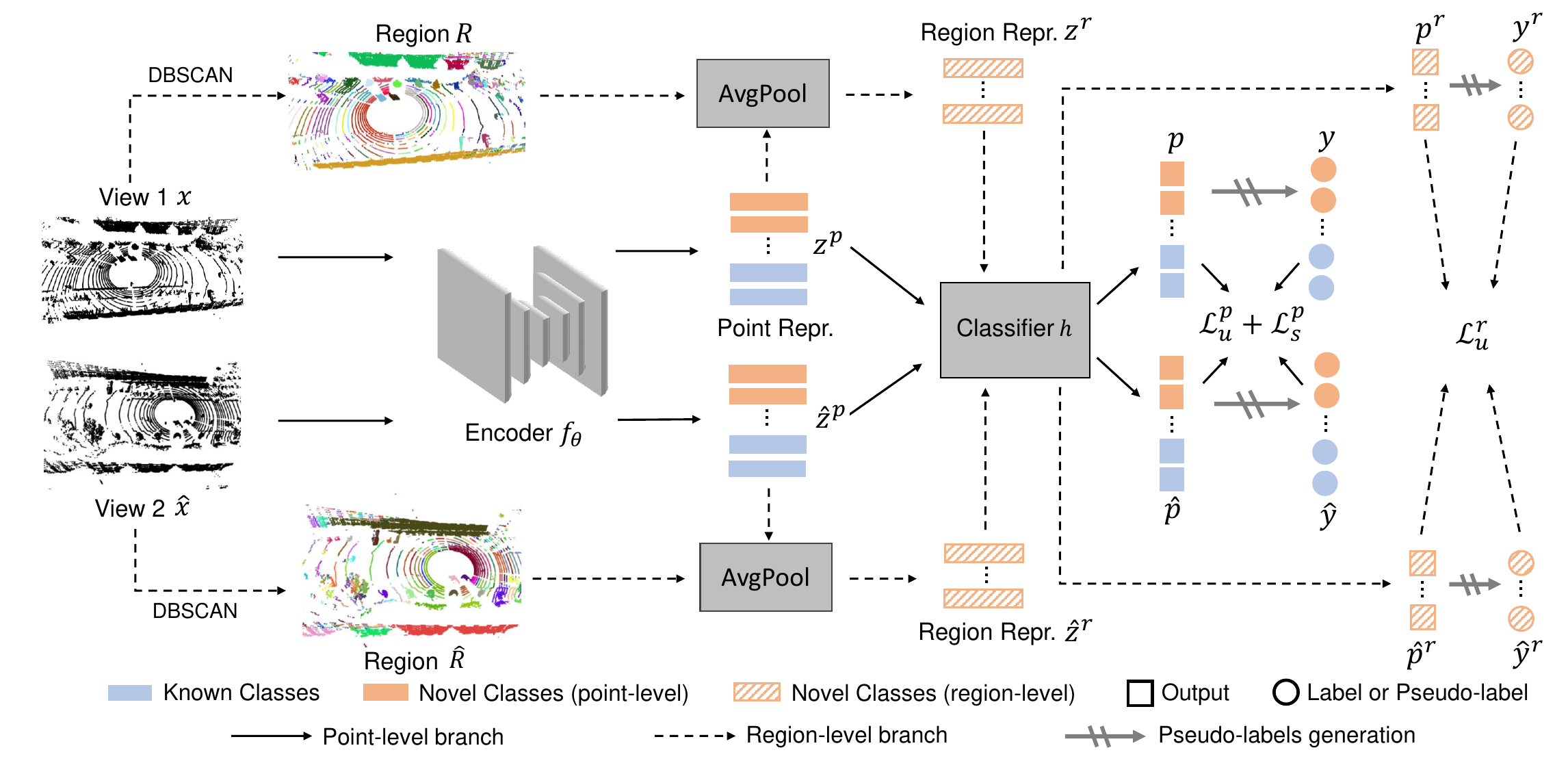}
    \caption{Our method starts with two views of the same point cloud ($x$ and $\hat{x}$) and clusters the points into corresponding regions. 
    Then, we extract individual point features via a forward pass and calculate regional representations 
    by averaging the point features within each region. Next, we make predictions $p$ by the classifier $h$, 
    and generate pseudo labels $y$ for unlabeled points and regions using our novel adaptive self-labeling algorithm (generating pseudo-labels does not involve gradients). 
    %Note that \textcolor{red}{$x$ and $\hat{x}$  correspond to the two different views.} 
    Lastly, we exchange the pseudo labels between the two views 
    and update the model accordingly.  
    %Specifically, for point-level, we compute $\mathcal{L}^p_u(y_2,p_1 )$ and $\mathcal{L}^p_u(y_1,p_2)$.  For region-level, we compute $\mathcal{L}^r_u(y_2^r,p_1^r)$ and $\mathcal{L}^r_u(y_1^r,p_2^r)$.
}
    
    \label{fig:enter-label}
\end{figure}

To tackle the challenge of discovering novel classes in point clouds, we introduce a dual-level adaptive self-labeling framework to learn a segmentation network for both known and novel classes. The key idea of our method includes two aspects: 1) utilizing the spatial smooth prior of point clouds to generate regions and developing a dual-level representation that incorporates regional consistency into the point-level classifier learning; 2) generating imbalance pseudo-labels with a novel adaptive regularization. An overview of our framework is depicted in Fig.\ref{fig:enter-label}. 
%An overview of our framework is illustrated in Fig.\ref{fig:enter-label}. 

% build a classification model on the joint label space from $\mathcal{D}^l$ and $\mathcal{D}^u$. 

%\subsection{Method Overview}\label{method:overview}
%

\subsection{Model Architecture}\label{sec:rep}

We adopt a generic segmentation model architecture consisting of a feature encoder for the input point cloud and a classifier head to generate the point-wise class label prediction. Note that to capture both known and novel classes, our feature encoder is shared by all the classes $C^s \cup C^u$ and the output space of our classifier head also includes known and novel classes. Below we first introduce our feature representation and encoder, followed by the classifier head.  

%As shown in Fig.\ref{fig:enter-label}, our model is a dual-branch network, consisting of a MinkowskiUNet \cite{choy20194d} for feature extraction ($f_\theta$) and prototypes for classifying known and novel classes, denoted as $h = [h^s, h^u] \in \mathbb R^{D\times(|C^s|+|C^u|)}$.

\paragraph{Dual-level Representation.}
% Our method utilizes an adaptive and imbalanced-aware self-labeling algorithm, enabling the clustering of novel classes at both the point-level and region-level. 
Instead of treating each point independently, we exploit the spatial smoothness prior to 3D objects in our representation learning. To this end, we adopt a dual-level representation of point clouds that describes the input scene at different granularity.  
Specifically, given an input point cloud $\mathbf{X}$, we first use a backbone network $f_\theta$ to compute a point-wise feature  $\mathbf{Z}^p=\{\mathbf{z}_i^p\}$, where $\mathbf{z}_i^p\in \mathbb{R}^{D \times 1}$. In this work, we employ MinkowskiUNet \cite{choy20194d} for the backbone. In addition, we cluster points into regions based on their coordinates and then compute regional features by average pooling of point features. Concretely, during training, we first utilize DBSCAN~\cite{ester1996density} to generate $K_i$ regions, $\mathcal{R}=\{r_k\}^{K_i}_{k=1}$, for unlabeled point in sample $i$, and calculate the regional features as follows,
\begin{equation}
	\{r_k\}^{K_i}_{k=1} \leftarrow \text{DBSCAN}(\{x^u_i\}^M_{i=1}), \quad \mathbf{z}^r_k =\text{AvgPool}\{{\mathbf{z}_i^p}|\mathbf{z}_i^p=f_\theta(x_i^u), \; x_i^u\in r_k\},
\end{equation}
where $\mathbf{z}^r_k$ is the feature of region $r_k$. 
Such a dual-level representation allows us to enforce regional consistency in representation learning.

\paragraph{Prototype-based Classifier.} 
We adopt a prototype-based classifier design for generating the point-wise predictions. Specifically, we 
introduce a set of prototypes for known and novel classes, denoted as $h = [h^s, h^u] \in \mathbb R^{D\times(|C^s|+|C^u|)}$, and $D$ denotes the dimension of the last-layer feature. For each point or region, we compute the cosine similarity between its feature and the prototypes, followed by Softmax to predict the class probabilities. Here we use the same set of prototypes for the points and regions, which enforces a consistency constraint within each region and results in a more compact representation for each class.  

%Then we employ the prototypes, denoted as $h^u_r$, to classify the regions into distinct semantic clusters. Instead of learning different prototypes for point-level and region-level representations, we share prototypes for both (i.e., $h^u_r=h^u$), which pull the point-level and region-level feature move forward in the same direction, and can achieve a more compact representation. What's more, one can adopt a more flexible approach, such as using the Mean Squared Error loss between $h^u_r$ and $h_u$, to impose constraints on $h^u_r$ and $h^u$.
%In this paper, we adopt a simple design and analyze its effectiveness in ablation studies. 
% To generate pseudo label for point and region, we propose a novel adaptive self-labeling algorithm. Then, we swap the pseudo label of different views to learn transformation invariance cluster. 

\subsection{Adaptive Self-labeling Framework} \label{sec:selflabeing}
To handle class-imbalanced data, we propose an adaptive self-labeling framework that dynamically generates imbalanced pseudo-labels. 
To this end, we adopt the following loss function for the known and novel classes, 
\begin{equation}\label{eq:totalloss}
	\mathcal L = \mathcal L_s+ \alpha\mathcal L_u^p+\beta\mathcal L_u^r,
\end{equation}
where $\mathcal L_s$ is the cross-entropy loss for known classes, $\mathcal{L}_u^p$ is point-level loss and $\mathcal{L}_u^r$ is region-level loss for novel classes. $\alpha$ and $\beta$ are weight parameters. 
For the novel classes, we first generate pseudo-labels for points and regions by solving a semi-relaxed Optimal Transport problem and then adopt the cross-entropy loss with the generated labels. 
The pseudo-code of our algorithm is shown in Appendix 
B
%\ref{append:pc} 
and below we will focus on our novel pseudo-label generation process.

\begin{algorithm}[t!]
\caption{Semi-relaxed Optimal Transport Algorithm}\label{alg:srot}
\SetAlgoLined
\footnotesize
\DontPrintSemicolon
\SetKwFunction{SelfLabeling}{Self-Labeling}
\SetKwProg{Fn}{Function}{}{end}
\Fn{\SelfLabeling{$-\log \mathbf{P}$, $\gamma$, $\epsilon$}}{
    $\mathbf{K}= \exp(\log \mathbf{P}/\epsilon), \quad f \leftarrow \frac{\gamma}{\gamma + \epsilon}$
    
    $\mathbf{\mu}, \mathbf{\nu} \leftarrow  \frac{1}{M}\mathbf{1}_M, \frac{1}{|C^u|}\mathbf{1}_{|C^u|}$ \Comment{Marginal distribution}

    $\mathbf{b}_0  \leftarrow \mathbf{1}_{|C^u|}$ \Comment{Initialize $\mathbf{b}$}
    
    % \Comment{Inner iteration for $\mathbf{\hat{Y}}$ and \mathbf{\nu}_l}
    \While{$\|\mathbf{b}_{t+1}-\mathbf{b}_{t}\|<1e-4$}{
    
    $\mathbf{a} \leftarrow \frac{\mathbf{\mu}}{\mathbf{K} \mathbf{b}_t}$
    
    $\mathbf{b}_{t+1} \leftarrow \left(\frac{\mathbf{\nu}}{\mathbf{K}^\top  \mathbf{a}}\right)^f$
    }
    
    $\mathbf{Q} \leftarrow M\text{diag}(\mathbf{a}) \mathbf{K} \text{diag}(\mathbf{b})$
    
    \KwRet $\mathbf{Q}$\;
}
\end{algorithm}

\paragraph{Imbalanced Pseudo Label Generation.} The pseudo-labels generation for balanced classes can be formulated as an optimal transport problem as follows~\cite{asano2020self,zhang2023novel}:
\begin{align}\label{eq:balance_sl}
	\min_{\mathbf Q}\frac{1}{M}\langle{\mathbf{Q}},-\log\mathbf P^u\rangle_F, \quad\text{s.t. }{\mathbf{Q}}\mathbf{1}_{|C^u|}=\mathbf{1}_M, \mathbf{Q}^{\top} \mathbf1_M= \frac{M}{|C^u|}\mathbf{1}_{|C^u|},
\end{align}
where $\mathbf{Q} \in \mathbb{R}^{M\times |C^u|}$ are the pseudo labels of unlabeled data,  $<,>_F$ is Frobenius inner product and $\mathbf{P}^u$ are the output probabilities of the model. For imbalanced point cloud data, we relax the second constraint on the class sizes in \cref{eq:balance_sl}, which leads to a parameterized semi-relaxed optimal transport problem as below:
\begin{align}\label{eq:imbalance_sl}
	\min_{{\mathbf{Q}}}\mathcal{F}_u({\mathbf{Q}},\gamma&)=\frac{1}{M}\langle{\mathbf{Q}},-\log\mathbf P^u\rangle_F+\gamma KL(\frac{1}{M}{\mathbf{Q}}^{\top} \mathbf1_M,\frac{1}{|C^u|}\mathbf{1}_{|C^u|}) \nonumber\\
	&\text{s.t. }{\mathbf{Q}} \in \{{\mathbf{Q}} \in \mathbb R^{M\times |C^u|}|{{\mathbf{Q}}}\mathbf{1}_{|C^u|}=\mathbf{1}_M\}, 
\end{align}
where $\gamma$ is a weight coefficient for balancing the constraint on cluster size distribution in the second term.
We further add an entropy term $-\epsilon \mathcal{H} (\frac{1}{M}\mathbf{Q})$ to \cref{eq:imbalance_sl} and for any given $\gamma$, this entropic semi-relaxed OT problem can be efficiently solved by fast scaling algorithms~\cite{cuturi2013sinkhorn,chizat2018scaling}. \cref{alg:srot} outlines the optimization process, and further details are provided in Appendix A. %\ref{appendix:srot}.

In this work, we propose a novel adaptive regularization strategy that adjusts the weight $\gamma$ according to the progress of model learning, significantly improving pseudo-label quality. Details of our strategy will be illustrated subsequently.

\paragraph{Adaptive Regularization Strategy.}\label{sec:adaptive_reg}
The objective \cref{eq:imbalance_sl} aims to strike a balance between the distribution represented by model prediction $\mathbf{P}^u$ and the uniform prior distribution. A large $\gamma$ tends to prevent the model from learning a degenerate solution, e.g. assigning all the samples into a single novel class, but it also restricts the model's capacity to learn the imbalanced data. %Conversely, a lower prior constraint enables the model more flexibility to learn data distribution but tends to learn a degenerate solution when the model's representation is worse. 
One of our key insights is that the imbalanced NCD learning requires an adaptive strategy for setting the value of $\gamma$ during the training. 
Intuitively, in the early training stage where the model performance is relatively poor, a larger constraint on ${\mathbf{Q}}^{\top} \mathbf1_M$ is needed to prevent degenerate solutions. As the training progresses, the model gradually learns meaningful clusters for novel classes, and the constraint should be relaxed to increase the flexibility of pseudo-label generation. %In summary, the annealed $\gamma$ should be adaptive with model learning. 

To achieve that, we develop an annealing-like strategy for adjusting $\gamma$, inspired by the ReduceLROnPlateau method that reduces the learning rate when the loss does not decrease. 
Here we employ the KL term in \cref{eq:imbalance_sl} as a guide for decreasing $\gamma$, as the value of the KL term reflects the relationship between the distribution of pseudo labels and the uniform distribution.
Specifically, our formulation for the adaptive regularization factor is as follows:
\begin{equation}\label{eq:gamma}
 \gamma_{t+1} = \lambda\gamma_{t}, \text{\ if $KL(\frac{1}{M}{\mathbf{Q}}^{\top} \mathbf1_M,\frac{1}{|C^u|}\mathbf{1}_{|C^u|})$ $\leq \rho$ consecutively for $T$ iter.}
\end{equation}
where $\rho, \lambda$, $T$ and $\gamma_0$ are hyperparameters. Compared to typical step decay and cosine decay strategies, our adaptive strategy is aware of the model learning process and allows for more flexible control of $\gamma$ based on the characteristics of the input itself.

% TODO: This section should be further polished. illustrate the reasoning of such an indicator. 
% 这里的实验还是要补一下。说明我们的strategy比step decay，以及cosine annearling好。前者主要是调节decay的大小，后者主要是最终gamma对大小。

\paragraph{Hyperparameter Search.}
To search the values of our hyperparameters, we design an indicator score that can be computed on the training dataset. Specifically, our indicator regularizes the total loss in \cref{eq:totalloss} with a KL term that measures the distance between the distribution of novel classes and the uniform distribution. Formally, the indicator is defined as follows:
\begin{equation}
 \mathcal{I} = \mathcal{L} + \gamma KL(\frac{1}{M}{\mathbf{Q}}^{\top} \mathbf1_M,\frac{1}{|C^u|}\mathbf{1}_{|C^u|}),
\end{equation}
% 如果indicator中没有KL loss，模型选择可能会存在trivial solution, where all the novel classes are assigned to a single cluster, and novel classes will be zero.
where $\gamma$ is obtained by \cref{eq:gamma}. Empirically, this indicator score provides a balanced evaluation of the model's performance in the known and novel classes. 

\subsection{Estimate the number of novel classes}\label{sec:estmatek_method}

To deal with realistic scenarios, where the number of novel classes ($C^u$) is unknown, we extend the classic estimation method~\cite{vaze2022generalized} in NCD to point clouds semantic segmentation for estimating $C^u$. Specifically, we extract representation from a known-class pre-trained model for training data, define a range of possible total class counts ( $|C^s| \textless |C_{all}|\textless$ max classes), and apply Kmeans to cluster the labeled and unlabeled point clouds across different $|C_{all}|$. Then, we evaluate the clustering performance of known classes under different $|C_{all}|$, and select $|C_{all}|$ with the highest clustering performance as the estimated $|C_{all}|$.

%% file: section/exp.tex
\section{Experiments}
\subsection{Experimental setup}
\paragraph*{Dataset.} We perform evaluation on the widely-used SemanticKITTI~\cite{geiger2012we,behley2021towards,behley2019semantickitti} and SemanticPOSS~\cite{pan2020semanticposs} datasets. The SemanticKITTI dataset consists of 19 semantic classes, while the SemanticPOSS dataset contains 13 semantic classes. Both datasets have intrinsic class imbalances. For a fair comparison with existing works~\cite{riz2023novel}, we divide the dataset into 4 splits and select one split as novel classes, while treating the remaining splits as the known classes. Additionally, to assess the effectiveness of our method under more challenging conditions, we further split the SemanticPOSS dataset into two parts, selecting one part as novel classes. The dataset details are provided in Appendix C. %\ref{appendix:dataset}. 

% We evaluate the performance of our method on typical CIFAR10, CIFAR100~\cite{krizhevsky2009learning}, ImageNet~\cite{deng2009imagenet}, OxfordIIIT-Pet~\cite{parkhi2012cats} and FGVC-Aircraft~\cite{maji2013fine}. Following ~\cite{cao2021open}, we first divide classes into 50\% seen and 50\% unseen, then select 50\% of seen classes as labeled, the rest as an unlabeled set, i.e., unlabeled data contains all unseen class data and half of seen class data. The details of the dataset split are shown in Appendix \cref{exp:dataset}.

% % What's more, we also refer to \cite{fini2021unified} to divide CIFAR100 into 50 seen and 50 unseen class, which is a more challenging setting. 

% \paragraph*{Baseline.} We compare our method with EUMS\cite{zhao2022novel} and NOPS\cite{riz2023novel}. 

%EUMS, extended to 3D by\cite{riz2023novel}, uses labeled base data and a saliency model for clustering novel classes.
%with entropy ranking and dynamic reassignment for clean pseudo labels.

\paragraph*{Evaluation Metric.} Following the official guidelines in SemanticKITTI and SemanticPOSS, we conduct evaluations on sequences 08 and 03, respectively. These sequences contain both known and novel classes. For the known classes, we report the IoU for each class. Regarding the novel classes, we employ the Hungarian algorithm to initially match cluster labels with their corresponding ground truth labels. Subsequently, we present the IoU values for each of these novel classes. Additionally, we calculate the mean of columns across all known and novel classes.
% for a comprehensive evaluation.

% The formulations of ClusterAcc is as follows:
% % Similar to previous work~\cite{cao2021open}, we evaluate our method on the unlabeled training set, which contains both seen and unseen classes. First, we adopt the Accuracy and ClusterAcc to measure the performance of seen and unseen classes separately. However, we can not distinguish unlabeled data as seen or unseen in advance, and such a separate metric ignores the confusion between seen and unseen classes. Therefore, we utilize all the unlabeled data to calculate ClusterAcc. The formulation of ClusterAcc is: 
% \begin{equation}
% ClusterAcc = \mathop{max}_{perm \in P} \frac{1}{N} \sum_{i=1}^{N} \mathbbm{1} \{y_i= perm (\hat{y}_i)\}
% \end{equation}
% where $y_i, \hat{y}_i$ represent the ground-truth and predicted labels for each point, respectively, and $P$ represents the set of all permutations and combinations. We use the Hungarian algorithm to optimize permutations. And we also report mean IOU over all known and novel classes.
% % In GNCD, we adopt Accuracy to measure the performance of labeled data. When calculating the overall performance, we still use ClusterAcc.
% \vspace{-0.7em}

\paragraph*{Implementation Details.} 
We follow~\cite{riz2023novel} to adopt the MinkowskiUNet-34C~\cite{choy20194d} network as our backbone. For the parameters in DBSCAN, we set the min\_samples to a reasonable value of 2, and select an epsilon value of 0.5, ensuring that 95\% of the point clouds are included in the region branch learning process. A detailed analysis of DBSCAN is included in Appendix J.  For the input point clouds, we set the voxel size as 0.05 and utilize the scale and rotation augmentation to generate two views. The scale range is from 0.95 to 1.05, and the rotation range is from -$\pi / 20$ to $\pi / 20$ for three axes. We train 10 epochs and set batch size as 4 for all experiments. The optimizer is Adamw, and the initial learning rate is 1e-3, which decreases to 1e-5 by a cosine schedule. For the hyperparameters, we set $\alpha=\beta =1$ and fix $\lambda$ at 0.5. 
We choose $T=10$ and $\rho=0.005$ based on the indicator mentioned in \cref{sec:selflabeing} and analyze them in the ablation study. Both the point- and region-level self-labeling algorithms employ the same parameters. All experiments are conducted on a single NVIDIA A100.
% \label{sec:id} We use ResNet18 as our backbone network for all dataset except Imagenet100, which we follow~\cite{cao2021open} to use ResNet50. The training consists of two stages. First, we utilize label data to supervise train 100 epochs and then train 200 or 90 epochs with unlabeled data. In the second stage, we adopt four views, two strong augmentation views and two weak augmentation views. And, we set the batch size to 512 for CIFAR datasets, and 256 for other datasets. The optimizer is SGD, and the learning rate first grows linearly and then cosine decays. Balance factor $\alpha,\beta$ is set to 1 in most of the experiments, and temperature $\tau$ is set to $0.1$. \textit{Without specially illustration, we adopt the self-labeling loss in the ablation study.} To reduce the effect of random factors, all results are the average of three repeated experiments. 
% DBSCAN has two key parameters: epsilon and min-samples. The explanation of parameters is in Appendix J. In our experiments, we set min-samples to be reasonable minimal 2, indicating that there must be at least two points in a region. For epsilon, we determine a value of 0.5 based on the proportion of outliers, ensuring that 95\% of the point clouds participate in region branch learning. 

\begin{table}[!t]
\centering
\scriptsize
\renewcommand\arraystretch{1.0}
\setlength\tabcolsep{1pt}
\caption{The novel class discovery results on SemanticPOSS dataset. `Number' denotes the number of points. `Full' denotes the results obtained by supervised learning. The gray values are the novel classes in each split.}
\label{tab:semantic_poss}
 
\begin{tabular}{cc|ccccccccccccc|ccc}
\toprule
\multicolumn{1}{c}{\rotatebox{45}{Split}} & \multicolumn{1}{c|}{\rotatebox{45}{Method}} & \multicolumn{1}{c}{\rotatebox{45}{bike}} & \multicolumn{1}{c}{\rotatebox{45}{build.}} & \multicolumn{1}{c}{\rotatebox{45}{car}} & \multicolumn{1}{c}{\rotatebox{45}{cone.}} & \multicolumn{1}{c}{\rotatebox{45}{fence}} & \multicolumn{1}{c}{\rotatebox{45}{grou.}} & \multicolumn{1}{c}{\rotatebox{45}{pers.}} & \multicolumn{1}{c}{\rotatebox{45}{plants}} & \multicolumn{1}{c}{\rotatebox{45}{pole}} & \multicolumn{1}{c}{\rotatebox{45}{rider}} & \multicolumn{1}{c}{\rotatebox{45}{traf.}} & \multicolumn{1}{c}{\rotatebox{45}{trashc.}} & \multicolumn{1}{c|}{\rotatebox{45}{trunk}} & \multicolumn{1}{c}{\rotatebox{45}{Novel}} & \multicolumn{1}{c}{\rotatebox{45}{Known}} & \multicolumn{1}{c}{\rotatebox{45}{All}} \\ \midrule
%&Number&$9.5e^6$ &$3.5e^7$&$1.4e^7$&$1.4e^5$& $2.4e^6$&$3.0e^7$ & $2.9e^6$& $5.8e^7$&$6.8e^5$&$5.7e^5$&$7.9e^5$ &$7.9e^4$ &$1.6e^6$&- &-&-\\ \midrule
& Full&45.0 & 83.3& 52.0&36.5& 46.7&77.6 & 68.2& 77.7&36.0&58.9&30.3 &4.2 &14.4&- &-&48.5\\ \midrule
\multirow{3}{*}{0} & EUMS & 25.7& \cellcolor{gray!30}{4.0}& \cellcolor{gray!30}{0.6} & 16.4& 29.4& \cellcolor{gray!30}{36.8} & 43.8& \cellcolor{gray!30}{28.5} & 13.1 & 26.8 & 18.2 & 3.3 & 16.9 & \cellcolor{gray!30}{17.4}& 21.5 & 20.3\\
& NOPS& 35.5& \cellcolor{gray!30}{30.4} & \cellcolor{gray!30}{1.2}& 13.5 & 24.1& \cellcolor{gray!30}{69.1} & 44.7 & \cellcolor{gray!30}{42.1} & 19.2& 47.7 & 24.4 & 8.2 & 21.8 & \cellcolor{gray!30}{35.7}& 26.6 & 29.4\\
& Ours& 46.3 & \cellcolor{gray!30}{\textbf{51.5}}& \cellcolor{gray!30}{\textbf{6.0}}& 35.7& 48.5& \cellcolor{gray!30}{\textbf{83.0}}& 67.9 & \cellcolor{gray!30}{\textbf{53.1}} & 35.5 & 59.3& 31.0 & 2.8 & 15.5 & \cellcolor{gray!30}{\textbf{48.4}} & 38.0 & 41.2 \\ \midrule
\multirow{3}{*}{1} & EUMS & \cellcolor{gray!30}{15.2}& 68.0& 28.0& 24.0 & \cellcolor{gray!30}{11.9}& 75.1 & \cellcolor{gray!30}{36.0}& 74.5 & 26.9& 48.6 & 26.0 & 5.6& 23.1& \cellcolor{gray!30}{21.0} & 40.0& 35.6 \\
& NOPS & \cellcolor{gray!30}{29.4} & 71.4 & 28.7 & 12.2& \cellcolor{gray!30}{3.9}& 78.2 & \cellcolor{gray!30}{\textbf{56.8}} & 74.2 & 18.3 & 38.9 & 23.3 & 13.7 & 23.5 & \cellcolor{gray!30}{30.0} & 38.2& 36.4 \\
& Ours & \cellcolor{gray!30}{\textbf{31.5}}& 83.2 & 48.7 & 25.4& \cellcolor{gray!30}{\textbf{23.9}} & 77.3 & \cellcolor{gray!30}{53.1} & 77.1& 32.5 & 57.3& 35.0 & 9.3 & 18.0 & \cellcolor{gray!30}{\textbf{36.2}} & 46.4& 44.0 \\ \midrule
\multirow{3}{*}{2} & EUMS & 40.1& 69.5 & 27.7 & 13.5 & 34.9 & 76.0 & 54.7 & 75.6 & \cellcolor{gray!30}{5.3} & 39.2 & \cellcolor{gray!30}{7.8} & 8.5 & \cellcolor{gray!30}{11.9} & \cellcolor{gray!30}{8.3} & 44.0 & 35.7 \\
& NOPS & 37.2 & 71.8 & 29.7& 14.6& 28.4& 77.5& 52.1& 73.0 & \cellcolor{gray!30}{\textbf{11.5}} & 47.1& \cellcolor{gray!30}{0.5} & 10.2& \cellcolor{gray!30}{\textbf{14.8}}& \cellcolor{gray!30}{9.0}& 44.2 & 36.0 \\
 & Ours& 45.3 & 82.8 & 49.8& 28.4& 46.3& 76.7& 66.2& 77.2 & \cellcolor{gray!30}{10.9} & 58.4& \cellcolor{gray!30}{\textbf{18.6}}& 7.3 & \cellcolor{gray!30}{8.2} & \cellcolor{gray!30}{\textbf{12.6}} & 53.8 & 44.3 \\ \midrule
\multirow{3}{*}{3} & EUMS& 41.2 & 70.7 & 28.1& \cellcolor{gray!30}{\textbf{4.3}} & 38.3& 76.7& 38.3& 75.4 & 25.8 & \cellcolor{gray!30}{34.3}& 28.3& \cellcolor{gray!30}{0.4} & 24.4& \cellcolor{gray!30}{13.0} & 44.7 & 37.4 \\
 & NOPS& 38.6 & 70.4 & 30.9& \cellcolor{gray!30}{0.0} & 29.4& 76.5& 56.0 & 71.8 & 17.0 & \cellcolor{gray!30}{31.9}& 26.2& \cellcolor{gray!30}{1.0} & 22.6& \cellcolor{gray!30}{10.9} & 43.9 & 36.3 \\
 & Ours& 45.5 & 82.9 & 47.7& \cellcolor{gray!30}{{0.0}} & 45.1& 77.8& 66.3& 77.7 & 34.3 & \cellcolor{gray!30}{\textbf{49.1}}& 35.6& \cellcolor{gray!30}{\textbf{4.0}} & 15.3& \cellcolor{gray!30}{\textbf{17.7}} & 52.8 & 44.7\\ \bottomrule
\end{tabular}
\end{table}

\begin{table}[!t]
\centering
\scriptsize
\renewcommand\arraystretch{1.0}
\setlength\tabcolsep{1pt}
\caption{Results on splits of SemanticPOSS dataset with more severe imbalance. The gray values are the novel classes in each split. NOPS is based on its released code.}
 
\label{tab:semantic_poss_harder}
\begin{tabular}{cc|ccccccccccccc|ccc}
\toprule
\multicolumn{1}{c}{\rotatebox{45}{Split}} & \multicolumn{1}{c|}{\rotatebox{45}{Method}} & \multicolumn{1}{c}{\rotatebox{45}{bike}} & \multicolumn{1}{c}{\rotatebox{45}{build.}} & \multicolumn{1}{c}{\rotatebox{45}{car}} & \multicolumn{1}{c}{\rotatebox{45}{cone.}} & \multicolumn{1}{c}{\rotatebox{45}{fence}} & \multicolumn{1}{c}{\rotatebox{45}{grou.}} & \multicolumn{1}{c}{\rotatebox{45}{pers.}} & \multicolumn{1}{c}{\rotatebox{45}{plants}} & \multicolumn{1}{c}{\rotatebox{45}{pole}} & \multicolumn{1}{c}{\rotatebox{45}{rider}} & \multicolumn{1}{c}{\rotatebox{45}{traf.}} & \multicolumn{1}{c}{\rotatebox{45}{trashc.}} & \multicolumn{1}{c|}{\rotatebox{45}{trunk}} & \multicolumn{1}{c}{\rotatebox{45}{Novel}} & \multicolumn{1}{c}{\rotatebox{45}{Known}} & \multicolumn{1}{c}{\rotatebox{45}{All}} \\ \midrule
& Full&45.0 & 83.3& 52.0&36.5& 46.7&77.6 & 68.2& 77.7&36.0&58.9&30.3 &4.2 &14.4&- &-&48.5\\ \midrule
\multirow{2}{*}{0} & NOPS& 37.4& \cellcolor{gray!30}{22.9} & \cellcolor{gray!30}{8.1}& \cellcolor{gray!30}{0.0} & 30.3 & 78.9 & \cellcolor{gray!30}{4.8} & 72.9 & \cellcolor{gray!30}{\textbf{1.0}} & 42.9 & 25.8 & 9.2 & \cellcolor{gray!30}{\textbf{9.6}} & \cellcolor{gray!30}{7.7}& 42.5 & 26.5\\
& Ours& 46.0& \cellcolor{gray!30}{\textbf{26.1}} & \cellcolor{gray!30}{\textbf{27.5}} & \cellcolor{gray!30}{\textbf{2.8}} & 46.9 & 77.6 & \cellcolor{gray!30}{\textbf{35.0}} & 77.8 & \cellcolor{gray!30}{0.2} & 58.6 & 30.5 & 3.2 & \cellcolor{gray!30}{0.0} & \cellcolor{gray!30}{\textbf{15.3}}& 48.7 & 33.2 \\ \midrule
\multirow{2}{*}{1} & NOPS & \cellcolor{gray!30}{6.1} & 71.3 & 35.6 & 21.2& \cellcolor{gray!30}{3.1}& \cellcolor{gray!30}{\textbf{42.9}} & 44.5 & \cellcolor{gray!30}{26.0} & 24.4 & \cellcolor{gray!30}{\textbf{0.7}} & \cellcolor{gray!30}{0.6} & \cellcolor{gray!30}{\textbf{0.1}} & 24.8 & \cellcolor{gray!30}{11.4} & 37.0& 23.2 \\
& Ours & \cellcolor{gray!30}{\textbf{26.3}} & 82.0 & 51.4 & 18.0 & \cellcolor{gray!30}{\textbf{10.4}}& \cellcolor{gray!30}{40.0} & 67.5 & \cellcolor{gray!30}{\textbf{32.5}} & 31.2 & \cellcolor{gray!30}{0.0} & \cellcolor{gray!30}{\textbf{6.3}} & \cellcolor{gray!30}{0.0} & 11.7 & \cellcolor{gray!30}{\textbf{16.5}} & 44.5 & 29.4 \\ 
\bottomrule
\end{tabular}
 
\end{table}

\subsection{Results}\label{sec:res}

\paragraph{SemanticPOSS Dataset.}
% As shown in \cref{tab:semantic_poss}, on novel classes, we outperform previous method on four split by a large margin, especially on split 0 and 1, where we improve SOTA by \textbf{18.4\%} and \textbf{9.3\%}. And the fully supervised upper bound of novel classes on split 0 and 1 are 72.7\% and 53.3\%, which is higher than our method by 18.6\% and 14.0\%, respectively. On the challenging split 2 and split 3, we achieve \textbf{4.9\%} and \textbf{2.8\%} gains, respectively. And the corresponding supervised upper bound are 26.9\% and 33.2\%, indicating its challenging compared to split 0 and 1. On the average, we achieve 30.5\% IOU on novel classes over four splits, which is higher than NOPS (21.4\%) by \textbf{9.1\%}.  ours head 53.18 nops head 46.83
% ours medium 36.5 nops medium 21.93 ours tail 10.6 nops tail 4.15
As presented in \cref{tab:semantic_poss}, our approach exhibits significant improvements in novel classes over the previous method across all four splits. Specifically, we achieve an increase of \textbf{12.7\%} and \textbf{6.2\%} in split 0 and 1, respectively. 
It is worth noting that the fully supervised upper bounds for novel classes in split 0 and 1 are 72.7\% and 53.3\%, respectively, and the performance gaps have been significantly reduced.  %surpassing our method by 24.3\% and 17.1\%.
In the more challenging split 2 and split 3, we observe gains of \textbf{3.6\%} and \textbf{4.7\%}, respectively. The corresponding upper bounds for these splits are 26.9\% and 33.2\%, indicating their increased difficulty compared to splits 0 and 1. On average, we achieve an IoU of 30.2\% for novel classes across all four splits, outperforming NOPS (22.5\%) by \textbf{7.7\%}.
In addition, we provide a detailed comparison with NOPS on head, medium, and tail classes in Appendix D, as well as under a more comparable setting that applies our training strategy to NOPS in Appendix E.

% 下面两段是原来版本的描述
% Compared with NOPS, our method achieves improvements in head(7.5\%), medium(8.9\%), and tail(6.8\%) classes. Details of dataset distribution are in Appendix D.

% Note that the improvement in known classes can be attributed to the training of NOPS not converging. In the Appendix E, we compare our method and NOPS*, which training is converging. The results show that, compared with NOPS, NOPS* achieves sizeable improvements in known classes but drops a lot in novel classes. Therefore, our method still outperforms NOPS* by a sizeable margin.

%ours 44.98 30.8  11.2
%nops 37.46 21.925 4.4

To further verify that our method can alleviate the imbalanced problem, we divided the SemanticPOSS dataset into two splits, creating a more severe imbalance scenario that poses a greater challenge for clustering novel classes. As shown in \cref{tab:semantic_poss_harder}, on novel classes, our method outperforms NOPS significantly on both splits, with a margin of \textbf{7.6\%} on split 0 and \textbf{5.1\%} on split 1. In particular, for the novel classes, we observe that our improvement mainly stems from the medium classes, such as person and bike. It is worth noting that NOPS employs extra training techniques, such as multihead and overclustering, whereas we use a simpler pipeline without needing them, further demonstrating our effectiveness.

%具体来说，由图2的第一行可得，NOPs由于均匀约束，预测结果非常noisy，head classes（building）和tail class （car）混合在一起。而ours在building和car上都得到不错的结果，这是因为ours adaptive regularization 和dual-level的representation learning 生成了高质量的imbalanced pseudo label

%have created a video to visualize the results of NOPS and our method on Split 0, which is included in the supplementary material.
%utilization of a learning rate of 0.01 in conjunction with SGD, which is excessively low, resulting in training 

%We conduct visualization on SemanticKITTI, and the visualizations for different splits are presented below.

\paragraph{SemanticKITTI Dataset.} The results in \cref{tab:semantic_kitti} demonstrate our superior performance compared to previous methods on different splits. Specifically, we achieve significant improvements of \textbf{8.6\%}, \textbf{3.3\%}, and \textbf{3.6\%} on splits 0, 1, and 2, respectively, for novel classes. The supervised upper bounds for these splits are 82.0\%, 42.4\%, and 39.6\%, respectively. In split 3, our results are slightly higher than NOPS by 0.2\%, possibly due to the scarce presence of these novel classes in split 3. On average across all four splits, our approach achieves an IoU of 27.5\%, surpassing NOPS (23.4\%) by \textbf{4.1\%} on novel classes. 

%与semanticposs结果类似，ours的预测结果大大减小了noisy，例如图2第二行的building与plants，第三行的car和parking。

\begin{figure}[!t]
    \centering
    \includegraphics[scale=0.23]{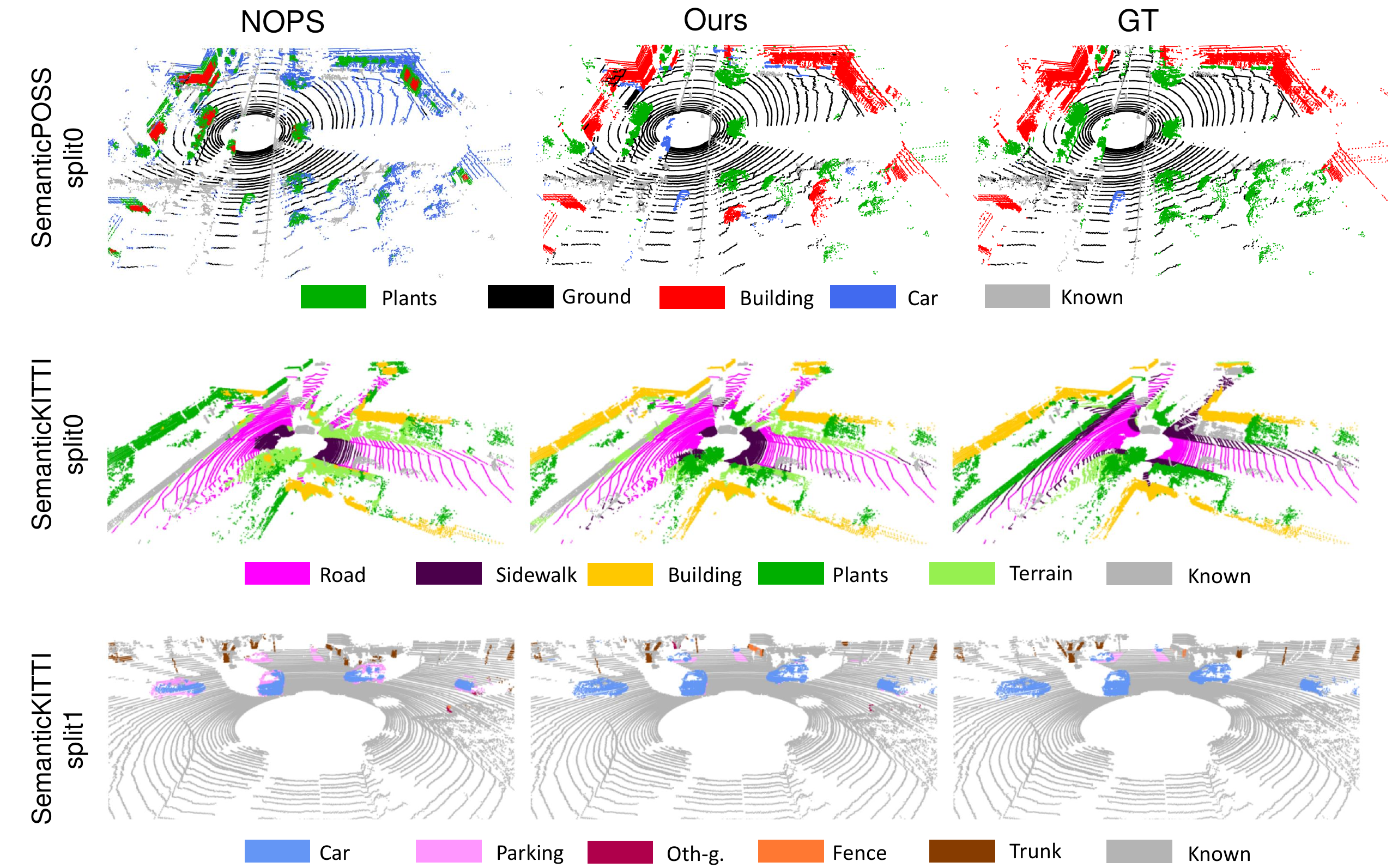}
    \caption{Visualization comparison between Our method and NOPS on the SemanticPOSS and SemanticKITTI datasets. In the first and second rows, compared to NOPS, our method achieves much better segmentation for the `Building', significantly reducing confusion with tail classes (such as `Car') or medium classes (like `Plants'). In the third row, our approach generates better segmentation for `Parking' and `Car'.}
    \label{fig:video}
    
\end{figure}

\paragraph{Visualization Analysis.}
Additionally, in \cref{fig:video}, we perform visual comparisons on the results between NOPS and our method, and it is evident that our method shows significant improvements compared to NOPS. Specifically, as shown in the first row of \cref{fig:video}, NOPS produces noisy predictions due to uniform constraints, mixing medium classes (e.g., building) and tail classes (e.g., car). 
In the second and third rows of \cref{fig:video}, NOPS often confuses between medium and head classes, such as building and plants, as well as parking and car.
In contrast, our method achieves better results for both datasets due to adaptive regularization and dual-level representation learning, generating high-quality imbalanced pseudo labels. More visual comparisons for additional splits are provided in the Appendix K.

% In \cref{fig:video}, we also visualize the comparison between NOPS and our method on SemanticKITTI, demonstrating significant performance improvements. 
% For instance, in the second row of \cref{fig:video}, categories such as building and plants improve, as well as car and parking in the third row. Please refer to Appendix D %\ref{appendix:anaysis} 
% for detailed analysis. 
%and Appendix L
%\ref{suppl:more_vis} 
%for visualization. 
% More details analysis are in Appendix \ref{appendix:anaysis}, and visualization analysis in Appendix \ref{suppl:more_vis}.

% To address this issue, NOPS incorporates a queue-based mechanism.

% \usepackage{multirow}
\begin{table}[!t]
\centering
\tiny
\scriptsize

\renewcommand\arraystretch{1.0}
\setlength\tabcolsep{0.1pt}
\caption{The novel class discovery results on the SemanticKITTI dataset. `Full' denotes the results obtained by supervised learning. The four groups represent the four splits in turn, and the gray values are the novel classes in each split.}
\label{tab:semantic_kitti}
\begin{adjustbox}{width=\textwidth}
\begin{tabular}{c|ccccccccccccccccccc|ccc}
\toprule
\rotatebox{45}{Method} & \rotatebox{45}{bi.cle} & \rotatebox{45}{b.clst}  & \rotatebox{45}{build.}  & \rotatebox{45}{car}     & \rotatebox{45}{fence}   & \rotatebox{45}{mt.cle}  & \rotatebox{45}{m.clst}  & \rotatebox{45}{oth-g.} & \rotatebox{45}{oth-v.}  & \rotatebox{45}{park.}   & \rotatebox{45}{pers.}   & \rotatebox{45}{pole}    & \rotatebox{45}{road}    & \rotatebox{45}{side2.}  & \rotatebox{45}{terra.}  & \rotatebox{45}{traff.}  & \rotatebox{45}{truck}   & \rotatebox{45}{trunk}   & \rotatebox{45}{veget.}  & \rotatebox{45}{Novel}   & \rotatebox{45}{Known} & \rotatebox{45}{All}  \\ \midrule
% Number   & $1.8e^5$    &  $1.2e^5$   & $1.4e^8$   & $3.7e^7$ & $6.6e^7$    & $3.6e^5$    & $4.1e^4$  &   $4.1e^6$    & $2.1e^6$    & $1.3e^7$    &$3.4e^5$   &$2.7e^6$ & $1.7e^8$    &  $1.2e^8$    & $7.8e^7$    & $6.6e^5$    & $1.7e^6$    & $5.9e^6$    & $2.7e^8$    & -        & - & - \\ \midrule
Full   & 2.9    & 55.4    & 89.5    & 93.5    & 27.9    & 27.4    & 0.0     & 0.9    & 19.9    & 35.8    & 31.2    & 60.0    & 93.5    & 77.8    & 62.0    & 39.8    & 50.8    & 53.9    & 87.0    & -        & - & 47.9 \\ \midrule
EUMS   & 5.3    & 40.0    & \cellcolor{gray!30}{15.8} & 79.2    & 9.0     & 16.9    & 2.5     & 0.1    & 11.4    & 14.4    & 12.7    & 29.2    & \cellcolor{gray!30}{42.6} & \cellcolor{gray!30}{\textbf{26.1}} & \cellcolor{gray!30}{0.1}  & 10.3    & 47.4    & 37.9    & \cellcolor{gray!30}{38.4} & \cellcolor{gray!30}{24.6} & 21.1  & 23.1 \\
NOPS   & 5.6    & 47.8    & \cellcolor{gray!30}{52.7} & 82.6    & 13.8    & 25.6    & 1.4     & 1.7    & 14.5    & 19.8    & 25.9    & 32.1    & \cellcolor{gray!30}{\textbf{56.7}} & \cellcolor{gray!30}{8.1}  & \cellcolor{gray!30}{\textbf{23.8}} & 14.3    & 49.4    & 36.2    & \cellcolor{gray!30}{44.2} & \cellcolor{gray!30}{37.1} & 26.5  & 29.3 \\
Ours   & 5.5    & 51.1    & \cellcolor{gray!30}{\textbf{74.6}} & 92.3    & 29.8    & 22.8    & 0.0     & 0.0    & 23.3    & 24.8    & 27.7    & 59.7    & \cellcolor{gray!30}{41.4} & \cellcolor{gray!30}{22.5} & \cellcolor{gray!30}{23.6} & 39.3    & 43.6    & 51.1    & \cellcolor{gray!30}{\textbf{66.4}} & \cellcolor{gray!30}{\textbf{45.7}} & 33.7  & 36.8 \\ \midrule
EUMS   & 7.5    & 42.4    & 80.0    & \cellcolor{gray!30}{\textbf{76.8}} & \cellcolor{gray!30}{8.6}  & 19.6    & 1.4     & \cellcolor{gray!30}{\textbf{0.6}} & 12.0    & \cellcolor{gray!30}{14.1} & 14.0    & 40.7    & 86.3    & 66.5    & 56.3    & 12.0    & 44.8    & \cellcolor{gray!30}{20.9} & 72.4    & \cellcolor{gray!30}{24.2} & 37.1  & 35.6 \\ 
NOPS   & 7.4    & 51.2    & 84.5    & \cellcolor{gray!30}{50.9} & \cellcolor{gray!30}{7.3}  & 28.9    & 1.8     & \cellcolor{gray!30}{0.0} & 22.2    & \cellcolor{gray!30}{19.4} & 30.4    & 37.6    & 90.1    & 72.2    & 60.8    & 16.8    & 57.3    & \cellcolor{gray!30}{\textbf{49.3}} & 85.1    & \cellcolor{gray!30}{25.4} & 46.2  & 40.7 \\
Ours   & 3.7    & 57.4    & 89.2    & \cellcolor{gray!30}{56.5} & \cellcolor{gray!30}{\textbf{17.3}} & 20.3    & 0.0     & \cellcolor{gray!30}{0.0} & 20.0    & \cellcolor{gray!30}{\textbf{30.6}} & 34.8    & 60.6    & 93.2    & 77.6    & 62.0    & 38.7    & 56.9    & \cellcolor{gray!30}{39.2} & 86.7    & \cellcolor{gray!30}{\textbf{28.7}} & 50.1  & 44.5 \\ \midrule
EUMS   & 8.3    & 50.8    & 83.0    & 88.1    & 17.9    & \cellcolor{gray!30}{2.8}  & 2.3     & 0.2    & \cellcolor{gray!30}{3.2}  & 25.4    & 25.0    & \cellcolor{gray!30}{20.2} & 88.3    & 71.0    & 57.9    & \cellcolor{gray!30}{8.6}  & \cellcolor{gray!30}{27.2} & 38.4    & 77.0    & \cellcolor{gray!30}{12.4} & 42.2  & 36.6 \\
NOPS   & 6.7    & 49.2    & 86.4    & 90.8    & 23.7    & \cellcolor{gray!30}{2.7}  & 0.6     & 1.9    & \cellcolor{gray!30}{\textbf{15.5}} & 29.5    & 27.9    & \cellcolor{gray!30}{\textbf{36.4}} & 90.3    & 73.4    & 61.2    & \cellcolor{gray!30}{\textbf{17.8}} & \cellcolor{gray!30}{10.3} & 46.2    & 84.3    & \cellcolor{gray!30}{16.5} & 48.0  & 39.7 \\
Ours   & 3.6    & 54.2    & 88.9    & 93.3    & 28.4    & \cellcolor{gray!30}{\textbf{10.2}} & 0.0     & 0.9    & \cellcolor{gray!30}{9.6}  & 33.4    & 32.2    & \cellcolor{gray!30}{36.1} & 92.7    & 77.4    & 62.2    & \cellcolor{gray!30}{10.7}  & \cellcolor{gray!30}{\textbf{34.2}} & 51.7    & 86.9    & \cellcolor{gray!30}{\textbf{20.1}} & 50.4  & 42.5 \\ \midrule
EUMS   & \cellcolor{gray!30}{\textbf{4.0}} & \cellcolor{gray!30}{2.5}  & 80.1    & 87.2    & 16.8    & 14.0    & \cellcolor{gray!30}{\textbf{15.0}} & 0.3    & 14.1    & 20.8    & \cellcolor{gray!30}{6.8}  & 37.6    & 86.8    & 66.5    & 55.3    & 16.2    & 40.6    & 38.4    & 76.2    & \cellcolor{gray!30}{7.1}  & 43.4  & 35.7 \\
NOPS   & \cellcolor{gray!30}{2.3} & \cellcolor{gray!30}{27.8} & 86.0    & 89.9    & 23.1    & 24.5    & \cellcolor{gray!30}{2.9}  & 3.1    & 18.2    & 30.1    & \cellcolor{gray!30}{\textbf{16.3}} & 39.9    & 90.7    & 73.5    & 61.0    & 17.4    & 49.8    & 44.0    & 83.2    & \cellcolor{gray!30}{12.4} & 49.0  & 41.2 \\
Ours   & \cellcolor{gray!30}{2.6} & \cellcolor{gray!30}{\textbf{32.5}} & 88.7    & 93.3    & 28.1    & 24    & \cellcolor{gray!30}{0.1}  & 1.0    & 23.7    & 35.6    & \cellcolor{gray!30}{15.3} & 59.8    & 93.2    & 77.6    & 61.4    & 37.8    & 56.6    & 52.1    & 86.7    & \cellcolor{gray!30}{\textbf{12.6}}  & 54.6  & 45.8 \\ \bottomrule
\end{tabular}
\end{adjustbox}
 
\end{table}

% \paragraph{Comparison with SOTA}

% The results show our method still outperforms NOPS by a large margin. 

%In addition, we conduct experiments in SemanticPOSS split 0 to compare the results of ours method and previous sota\cite{riz2023novel} with the estimated $|C_u|$. As shown in the table below, our method can still significantly outperform NOPS with estimated $|C_u|$.

%  \begin{table}[H]
%     \centering
%     \caption{ \textcolor{red}{Estimated novel classes number in SemanticPOSS split 0}}
% \label{tab:estimatesnumber}
% \begin{tabular}{@{}ccc@{}}
%     \toprule
%     max classes             & estimated $|C_u|$ & GT \\ \midrule
%     \multicolumn{1}{c|}{30} & 2                 & 4  \\
%     \multicolumn{1}{c|}{40} & 2                 & 4  \\
%     \multicolumn{1}{c|}{50} & 3                 & 4  \\ \bottomrule
%     \end{tabular}
% \end{table}

% The results indicate that there may be errors in estimating $|C_u|$ using this method. In addition, we conduct experiments on SemanticPoss split 0 with 2,3,4,5,6 classes(the correct number of classes is 4). As shown in the table below, when there is an error between the $|C_u|$ value and the ground truth, our method can also be significantly higher than NOPS. 

\subsection{Ablation Study}

\begin{figure}[!t]
  \centering
  \begin{subfigure}{0.22\textwidth}
    \centering
     \includegraphics[width=\linewidth]{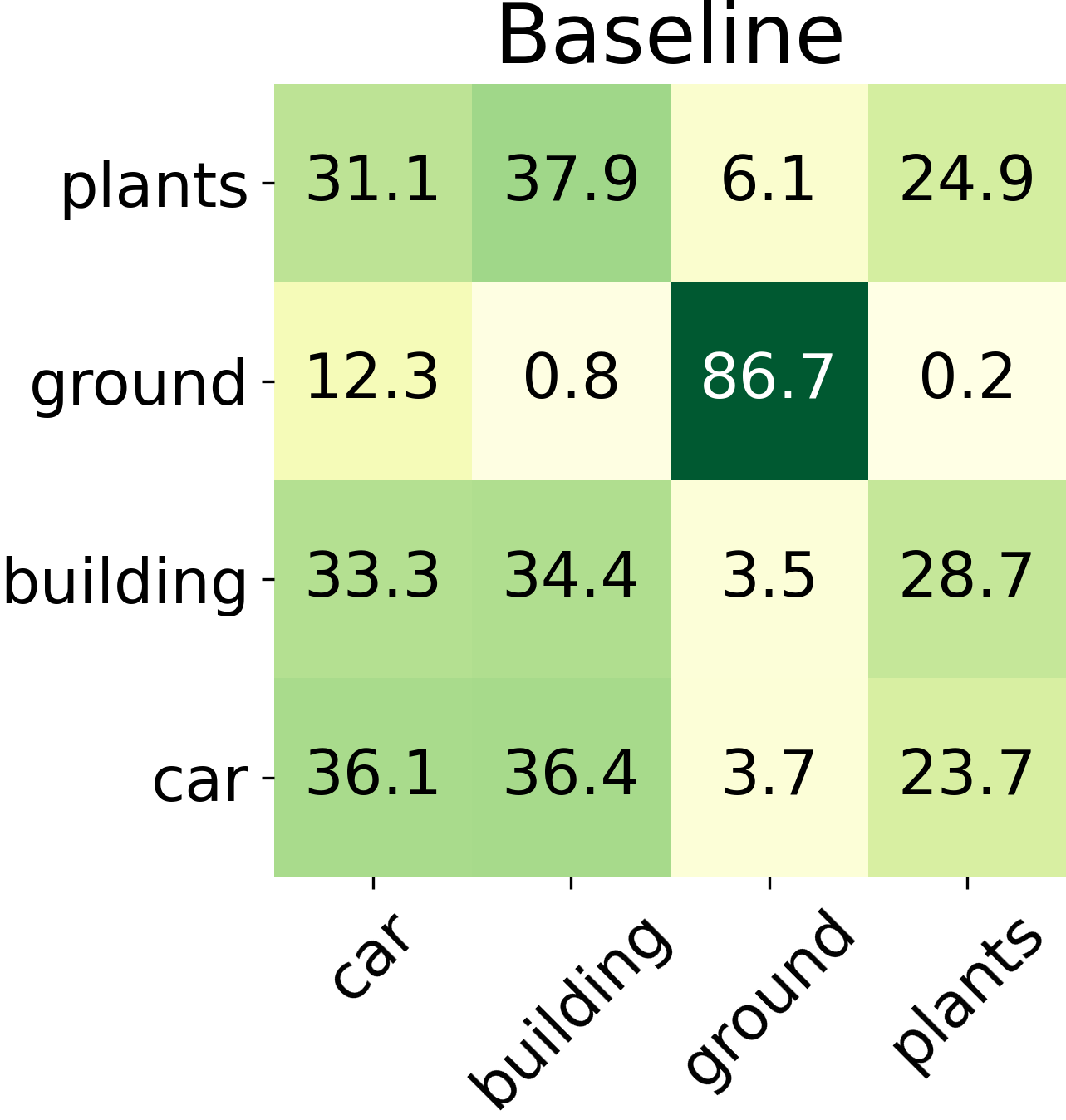}
  \end{subfigure}%
  \hfill
  \begin{subfigure}{0.22\textwidth}
    \centering
    \includegraphics[width=\linewidth]{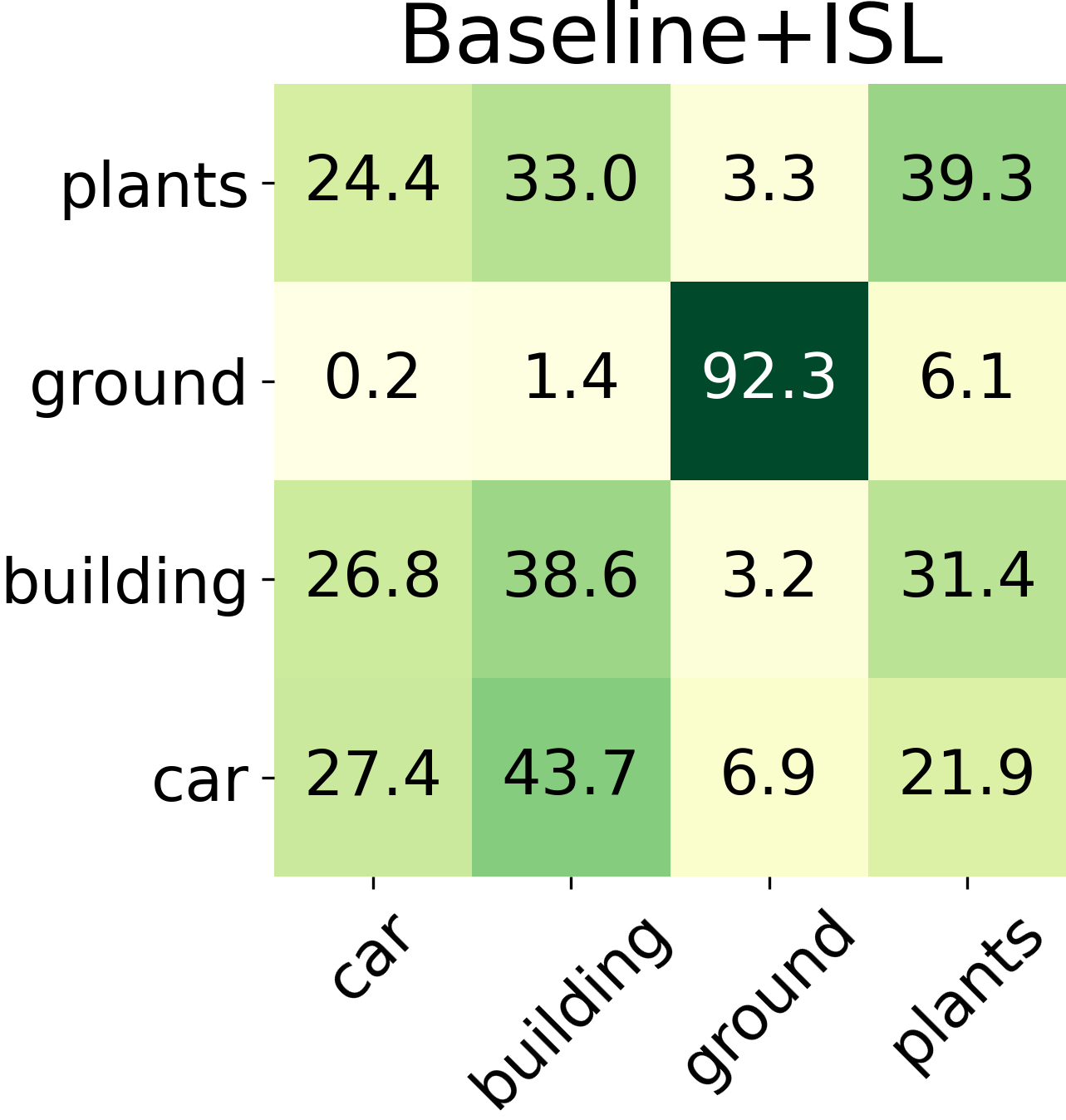}
  \end{subfigure}%
  \hfill
  \begin{subfigure}{0.22\textwidth}
    \centering
    \includegraphics[width=\linewidth]{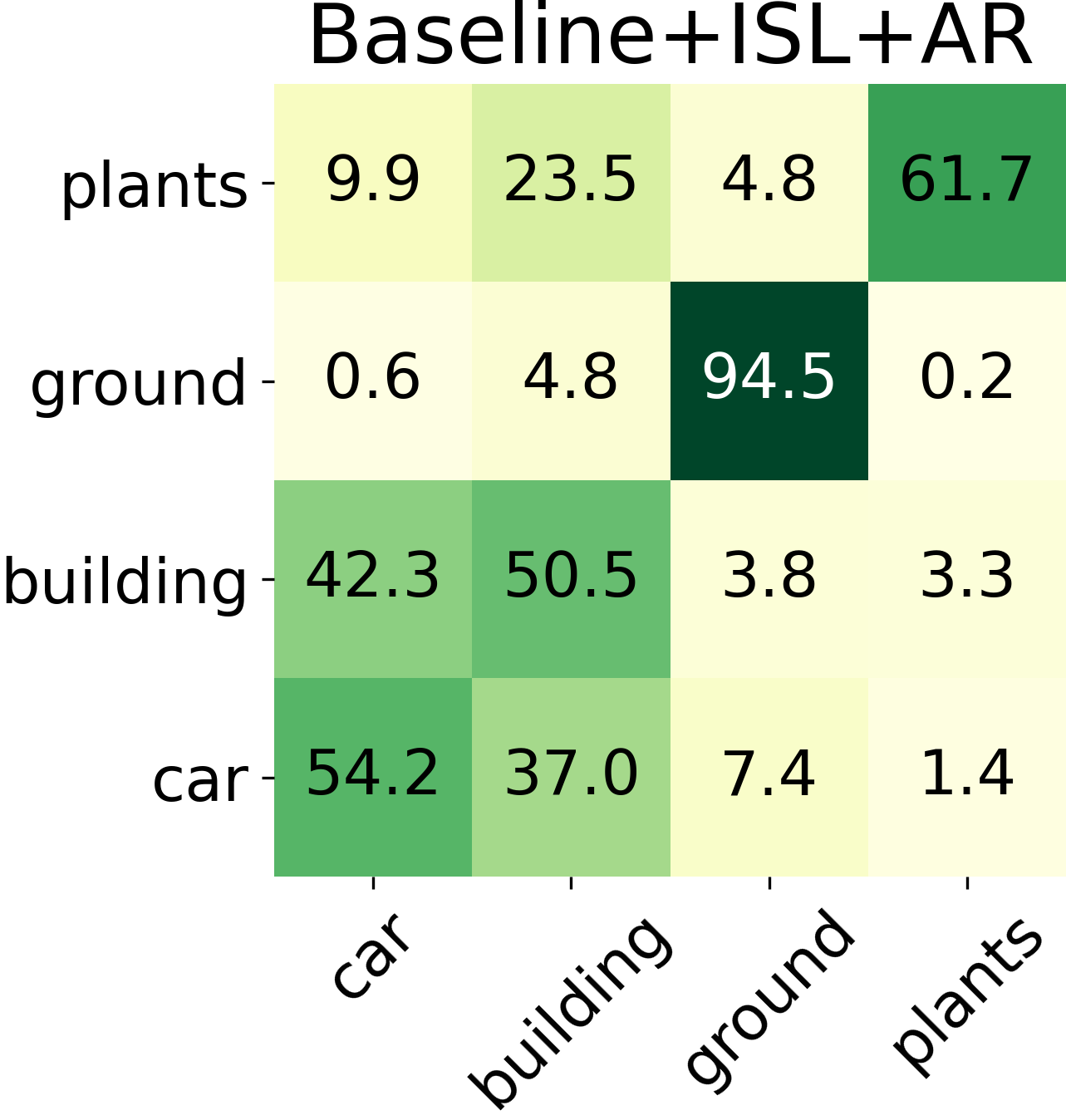}
  \end{subfigure}%
  \hfill
  \begin{subfigure}{0.22\textwidth}
    \centering
    \includegraphics[width=\linewidth]{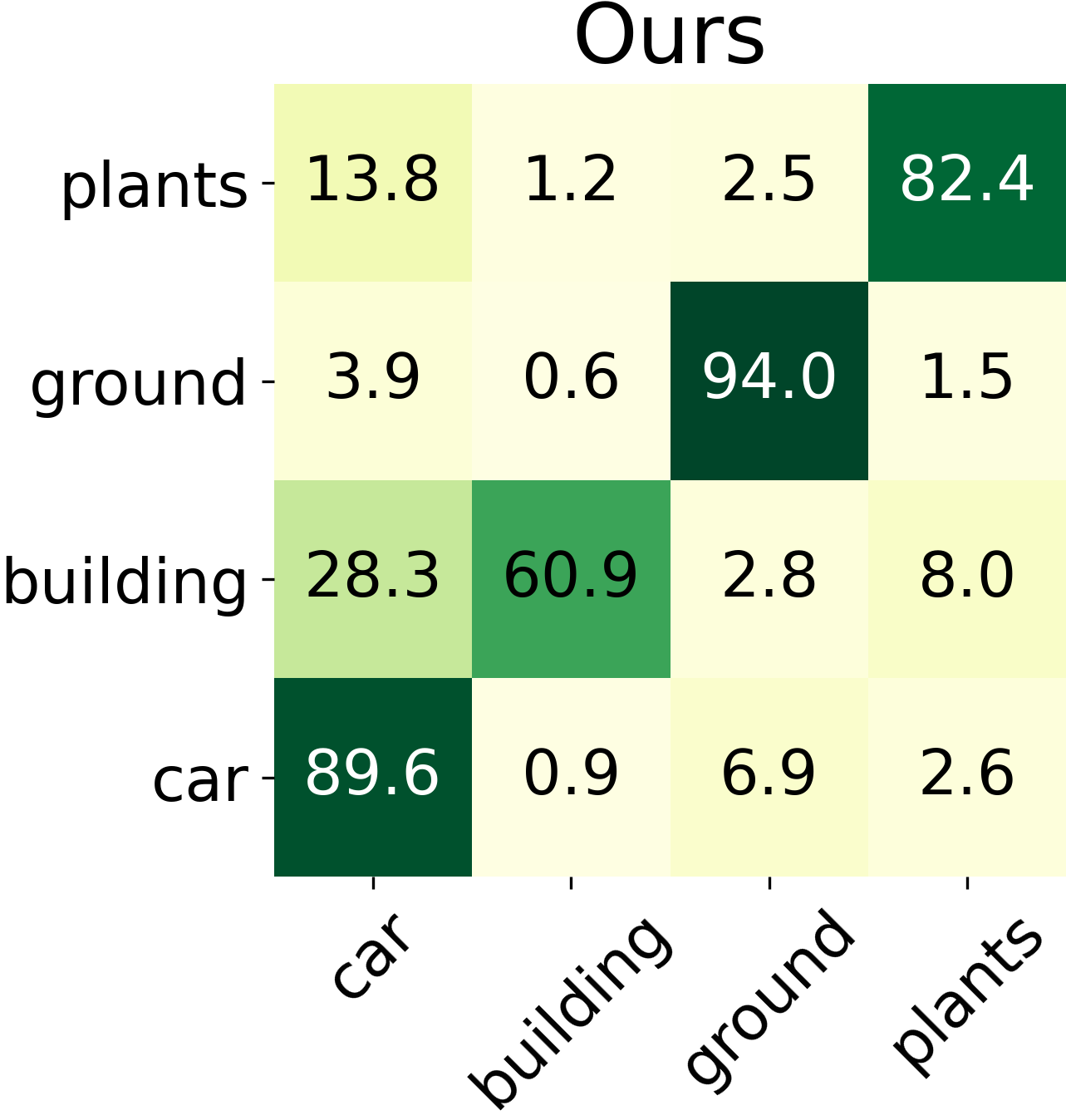}
  \end{subfigure}
  \caption{Confusion Matrix, GT on the y-axis, Pseudo Label on the x-axis. $(i, j)$ 
 represents the \% of GT in class $j$ assigned pseudo label $i$. We categorize `plants' and `ground' as head classes, `building' as medium, and `car' as tail classes.}
  \label{fig:cost}
  
\end{figure}

\begin{figure}[!t]
    \centering
    \begin{subfigure}{\textwidth}
        \centering
        \includegraphics[scale=0.6]{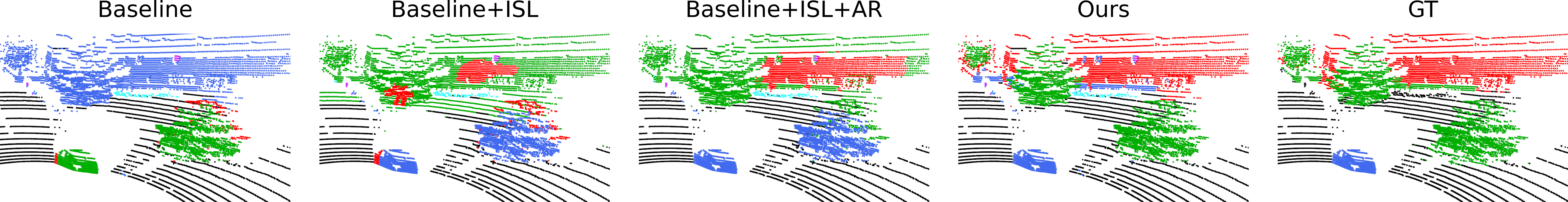}
    \end{subfigure}
    \begin{subfigure}{\textwidth}
        \centering
        \includegraphics[scale=0.1]{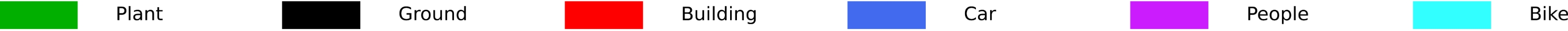}
    \end{subfigure}
    \caption{Visualization analysis. The introduction of ISL notably reduces the misclassification between `Plant' and `Car'. Then, the integration of AR further mitigates the confusion between `Plant' and `Building'. Ultimately, the incorporation of Region component (Ours) effectively minimizes the mix-up between `Plant', `Car', and `Building'.}
    \label{fig:ablation}
    
\end{figure}

\paragraph{Component Analysis.}
To analyze the effectiveness of each component, we conduct extensive experiments on split 0 of the SemanticPOSS dataset. Here we provide ablation on three components, including Imbalanced Self-Labeling (ISL), Adaptive Regularization (AR), and Region-Level Branch (Region).  
% , and the results are presented in \cref{tab:component}.
% \cref{fig:cost} visualizes the confusion matrix of the training set for novel classes to show the quality of pseudo-labels, while \cref{fig:ablation} provides a visualization.
As shown in \cref{tab:component}, compared to baseline which employs equal-size constraints, imbalanced self-labeling improves performance by \textbf{4.2\%}. The confusion matrix in \cref{fig:cost} indicates that except for the highly-accurate class ``ground", there is a significant improvement in the head and medium classes. This phenomenon is clearly depicted in \cref{fig:ablation}, where the predictions of the baseline exhibit noticeable noise.
%Notably, in \cref{tab:component}, the first and second rows reveal a significant \textbf{4.2\%} increase in novel mIoU with the incorporation of imbalanced self-labeling. This improvement is primarily observed in the head classes (ground, plants, building). The rationale behind this enhancement lies in the fact that the baseline approach enforces equal cluster sizes during point cloud clustering for novel classes. 
%However, this doesn't align with the distribution of the original point cloud data, resulting in the head classes (plants, building) becoming mixed with the tail class (car). 
%where the Baseline assigns a significant number of point clouds to the tail class (car) compared to Baseline+ISL, resulting in noisy pseudo-labels.
From the second and third rows of \cref{tab:component}, the adaptive regularization leads to a significant improvement of \textbf{8.2\%} in split0 and \textbf{4.5\%} in overall splits. As shown in \cref{fig:cost}, adaptive regularization enhances the quality of pseudo-labels for each class, especially for the head class (plants). We also visualize the class distribution of pseudo-labels in Appendix  F%\ref{appendix:adaptive_reg}
, which shows adaptive regularization provides greater flexibility than fixed regularization term. 
% As mentioned in \cref{sec:selflabeing}, adaptive regularization provides greater flexibility than a fixed regularization factor,  (see Appendix \ref{appendix:adaptive_reg}). 
% The superior experimental results validate the effectiveness of our adaptive regularization.
%allowing the model to focus more on learning meaningful cluster information and generating pseudo labels that better match the actual point cloud distribution.
%Experimentally, this adjustment makes the distribution of pseudo-labels align more closely with the actual point cloud distribution.  
%%%suo 符合xx
According to the third, fourth and
last rows of \cref{tab:component}, the inclusion of the region-level branch leads to a \textbf{9.1\%} improvement and an additional \textbf{4.2\%} improvement built upon the AR. In addition, more experiments and analysis on prototype learning are included in Appendix G. 
In \cref{fig:cost}, there's a significant improvement in pseudo-labels for each category, particularly for the tail class (car) and the head class (plants). From \cref{fig:ablation}, it is evident that the region-level branch can correct cases where a single object is mistakenly labeled as multiple categories. Due to the utilization of spatial priors, where closely-located points are highly likely to belong to the same category, our region-level branch can correct misclassifications by considering context from neighboring points, preventing splitting a single object into multiple entities. Those experiments validate the effectiveness of each component in our method.

\begin{table}[t]
\caption{Ablation study on SemanticPOSS, focusing on novel classes.
\textbf{ISL} and \textbf{AR} denote imbalanced self-labeling and regularization. \textbf{Region} denotes region-level learning. The last two columns represent the average mIoU for split 0 and across all splits.}
 
\label{tab:component}
\centering
\resizebox{0.55\textwidth}{!}{
  \begin{tabular}{@{}ccc|ccccc|c@{}}
    \toprule
                 &                           &                               & \multicolumn{5}{c|}{Split0}                                    & Overall \\
    ISL          & {\color[HTML]{1F1F1F} AR} & {\color[HTML]{1F1F1F} Region} & {\color[HTML]{1F1F1F} Building} & Car & Ground & Plants & Avg  & Avg     \\ \midrule
                 & {\color[HTML]{1F1F1F} }   & {\color[HTML]{1F1F1F} }       & {\color[HTML]{1F1F1F} 21.6}     & 2.7 & 76.6   & 26.1   & 31.8 & 20.9    \\
    $\checkmark$ & {\color[HTML]{1F1F1F} }   & {\color[HTML]{1F1F1F} }       & {\color[HTML]{1F1F1F} 27.6}     & 3.1 & 81.2   & 32.1   & 36.0 & 23.9    \\
    
    $\checkmark$ & $\checkmark$              &                               & 53.1                            & 5.3 & 81.1   & 37.4   & 44.2 & 28.4    \\
    $\checkmark$   &    & $\checkmark$      & 41.9     & 9.3 & 83.6   & 45.6   & 45.1 & 26.9 \\
    
    $\checkmark$ & $\checkmark$              & $\checkmark$                  & 51.5                            & 6.0 & 83.0   & 53.1   & 48.4 & 30.2    \\ \bottomrule
    \end{tabular}
    }
 
\end{table}

\begin{table}[!t]
\centering
\footnotesize
\caption{Analysis of adaptive regularization on SemanticPOSS dataset. GT denotes we directly assign the ground truth distribution of cluster size.}

\label{tab:gamma_analysis}
\resizebox{0.6\textwidth}{!}{
\begin{tabular}{@{}c|ccccccc|cc@{}}
\toprule
$\gamma$                      & 0.01                        & 0.05                        & 0.1                         & 0.5                         & 1                           & 5                           & $+\infty$                                        & GT                                              & Adaptive                                        \\ \midrule
{ Split0} & { 10.1} & { 33.8} & { 33.3} & { 36.0} & { 32.2} & { 33.8} & 31.8 & \multicolumn{1}{c}{{32.5}} & \multicolumn{1}{c}{{ 44.2}} \\ \bottomrule
\end{tabular}}
\end{table}

\paragraph{Estimate the number of novel classes.} \label{sec:estimatesk}
%Moreover, to validate the effectiveness of our method in realistic scenarios, where the number of novel classes is unknown, we estimate the number of novel classes and conduct experiments. To estimate $|C^u|$, we extend the classic estimation method~\cite{vaze2022generalized} in NCD to point clouds semantic segmentation. Specifically, we extract representation from a known-class pre-trained model for training data, set the candidate range of the total number of categories ( $|C^s| \textless |C_{all}|\textless$ max classes), and apply Kmeans to cluster the labeled and unlabeled point clouds. Then, we evaluate the clustering performance of known classes under different $|C_{all}|$, and select $|C_{all}|$ with the highest clustering performance as the estimated $|C_{all}|$.
For computational simplicity, we conduct experiments on splits 0 of the SemanticPOSS dataset and randomly sample 800,000 points from all scenes to estimate $|C^u|$. We set max classes to 50, which is an estimate of the maximum number of new classes that might appear in a typical scene. The estimated $|C^u|$ is 3, which is close to the ground truth value (GT is 4). Finally, we conduct experiments with $|C^u|$ as 3. As \cref{tab:error} illustrated our method still outperforms NOPS by a large margin.

\begin{table}[t]
\centering
\caption{Comparison between Ours and NOPS with the estimated number of novel classes in Split 0 of SemanticPOSS. The estimated $C^u$ is 3, and the ground truth is 4.}
\label{tab:error}

\resizebox{0.45\textwidth}{!}{\begin{tabular}{@{}c|ccccc@{}}
    \toprule
    Method     &  Building  & Car       & Ground  & Plants    & Avg \\ \midrule
    % \multirow{2}{*}{K=2} & NOPS      & 0.00                          & 0.00                          & 46.87                       & 42.41                         & 22.32                    \\
    %                  & Ours      & 0.00                          & 0.00                          & 81.49                       & 53.11                         & \textbf{33.65}           \\ \midrule
    NOPS      & 25.54                         & 0.00                          & 68.15                       & 34.12                         & 31.95                    \\
    Ours      & 64.05                         & 0.00                          & 82.22                       & 67.63                         & \textbf{53.47}           \\ \bottomrule
    % \multirow{2}{*}{K=5}  & NOPS      & 23.21                         & 10.78                         & 77.60                       & 32.84                         & 36.11                    \\
    %                  & Ours      & 66.04                         & 11.81                         & 67.99                       & 53.78                         & \textbf{49.91}           \\ \midrule
    % \multirow{2}{*}{K=6}  & NOPS      & 24.75                         & 9.57                          & 52.61                       & 29.22                         & 29.04                    \\
    %                                       & Ours      & 42.10                         & 10.30                         & 54.87                       & 44.01                         & \textbf{37.82}           \\ \bottomrule
    \end{tabular}}
    
\end{table}

\paragraph{Adaptive Regularization and Hyperparameters Selection.}

\begin{figure}[!t]
  \centering
\begin{minipage}{0.3\linewidth}
  \centering
 \includegraphics[scale=0.23]{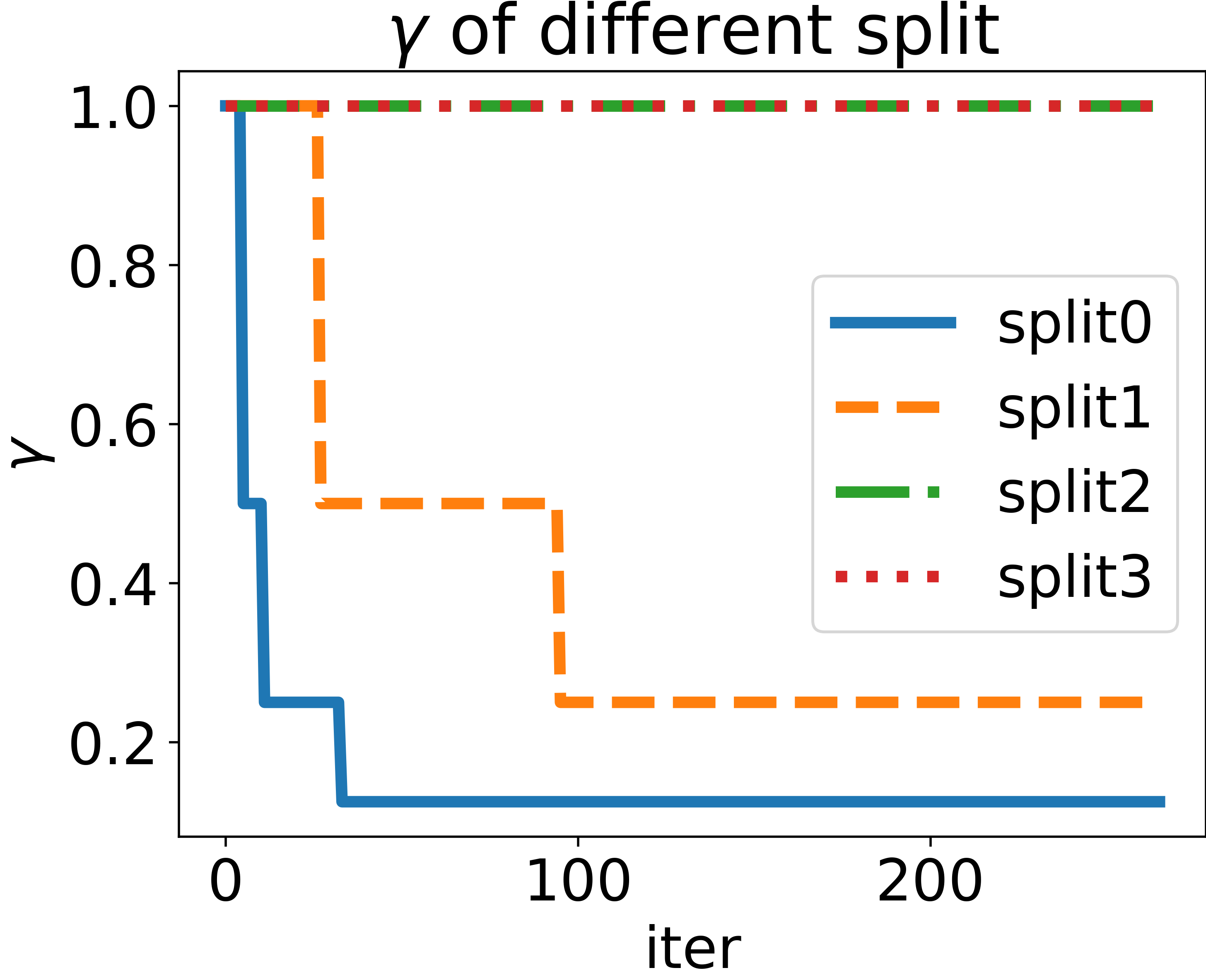}
 \caption{$\gamma$ variation}
 \label{fig:g}
 \end{minipage}
 \begin{minipage}{0.3\linewidth}
 \centering
 \includegraphics[scale=0.23]{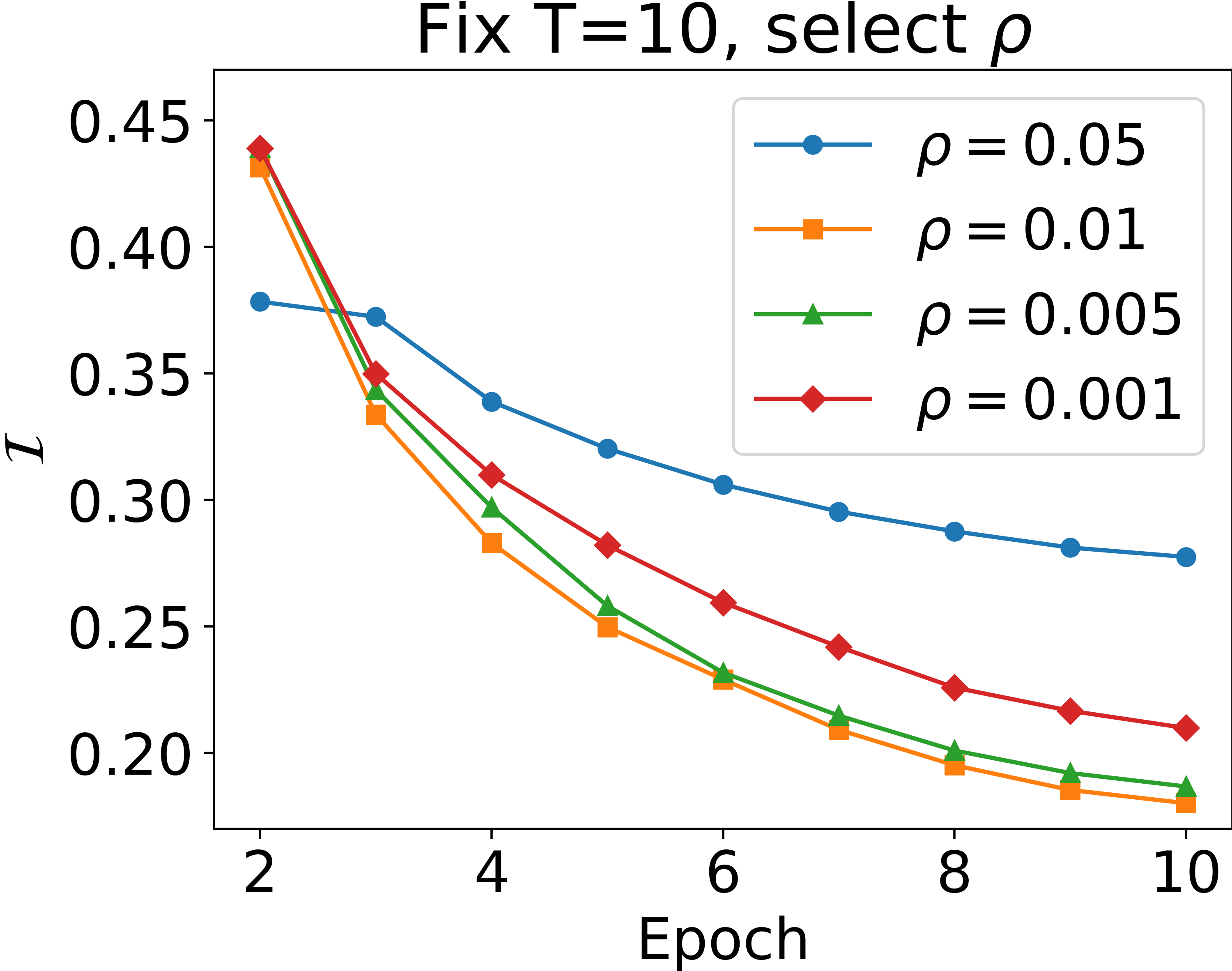}
 \caption{Selecting $\rho$}
   \label{fig:tau}
 \end{minipage}%
 \begin{minipage}{0.3\linewidth}
 \centering
 \includegraphics[scale=0.23]{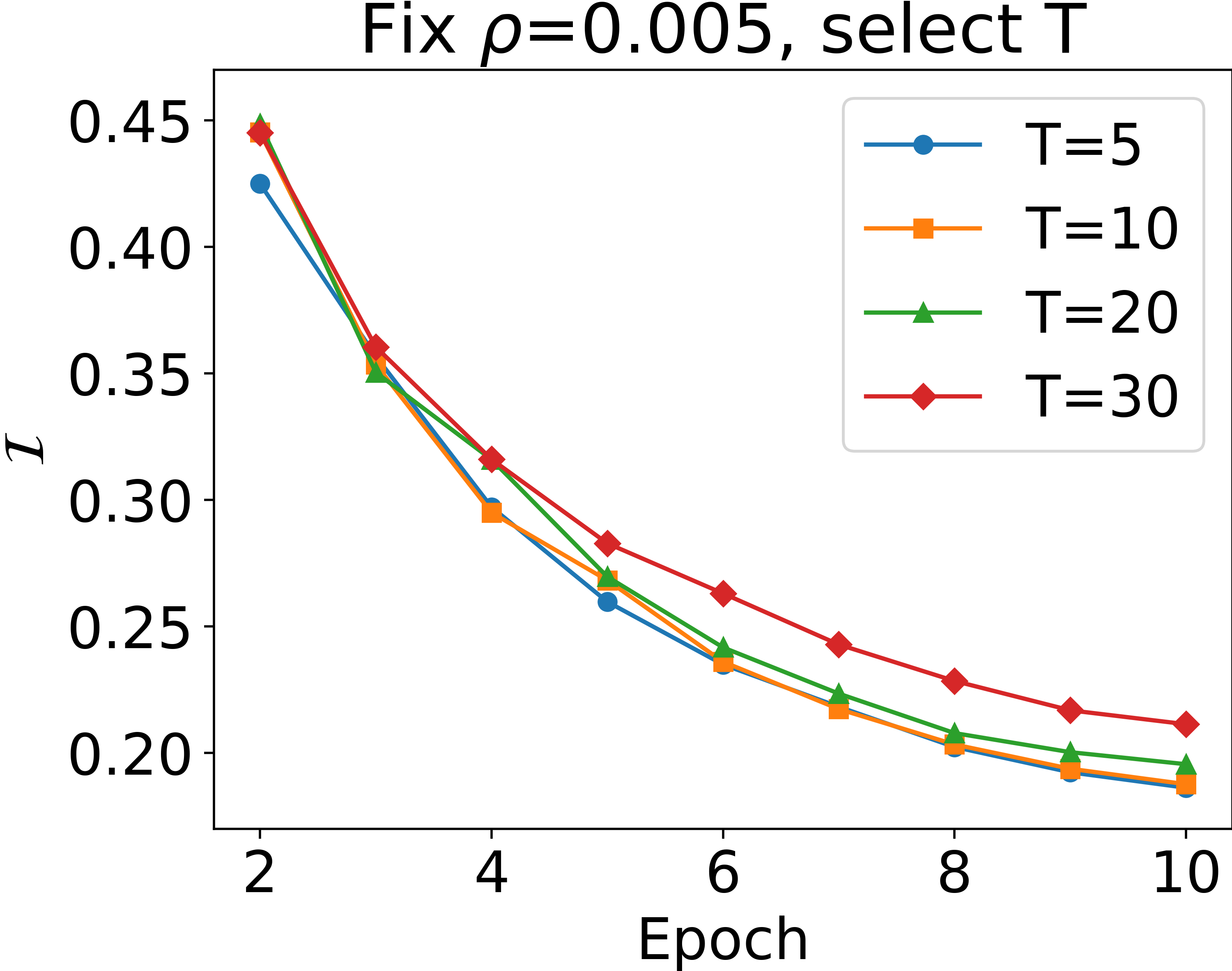}
 \caption{Selecting $T$ }
 \label{fig:T}
 \end{minipage}
  
 \end{figure}

% 展示w和不同gamma的关系图，说明adaptive的必要性。模型前期需要一个略大的gamma，随着学习的进行，gamma需要减小，而不是一个定值。
To analyze the impact of adaptive regularization, we compare it with various fixed regularization factors, as illustrated in \cref{tab:gamma_analysis}. We notice that employing a very small fixed $\gamma$, such as 0.05 as indicated in the table, results in a weak prior constraint, and the model tends to learn a degenerate solution where all samples are assigned to a single cluster. When the $\gamma$ increases to 0.5, the model achieves optimal results, but the increment decreases when the $\gamma$ further increases. Compared with adaptive $\gamma$, the optimal results of fixed $\gamma$ is nearly \textbf{8.2\%} lower, demonstrating that the adoption of an adaptive $\gamma$ not only enhances the model's flexibility but also prevents any performance degradation. Furthermore, we experiment with the setup adopting the GT class distribution and substituting the KL constraint in \cref{eq:imbalance_sl} with an equality constraint. Surprisingly, the results indicate that the GT class distribution constraint is not the optimal solution for clustering imbalanced novel classes. 
At last, in \cref{fig:g}, we visualize the $\gamma$ curves for SemanticPOSS in four splits. Split 0 exhibits the highest rate of change, followed by Split 1, while Splits 2 and 3 remain constant, indicating that our strategy is adaptive to each dataset.
% This phenomenon potentially due to the uniform constraint smoothing the pseudo-labels of head class samples. This smoothing process helps alleviate biased learning of head classes, ultimately leading to an improvement in the performance of medium and tail classes.
%However, when a larger fixed gamma is used, the results are also not optimal (the best result is nearly 8\% lower than adaptive). This is because a strong prior constraint forces the model to generate uniform pseudo labels, limiting its ability to learn the data distribution. We believe that in the later stages of training, a strong prior constraint like before is unnecessary. By utilizing an adaptive gamma, the model's flexibility can be increased while avoiding degradation. 
%Simultaneously, we also noticed that using ground truth (GT) to constrain the distribution led to worse results. We analyze that this could be attributed to the uniform constraint smoothing the pseudo-labels of head class samples. This smoothing mitigates biased learning of head classes, ultimately enhancing the performance of medium and tail classes.

To further validate the effectiveness of adjusting $\gamma$ based on KL divergence, we also compare it with typical step decay and cosine annealing strategies. For the step decay, we set the initial $\gamma$ to 1 and decay it by multiplying it with $\lambda$ every epoch. For the cosine annealing approach, we also set the initial $\gamma$ to 1 and reduce it to the minimum value (min $\gamma$). From the \cref{tab:step} and \cref{tab:cos}, we observe that the results of simple step decay and cosine annealing are nearly \textbf{10\%} worse than adaptive $\gamma$ (which is 44.2). We believe that these two typical strategies lack flexibility compared to adaptive $\gamma$. They might not facilitate the adaptive control of the $\gamma$ decay process based on the model learning process.
  
To choose the hyperparameters $\rho$ and $T$ according to the indicator outlined in \cref{sec:selflabeing}, we conduct experiments for various values of $\rho$ and $T$. The results are displayed in \cref{tab:tau} and \cref{tab:T}. Additionally, we plot the indicator's curve for each experiment in \cref{fig:tau} and \ref{fig:T}.  The plots reveal that when $\rho$ falls within the range of 0.01 to 0.005, and $T$ is set between 5 and 20, the indicator value remains low while achieving a high novel IoU. Those results demonstrate the efficiency of our hyperparameters selection strategy and the robustness of our method.

\begin{table}[t]
\centering
\begin{minipage}{0.5\linewidth}
\centering
\caption{Step decay results}
 
\label{tab:step}
\resizebox{0.8\linewidth}{!}{
\begin{tabular}{@{}c|ccccc@{}}
\toprule
 $\lambda$ & 0.1 & 0.3 & 0.5 & 0.7 & 0.9 \\ \midrule
Step decay & 34.0 & 34.2 & 32.9 & 34.6 & 33.3\\ \bottomrule
\end{tabular}}
\end{minipage}%
\begin{minipage}{0.5\linewidth}
\centering
\caption{Cosine annealing results}
 
\resizebox{1.0\linewidth}{!}{
\begin{tabular}{@{}c|ccccc@{}}
\toprule
\label{tab:cos}
min $\gamma$ & 0.1 & 0.05 & 0.01 & 0.005 & 0.001 \\ \midrule
Cosine annealing & 32.2 & 32.5 & 35.8 & 36.1 & 32.0 \\ \bottomrule
\end{tabular}}
\end{minipage}
 
\end{table}

 \begin{table}[t]
 \centering
 \begin{minipage}{0.5\linewidth}
 \centering
 \caption{The results for different $\rho$}
  
 \label{tab:tau}
 \resizebox{0.7\linewidth}{!}{
 \begin{tabular}{@{}c|cccc@{}}
 \toprule
 $\rho$                 & 0.05                         & 0.01                         & 0.005                        & 0.001                        \\ \midrule
 { Split0} & { 10.08} & { 45.84} & { 44.21} & { 33.11} \\ \bottomrule
 \end{tabular}}
 \end{minipage}%
 \begin{minipage}{0.5\linewidth}
 \centering
 \caption{The results for different $T$}
  
 \label{tab:T}
 \resizebox{0.7\linewidth}{!}{
 \begin{tabular}{@{}c|cccc@{}}
 \toprule
 $T$                    & 5                            & 10                           & 20                           & 30                           \\ \midrule
 { Split0} & { 44.46} & { 44.21} & { 44.12} & { 32.48} \\ \bottomrule
 \end{tabular}}
 \end{minipage}
 
 \end{table}

\paragraph{Limitations.} 
One limitation is our problem setup which follows ~\cite{riz2023novel} and only addresses scenarios where unlabelled data constitutes novel classes. In contrast, a more realistic open-world setting necessitates handling situations where both known classes and novel classes lack labels.
Nevertheless, we anticipate that our method will establish a robust baseline and stimulate further research aimed at addressing the challenges presented by practical open-world situations.

%% file: section/conclusion.tex
\section{Conclusion}
%In this paper, we propose a novel dual-level self-labelling algorithm for point clouds novel class semantic segmentation. We formulate the pseudo label generation process as a Semi-relaxed Optimal Transport problem and incorporate an adaptive regularization factor to dynamically adjust the constraint of the uniform prior, thereby generating higher-quality imbalanced pseudo labels.
%preventing undesirable solutions while enhancing the model's adaptability.
%To leverage the spatial prior, which is an important cue for point cloud understanding, we employ a DBSCAN to cluster the nearest points into regions and develop a dual-level learning strategy that discovers novel classes at the point- and region-level.  Extensive experiments on two datasets, SemanticKITTI and SemanticPOSS demonstrate the superiority of our method.
In this paper, we propose a novel dual-level adaptive self-labeling framework for novel class discovery in point cloud segmentation. Our framework formulates the pseudo label generation process as a Semi-relaxed Optimal Transport problem and incorporates a novel data-dependent adaptive regularization factor to gradually relax the constraint of the uniform prior based on the distribution of pseudo labels, thereby generating higher-quality imbalanced pseudo labels for model learning. In addition, we develop a dual-level representation that leverages the spatial prior to generate region representation, which reduces the noise in generated segmentation and enhances point-level classifier learning. Furthermore, we propose a hyperparameters search strategy based on training sets. Extensive experiments on two widely used datasets, SemanticKITTI and SemanticPOSS, demonstrate the effectiveness of each component and the superiority of our method.
%We tackle the novel class segmentation in point cloud, which leverages known class knowledge to densely discover novel classes. Existing works assume a uniform distribution on cluster size and formulate pseudo label generation process as an optimal transport problem. However, they achieve inferior performance due to the imbalanced distribution within point clouds. To mitigate this issue, we propose a novel adaptive self-labeling algorithm that adaptively generates higher-quality imbalanced pseudo-labels, improving the clustering of novel classes. Additionally,predicting new classes densely without any prior knowledge is highly challenging.
%To address this, we employ a spatial smooth prior within the point clouds to cluster the nearest points into regions and develop a dual-level learning strategy which discovery novel classes in point- and region-level. Extensive experiments onexisting benchmarks demonstrate the superiority of our method.

\subsubsection*{Acknowledgments}
This work was supported by National Science Foundation of China under grant 62350610269, Shanghai Frontiers Science Center of Human-centered Artificial Intelligence, and MoE Key Lab of Intelligent Perception and Human-Machine Collaboration (ShanghaiTech University).

%% file: section/suppl.tex
\appendix
\section{Semi-relaxed Optimal Transport} \label{appendix:srot}
In this section, we first discuss the advantages of our formulation. We then demonstrate how to solve the semi-relaxed optimal transport using an efficient scaling algorithm. Finally, we analyze the differences between our algorithm and \cite{zhang2023novel} in detail. 

\subsection{Analysis of our OT formulation}
Without loss of generality, we consider our formulation as the following generic form:
\begin{align}\label{eq:gen_imbalance_sl}
    &\min_{\mathbf Q}\langle\mathbf{Q},\mathbf C\rangle_F+\gamma KL(\mathbf{Q}^{\top} \mathbf1_M,\bm{\mu}) \\
    &\text{s.t. }\mathbf{Q} \in \{\mathbf{Q} \in \mathbb R^{M\times N}|{\mathbf{Q}}\mathbf{1}_N=\bm{\nu}\},
\end{align}
where $\bm{\mu,\nu}$ are the prior marginal distribution of $\mathbf Q$, and $\mathbf C$ is the cost matrix. 
% As analyzed in Appendix \ref{appendix:pro1}, the representation and prototype gradually bootstrap. Consequently, 
As analyzed in Appendix \ref{appendix:pro1},
we observe that the imbalanced property of the dataset is reflected in the model's prediction $P$. OT generates pseudo labels $Q$ based on $P$ and several distribution constraints. Unlike typical OT, which imposes two equality constraints and enforces a uniform distribution, our relaxed OT utilizes a relaxed KL constraint on cluster size.
In optimizing our relaxed OT, the optimal $Q$ for $<Q, -\log P>$ is assigned based on the largest prediction in $P$, capturing its imbalanced property. The KL constraints ensure that the marginal distribution of $Q$ remains close to the prior uniform distribution, avoiding degenerate solutions while providing flexibility. Consequently, the optimal $Q$ in our relaxed OT accounts for both the inherent imbalanced distribution of classes in $P$ and the prior uniform distribution, generating pseudo labels that reflect the imbalanced characteristics of $P$.

\subsection{Efficient Solver}
While the above formulation suits our problem, its quadratic time complexity is unaffordable for large-scale problems. To solve this efficiently, motivated by \cite{cuturi2013sinkhorn}, we first introduce an entropic constraint, $-\epsilon \mathcal{H}(\mathbf{Q})$. Due to  
\begin{equation}
    \langle\mathbf{Q},\mathbf C\rangle_F - \epsilon \mathcal{H}(\mathbf{Q}) = \epsilon \langle\mathbf{Q}, \mathbf{C}/\epsilon + \log \mathbf{Q}\rangle_F = \epsilon \langle\mathbf{Q}, \log \frac{\mathbf{Q}}{e^{-\mathbf{C}/\epsilon}}\rangle_F,
\end{equation}
The entropic semi-relaxed optimal transport can be reformulated as:
\begin{align}\label{eq:gen_imbalance_sl_en}
    &\epsilon \langle\mathbf{Q}, \log \frac{\mathbf{Q}}{e^{-\mathbf{C}/\epsilon}}\rangle_F+\gamma KL(\mathbf{Q}^{\top} \mathbf1_M,\bm{\mu}) \\
    &\text{s.t. }\mathbf{Q} \in \{\mathbf{Q} \in \mathbb R^{M\times N}|{\mathbf{Q}}\mathbf{1}_N=\bm{\nu}\}.
\end{align}
This problem can then be approximately solved by an efficient scaling algorithm. For more details, refer to \cite{chizat2018scaling}.

%\textcolor{red}{illustrate the algorithm proposed by Zhang et al, and point out the weaknesses and strengths of their paper.}

\subsection{Comparison with \cite{zhang2023novel}} 
Zhang et al.~\cite{zhang2023novel} introduce an imbalanced self-labeling learning framework to tackle the issue of novel class discovery in long-tailed scenarios. To generate imbalanced pseudo-labels, they introduce an auxiliary variable $\mathbf{w} \in \mathbb{R}^{N}$, which is dynamically inferred during learning and encodes constraints on the cluster-size distribution. Their formulation is as follows:
\begin{align}\label{eq:gen_imbalance_tmlr}
    &\min_{\mathbf Q}\langle\mathbf{Q},\mathbf C\rangle_F+\gamma KL(\mathbf{w},\bm{\mu}) \\
    &\text{s.t. }\mathbf{Q} \in \{\mathbf{Q} \in \mathbb R^{M\times N}|{\mathbf{Q}}\mathbf{1}_N=\bm{\nu}, 
    {\mathbf{Q}^\top}\mathbf{1}_M=\mathbf{w}
    \},
\end{align}
Unlike our approach, they adopt a fixed $\gamma$. To optimize Equ.(\ref{eq:gen_imbalance_tmlr}), they propose a bi-level optimization strategy, alternately estimating cluster distributions and generating pseudo labels by solving an optimal transport problem. Specifically, they start with a fixed $\mathbf{w}$ and first minimize Equ.(\ref{eq:gen_imbalance_tmlr}) with respect to $\mathbf{Q}$. Since the KL constraint term remains constant, this task reduces to a standard optimal transport problem, which can be efficiently solved using the Sinkhorn-Knopp Algorithm. Then, they optimize Equ.(\ref{eq:gen_imbalance_tmlr}) with respect to $\mathbf{w}$ using simple gradient descent.

%Unlike us, they adopt a fixed $\gamma$ in their approach. To optim Equ.(\ref{eq:gen_imbalance_tmlr}), they propose a bi-level optimization strategy, which alternately estimates cluster distributions and generates pseudo labels by solving an optimal transport problem. Specifically, they start from a fixed $\mathbf{w}$ and first minimize Equ.(\ref{eq:gen_imbalance_tmlr}) $w.r.t$ $\mathbf{Q}$. As the KL constraint term remains constant, the task turns into a standard optimal transport problem, which can be efficiently solved by the Sinkhorn-Knopp Algorithm. Then they optimize Equ.(\ref{eq:gen_imbalance_tmlr}) $w.r.t$ $\mathbf{w}$  with simple gradient descent. 

\begin{figure}[!t]
\footnotesize
\centering
\begin{minipage}{0.45\linewidth}
\centering
\includegraphics[scale=0.35]{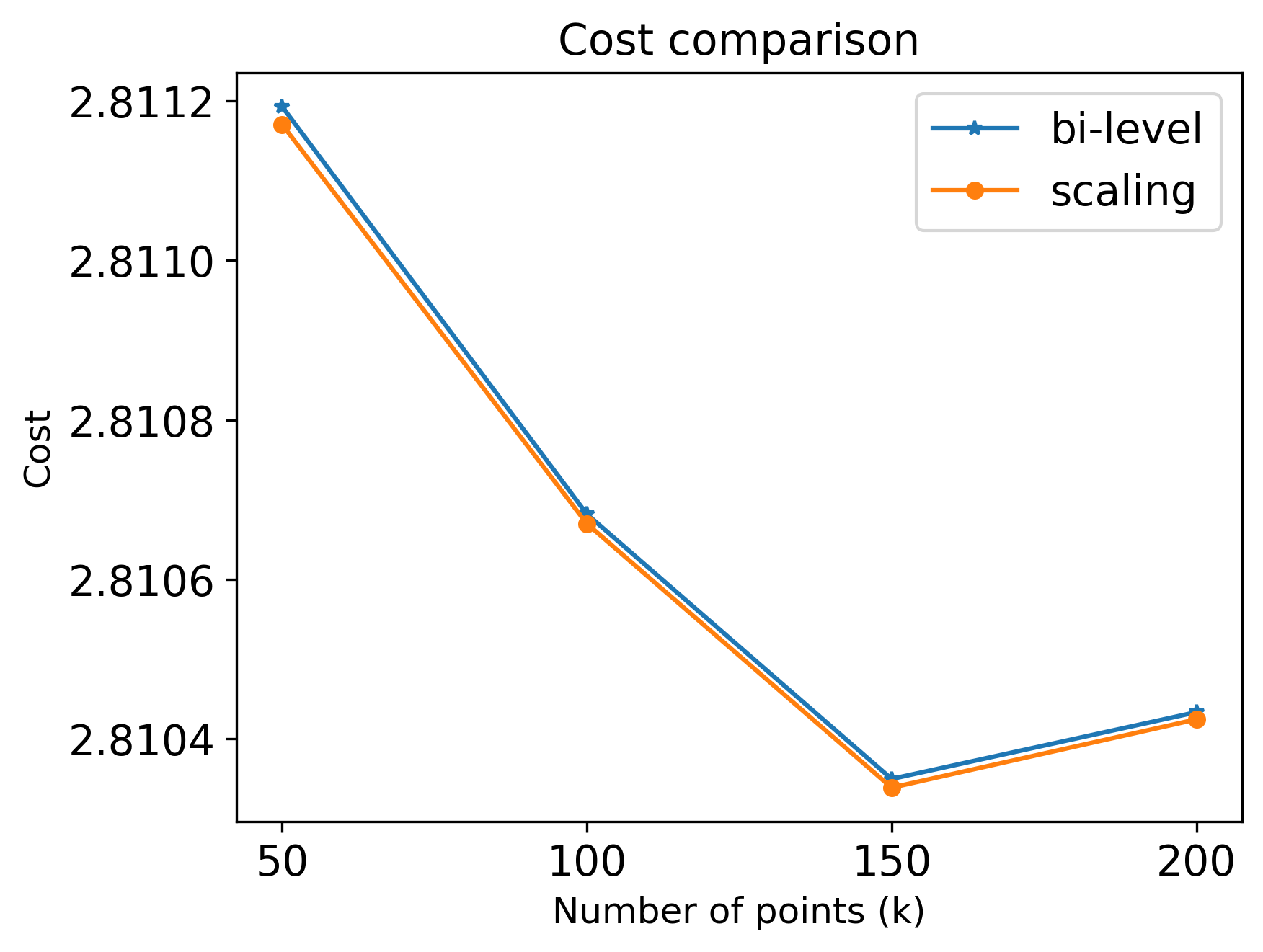}
% \caption{Selecting $\rho$ based on the indicator}
\label{fig:sink}
\end{minipage}%
\begin{minipage}{0.45\linewidth}
\centering
\includegraphics[scale=0.35]{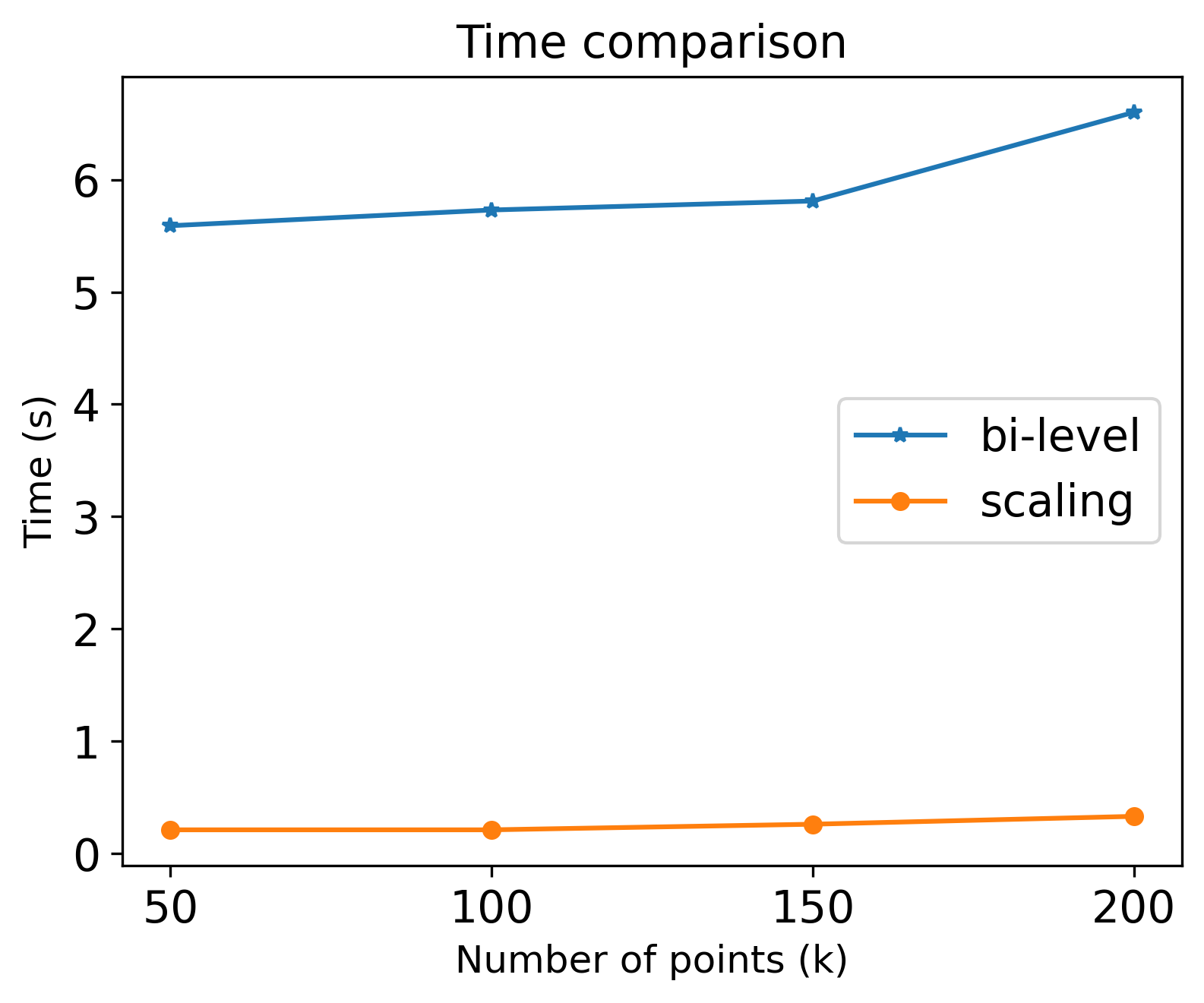}
\end{minipage}
\caption{Comparison of two optimization algorithms.}
\label{fig:scaling_bilevel}
\end{figure}

While their bi-level optimization approximates the objective function, it consumes significantly more time compared to the direct application of the light-speed scaling algorithm. Additionally, it introduces extra hyperparameters $\mathbf{w}$ for inner-loop optimization.
As shown in \cref{fig:scaling_bilevel}, the scaling algorithm is faster compared to the bi-level optimization strategy proposed by \cite{zhang2023novel}, making it feasible to solve large-scale problems.

\section{Pseudo Code of our method}\label{append:pc}
We provide a detailed description of our method in Alg.\ref{alg:PICA} for clarity and have released our code in \href{https://github.com/RikkiXu/NCD_PC}{https://github.com/RikkiXu/NCD\_PC}.

\usepackage[utf8]{inputenc}
\usepackage{amsmath}
\usepackage{algorithm2e}

\SetKwComment{Comment}{//}{}
\begin{algorithm}[t!]
\caption{Dual-level Imbalanced-aware Self-labeling and Learning Algorithm}\label{alg:PICA}
\SetAlgoLined
\footnotesize
\DontPrintSemicolon
\KwIn{Trainset $\mathcal{D}$, softmax function $\sigma$, encoder $f_{\theta}$,  classifier $h = [h^s, h^u] \in \mathbb R^{D\times(|C^s|+|C^u|)}$, $\bm{\mu} = \mathbf{1}$, uniform distribution $\bm{\nu}$, hyperparameters $\gamma_p, \gamma_r,\lambda,\rho,T,\alpha, \beta, \epsilon$}    
 %\bm{\nu} = \mathbf{1}_{|C^u|\times 1}

\For{$ e \in 1, 2,.., Epoch $}{
\For{$ s \in 1, 2, ..., Step $}{

$\{(x^s_i, y^s_i)\}^N_{i=0}, \{x^u_j\}^M_{j=0} \leftarrow \text{Sample}(\mathcal{D})$

\Comment{Cluster point into different regions by DBSCAN}
$\{r_k\}^K_{k=0} \leftarrow DBSCAN(\{x^u_j\}^M_{j=0})$

\Comment{Forward the model}
$p^s = \softmax (h^s \circ f_\theta(x^s) / \tau)$

$\mathbf{z}_p = f_{\theta}(x^u),\mathbf{z}_r =\{\text{AvgPool}(\mathbf{z}_p)|r_k \text{ is same}\}$ 

$p_p^u = \softmax (h^u \circ \mathbf{z}_p / \tau), p_r^u = \softmax (h^u \circ \mathbf{z}_r / \tau)$

\Comment{CE loss for known classes}
$\mathcal{L}_s=-\frac{1}{N}\sum^N_{i=1}y^s_i\log p_i^s$

\Comment{Point level self-labeling}
$\mathbf{Q}^u_p$ = \SelfLabeling{ $-\log \mathbf{P}_p^u$, $\gamma^p_t$, $\epsilon$}

$\mathcal{L}^u_p=\frac{1}{M}\langle \mathbf{Q}^u_p, -\log\mathbf{P}_r^u\rangle_F$

\Comment{Region level self-labeling}
$\mathbf{Q}^u_r$ = \SelfLabeling{ $-\log \mathbf{P}_r^u$, $\gamma^r_t$, $\epsilon$ }

$\mathcal{L}^r_u=\frac{1}{K}\langle \mathbf{Q}^u_r, -\log\mathbf{P}_r^u\rangle_F$

\Comment{Update $\gamma^p_{t+1}$ and $\gamma^r_{t+1}$}
\If{$KL(\frac{1}{M}{\mathbf{Q}^u_p}^\top\mathbf{1}_M, \bm\nu)$ \text{ $\leq \rho$ consistently for $T$ iterations}}
    {$\gamma^p_{i+1}=\lambda\gamma^p_{i}$}

\If{$KL(\frac{1}{K}{\mathbf{Q}^u_r}^\top\mathbf{1}_M, \bm\nu)$ $\leq \rho$ consistently for $T$ iterations}
    {$\gamma^r_{i+1}=\lambda\gamma^r_{i}$}

\Comment{Total loss}
minimize $\mathcal{L}_s + \alpha \mathcal{L}_u^p + \beta \mathcal{L}^r_u\ w.r.t \ \theta$
}
}
\end{algorithm}

\section{Dataset Splits} \label{appendix:dataset}  
\begin{figure}
    \begin{minipage}[b]{0.35\textwidth}
        \centering
        \includegraphics[width=0.85\linewidth]{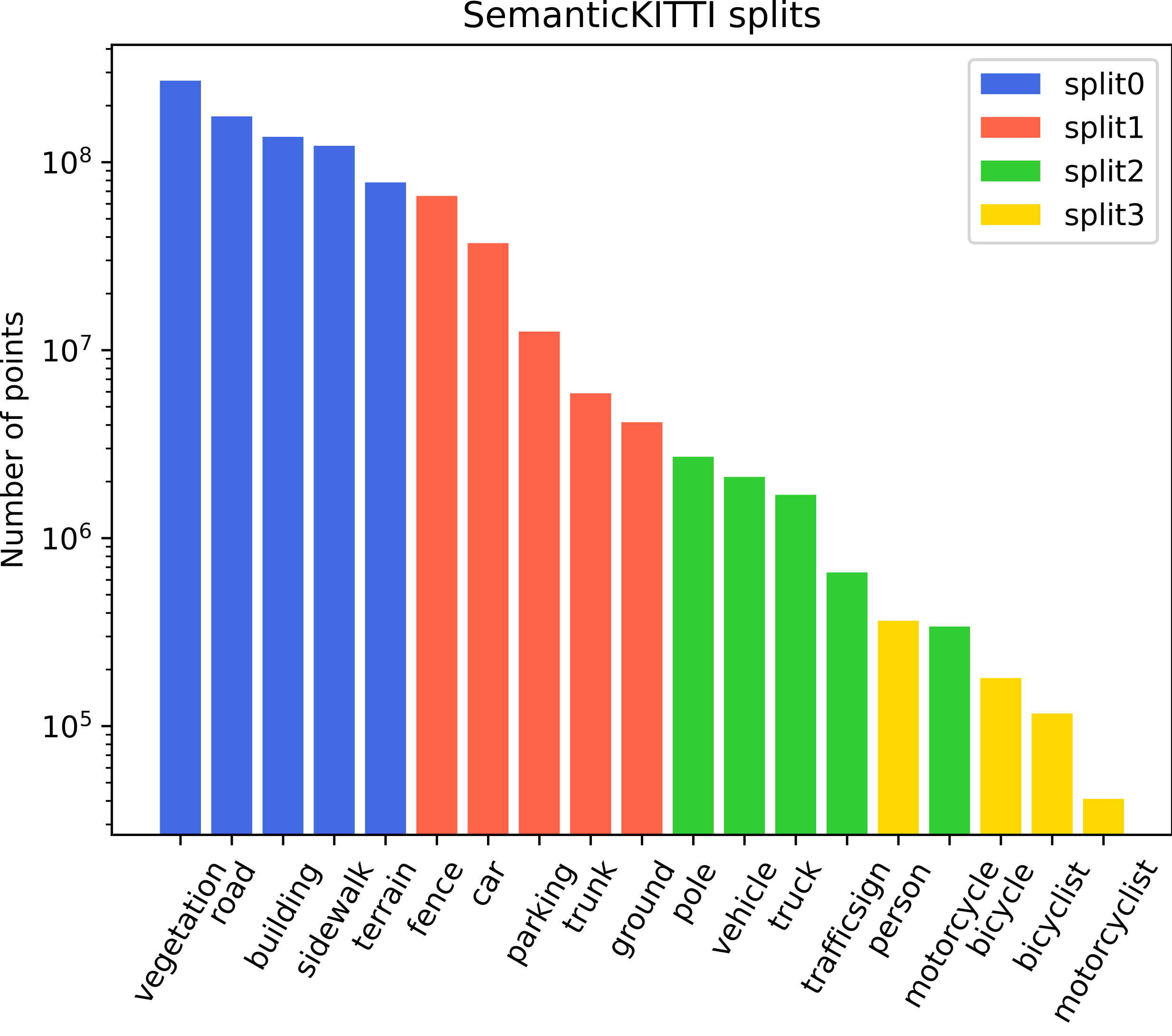}
    \end{minipage}%
    \begin{minipage}[b]{0.35\textwidth}
        \centering
        \includegraphics[width=0.85\linewidth]{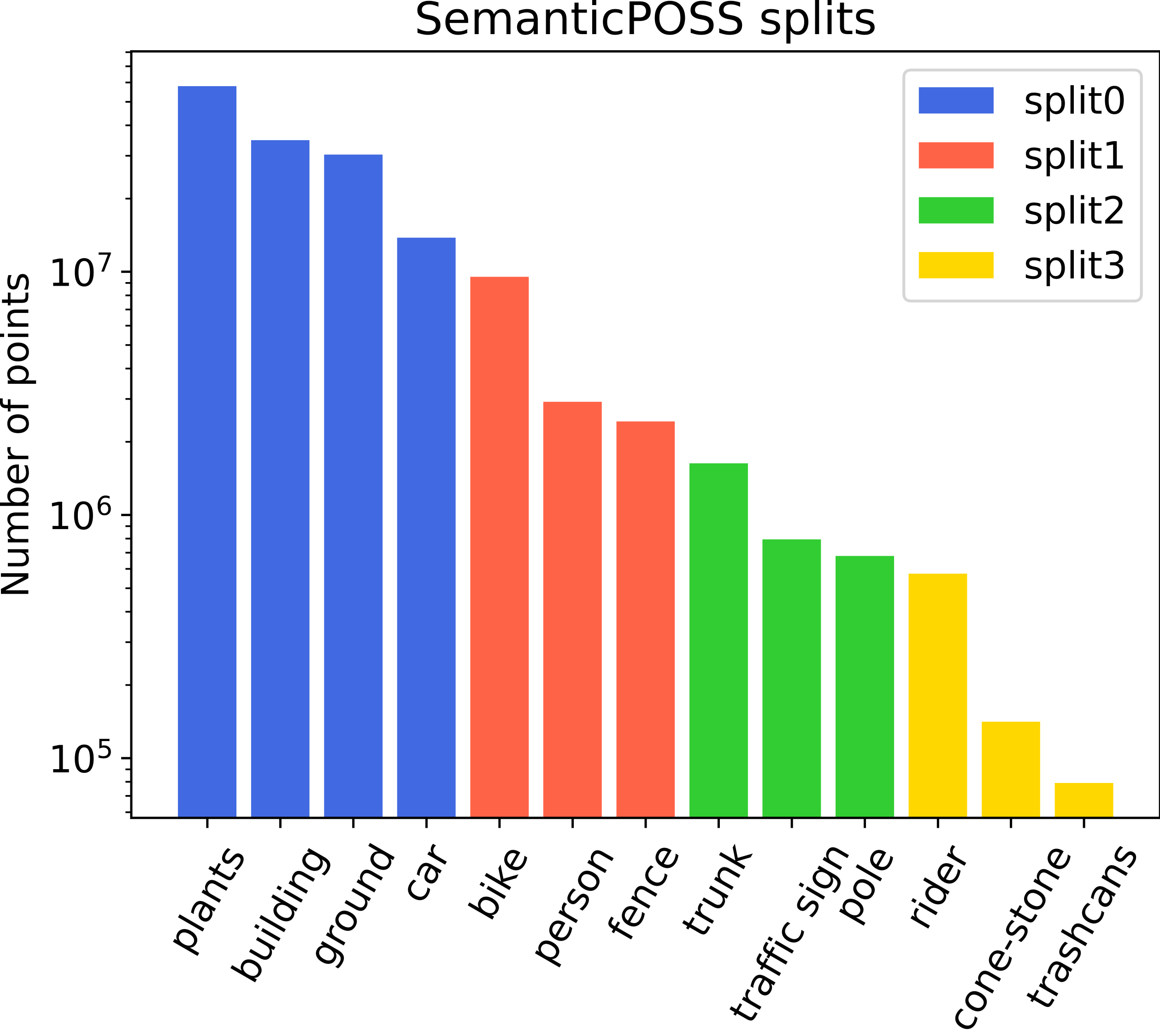}
    \end{minipage}%
    \begin{minipage}[b]{0.35\textwidth}
        \centering
        \includegraphics[width=0.85\linewidth]{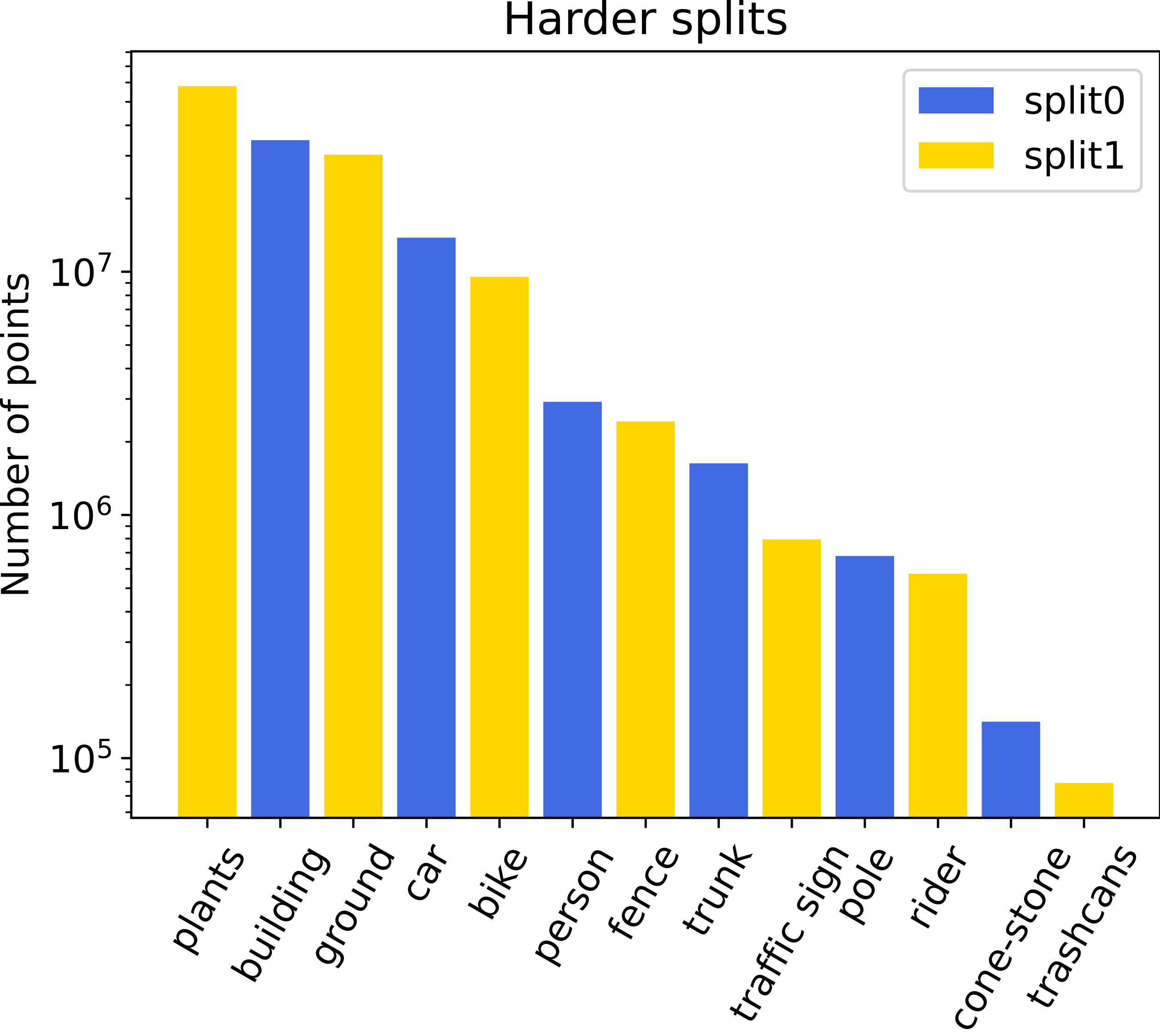}
    \end{minipage}
    \caption{Distribution plot of the SemanticPOSS dataset. Each class has been assigned the color of the split which has to be considered novel.}
    \label{fig:data_bar}
\end{figure}

We follow~\cite{riz2023novel} and partition SemanticKITTI and SemanticPOSS into four splits, detailed in \cref{tab:dataset} and \ref{tab:dataset2}. It's important to note that~\cite{riz2023novel} balance the distribution of novel classes across splits to prevent the most frequent novel class from biasing other classes and to leverage semantic relationships between known and novel classes. The left and middle plots in \cref{fig:data_bar} illustrate the distributions of SemanticKITTI and SemanticPOSS across these splits. However, this deliberate selection of novel classes may not adequately address point cloud data imbalance.

To assess the generalization of our algorithm, we conduct experiments on a more challenging benchmark. Here, we select half of the classes from SemanticPOSS as novel classes, as depicted on the right side of \cref{fig:data_bar}. Results and analysis are presented in Section 4.2.

%It is worth noting that, to avoid the most frequent novel class affecting the other classes and to better utilize the semantic relationships between known and novel classes, they balance the distribution of the novel classes in each split. \cref{fig:data_bar}'s left and middle plots illustrate the distribution of SemanticKITTI and SemanticPOSS in each split. However, deliberately selecting novel classes in this manner is not reasonable as it avoids addressing the issue of point cloud data imbalance. To validate the generalization of our algorithm, we conduct experiments on a more challenging benchmark. In this benchmark, we cross-select half of the classes from SemanticPOSS as novel classes, as shown in the right side of \cref{fig:data_bar}. We analyze the results in Sec.4.2.

% Compared with NOPS, our method achieves improvements in head(7.5\%), medium(8.9\%), and tail(6.8\%) classes. 
% On average, we achieve an IoU of 30.2\% for novel classes across all four splits, outperforming NOPS (22.5\%) by \textbf{7.7\%}. Details of dataset distribution are in Appendix C.
%\ref{appendix:dataset}. 
% Note that the improvement in known classes can be attributed to the training of NOPS not converging. 

% In the Appendix E
% %\ref{appendix:nops*}
% , we compare our method and NOPS*, which training is converging. The results show that, compared with NOPS, NOPS* achieves sizeable improvements in known classes but drops a lot in novel classes. Therefore, our method still outperforms NOPS* by a sizeable margin.
%ours 44.98 30.8  11.2
%nops 37.46 21.925 4.4

\begin{table}[!t]
\centering
\caption{The detail of novel classes in each split. }
\label{tab:dataset}
\begin{tabular}{c|cccccc|ccccc}
\toprule
Split & \multicolumn{6}{c|}{SemanticKITTI}                                    & \multicolumn{5}{c}{SemanticPOSS}               \\ \midrule
0     & \multicolumn{6}{c|}{building,road,sidewalk,terrain,vegetation}        & \multicolumn{5}{c}{building,car,ground,plants} \\
1     & \multicolumn{6}{c|}{car,fence,other-ground,parking,trunk}             & \multicolumn{5}{c}{bike,fence,person}          \\
2     & \multicolumn{6}{c|}{motorcycle,other-vehicle,pole,traffic-sign,truck} & \multicolumn{5}{c}{pole,traffic-sign,trunk}    \\
3     & \multicolumn{6}{c|}{bicycle,bicyclist,motorcyclist,person}            & \multicolumn{5}{c}{cone-stone,rider,trashcan} \\ \bottomrule
\end{tabular}
\end{table}

\begin{table}[!t]
\centering
\caption{The detail of novel classes on the challenging setting of SemanticPOSS dataset.}
\label{tab:dataset2}
\begin{tabular}{c|ccccc} 
\toprule
Split & \multicolumn{5}{c}{SemanticPOSS}                                      \\ \midrule
0     & \multicolumn{5}{c}{building,car,ground,plants,bike,fence,person}      \\
1     & \multicolumn{5}{c}{pole,traffic-sign,trunk,cone-stone,rider,trashcan} \\ \bottomrule
\end{tabular}
\end{table}
\section{Analysis on Detailed Results}\label{appendix:anaysis}

\begin{table}[h]
\begin{minipage}{0.45\linewidth}
\centering
\caption{Detailed results for SemanticPOSS}
\resizebox{0.7\linewidth}{!}{%
\begin{tabular}{@{}l|ccc@{}}
\toprule
Method & {\color[HTML]{1F1F1F} Head}  & {\color[HTML]{1F1F1F} Medium} & {\color[HTML]{1F1F1F} Tail} \\ \midrule
NOPS   & {\color[HTML]{1F1F1F} 37.5} & {\color[HTML]{1F1F1F} 21.9}   & {\color[HTML]{1F1F1F} 4.4} \\
Ours   & {\color[HTML]{1F1F1F} 45.0} & {\color[HTML]{1F1F1F} 30.8}  & {\color[HTML]{1F1F1F} 11.2}  \\ \bottomrule
\end{tabular}
}
\label{tab:results_a}
\end{minipage}%
\hfill
\begin{minipage}{0.45\linewidth}
\centering
\caption{Detailed results for SemanticKITTI}
\resizebox{0.7\linewidth}{!}{%
\begin{tabular}{@{}l|ccc@{}}
\toprule
Method & {\color[HTML]{1F1F1F} Head}  & {\color[HTML]{1F1F1F} Medium} & {\color[HTML]{1F1F1F} Tail} \\ \midrule
NOPS   & {\color[HTML]{1F1F1F} 26.1} & {\color[HTML]{1F1F1F} 28.3}   & {\color[HTML]{1F1F1F} 7.4} \\
Ours   & {\color[HTML]{1F1F1F} 33.8} & {\color[HTML]{1F1F1F} 32.2}  & {\color[HTML]{1F1F1F} 8.5}  \\ \bottomrule
\end{tabular}
}
\label{tab:results_b}
\end{minipage}
\end{table}

To validate our method's ability to handle imbalanced data distributions, we compute the mIoU for head classes, medium classes, and tail classes within each split. The results for both datasets are presented in \cref{tab:results_a} and \cref{tab:results_b}.
Specifically, for SemanticPOSS, in the first split, there are a total of 4 novel classes. We choose the two classes with the highest count as head classes, and the remaining two are designated as medium and tail classes. In the other splits, which all have 3 novel classes, we sort them by size and assign them as head, medium, and tail classes accordingly. 
For SemanticKITTI, we designated the largest class and the smallest class within each split as head and tail classes, respectively, with the rest categorized as medium classes.

%To validate that our method can handle imbalanced data distributions, we compute the mIoU for head classes, medium classes, and tail classes within each split. The results for both datasets are shown in \cref{tab:results_a} and \cref{tab:results_b}.
%Specifically, for SemanticPOSS, in the first split, there are a total of 4 novel classes. We choose the two classes with the highest count as head classes, and the remaining two are designated as medium and tail classes. In the other splits, which all have 3 novel classes, we sort them by size and assign them as head, medium, and tail classes accordingly. 
%For SemanticKITTI, we designated the largest class and the smallest class within each split as head and tail classes, respectively, with the rest categorized as medium classes.

As shown in \cref{tab:results_a},
in SemanticPOSS, our approach achieve improvements of 7.5\%, 8.9\%, and 6.8\% for head, medium, and tail classes, respectively, compared to NOPS. Similarly, in SemanticKITTI, our method yielded improvements of 7.7\%, 3.9\%, and 1.1\% for head, medium, and tail classes, respectively. These results demonstrate that our method effectively addresses imbalanced data scenarios by improving performance across all class categories. It's worth noting that NOPS utilizes additional techniques such as multi-head and overclustering during training to enhance its performance, whereas our method achieves these improvements without employing such techniques, highlighting its efficiency and effectiveness.
%30.65 28.7
%42.975 27.55
%15.185, 14.94

\section{Comparison with NOPS variant}\label{appendix:nops*}
We observed that NOPS~\cite{riz2023novel} employs a learning rate of 0.01 with SGD, which proved excessively low and caused training not to converge. To ensure convergence, we modified the optimization strategy by switching to the AdamW optimizer with an initial learning rate of 1e-3, gradually decreasing to 1e-5 following a cosine schedule. This adjustment successfully led to the convergence of NOPS during training. The results, as shown in \cref{tab:imporved_nops_semantic_poss} and \cref{tab:suppl_semantic_kitti}, indicate that while the mIoU for known classes improves, there is a decrease in mIoU for novel classes as a consequence. Nevertheless, our method continues to outperform NOPS by a significant margin.
\begin{table}[!t]
\centering
\scriptsize
\renewcommand\arraystretch{1.0}
\setlength\tabcolsep{0.5pt}
\caption{NOPS* denotes NOPS with our training setting.}

\label{tab:imporved_nops_semantic_poss}
\begin{tabular}{cc|ccccccccccccc|ccc}
\toprule
\multicolumn{1}{c}{\rotatebox{45}{Split}} & \multicolumn{1}{c|}{\rotatebox{45}{Method}} & \multicolumn{1}{c}{\rotatebox{45}{bike}} & \multicolumn{1}{c}{\rotatebox{45}{build.}} & \multicolumn{1}{c}{\rotatebox{45}{car}} & \multicolumn{1}{c}{\rotatebox{45}{cone.}} & \multicolumn{1}{c}{\rotatebox{45}{fence}} & \multicolumn{1}{c}{\rotatebox{45}{grou.}} & \multicolumn{1}{c}{\rotatebox{45}{pers.}} & \multicolumn{1}{c}{\rotatebox{45}{plants}} & \multicolumn{1}{c}{\rotatebox{45}{pole}} & \multicolumn{1}{c}{\rotatebox{45}{rider}} & \multicolumn{1}{c}{\rotatebox{45}{traf.}} & \multicolumn{1}{c}{\rotatebox{45}{trashc.}} & \multicolumn{1}{c|}{\rotatebox{45}{trunk}} & \multicolumn{1}{c}{\rotatebox{45}{Novel}} & \multicolumn{1}{c}{\rotatebox{45}{Known}} & \multicolumn{1}{c}{\rotatebox{45}{All}} \\ \midrule
& Full&45.0 & 83.3& 52.0&36.5& 46.7&77.6 & 68.2& 77.7&36.0&58.9&30.3 &4.2 &14.4&- &-&48.5\\ \midrule
\multirow{2}{*}{0} & NOPS& 35.5& \cellcolor{gray!30}{30.4} & \cellcolor{gray!30}{1.2}& 13.5 & 24.1& \cellcolor{gray!30}{69.1} & 44.7 & \cellcolor{gray!30}{\textbf{42.1}} & 19.2& 47.7 & 24.4 & 8.2 & 21.8 & \cellcolor{gray!30}{35.7}& 26.6 & 29.4\\
 & NOPS*& 46.9& \cellcolor{gray!30}{16.1} & \cellcolor{gray!30}{4.2}& 35.4 & 47.8 & \cellcolor{gray!30}{54.9} & 67.1 & \cellcolor{gray!30}{37.9} & 36.1 & 62.3 & 28.9 & 1.8 & 20.2 & \cellcolor{gray!30}{28.3}& 38.1 & 35.3\\
& Ours & 46.5 & \cellcolor{gray!30}{\textbf{70.3}}& \cellcolor{gray!30}{\textbf{7.6}}& 31.1& 49.4& \cellcolor{gray!30}{\textbf{82.2}}& 67.1 & \cellcolor{gray!30}{41.7} & 37.5 & 57.5& 29.8 & 8.4 & 14.1 & \cellcolor{gray!30}{\textbf{50.4}} & 37.9 & 41.8 \\ \midrule
\multirow{2}{*}{1} & NOPS & \cellcolor{gray!30}{29.4} & 71.4 & 28.7 & 12.2& \cellcolor{gray!30}{3.9}& 78.2 & \cellcolor{gray!30}{\textbf{56.8}} & 74.2 & 18.3 & 38.9 & 23.3 & 13.7 & 23.5 & \cellcolor{gray!30}{30.0} & 38.2& 36.4 \\
 & NOPS* & \cellcolor{gray!30}{21.5} & 83.2 & 49.9 & 29.7 & \cellcolor{gray!30}{20.0} & 77.7 & \cellcolor{gray!30}{29.6} & 77.3 & 37.4 & 58.0 & 26.2 & 3.4 & 15.5 & \cellcolor{gray!30}{23.7} & 45.8 & 40.7 \\
& Ours & \cellcolor{gray!30}{\textbf{31.5}}& 83.2 & 48.7 & 25.4& \cellcolor{gray!30}{\textbf{23.9}} & 77.3 & \cellcolor{gray!30}{53.1} & 77.1& 32.5 & 57.3& 35.0 & 9.3 & 18.0 & \cellcolor{gray!30}{\textbf{36.2}} & 46.4& 44.0 \\ \midrule
\multirow{2}{*}{2} & NOPS & 37.2 & 71.8 & 29.7& 14.6& 28.4& 77.5& 52.1& 73.0 & \cellcolor{gray!30}{11.5} & 47.1& \cellcolor{gray!30}{0.5} & 10.2& \cellcolor{gray!30}{\textbf{14.8}}& \cellcolor{gray!30}{9.0}& 44.2 & 36.0 \\
& NOPS* & 44.5 & 83.0 & 51.4& 25.2& 47.5& 77.0& 66.0& 76.4 & \cellcolor{gray!30}{\textbf{23.0}} & 59.9& \cellcolor{gray!30}{4.2} & 8.4& \cellcolor{gray!30}{4.4}& \cellcolor{gray!30}{10.5}& 53.9 & 43.9 \\
 & Ours& 45.3 & 82.8 & 49.8& 28.4& 46.3& 76.7& 66.2& 77.2 & \cellcolor{gray!30}{10.9} & 58.4& \cellcolor{gray!30}{\textbf{18.6}}& 7.3 & \cellcolor{gray!30}{8.2} & \cellcolor{gray!30}{\textbf{12.6}} & 53.8 & 44.3 \\ \midrule
\multirow{2}{*}{3} & NOPS& 38.6 & 70.4 & 30.9& \cellcolor{gray!30}{0.0} & 29.4& 76.5& 56.0 & 71.8 & 17.0 & \cellcolor{gray!30}{31.9}& 26.2& \cellcolor{gray!30}{1.0} & 22.6& \cellcolor{gray!30}{10.9} & 43.9 & 36.3 \\
 & NOPS*& 44.4 & 83.1 & 46.9 & \cellcolor{gray!30}{\textbf{0.2}} & 43.0& 77.9& 65.2 & 78.0 & 34.8 & \cellcolor{gray!30}{45.3}& 32.4& \cellcolor{gray!30}{1.9} & 14.5& \cellcolor{gray!30}{15.8} & 52.0 & 43.7 \\
 & Ours& 45.5 & 82.9 & 47.7& \cellcolor{gray!30}{{0.0}} & 45.1& 77.8& 66.3& 77.7 & 34.3 & \cellcolor{gray!30}{\textbf{49.1}}& 35.6& \cellcolor{gray!30}{\textbf{4.0}} & 15.3& \cellcolor{gray!30}{\textbf{17.7}} & 52.8 & 44.7\\ \bottomrule
\end{tabular}
\end{table}

\begin{table}[!t]
    \centering
    
    \scriptsize
    \renewcommand\arraystretch{1.0}
    \setlength\tabcolsep{0.3pt}
    \caption{The novel class discovery results on the SemanticKITTI dataset. `Full' denotes the results obtained by supervised learning. The gray values are the novel classes in each split.}
    \label{tab:suppl_semantic_kitti}
    \begin{adjustbox}{width=\textwidth}
    \begin{tabular}{c|ccccccccccccccccccc|ccc}
    \toprule
    \rotatebox{45}{Method} & \rotatebox{45}{bi.cle} & \rotatebox{45}{b.clst}  & \rotatebox{45}{build.}  & \rotatebox{45}{car}     & \rotatebox{45}{fence}   & \rotatebox{45}{mt.cle}  & \rotatebox{45}{m.clst}  & \rotatebox{45}{oth-g.} & \rotatebox{45}{oth-v.}  & \rotatebox{45}{park.}   & \rotatebox{45}{pers.}   & \rotatebox{45}{pole}    & \rotatebox{45}{road}    & \rotatebox{45}{side2.}  & \rotatebox{45}{terra.}  & \rotatebox{45}{traff.}  & \rotatebox{45}{truck}   & \rotatebox{45}{trunk}   & \rotatebox{45}{veget.}  & \rotatebox{45}{Novel}   & \rotatebox{45}{Known} & \rotatebox{45}{All}  \\ \midrule
    % Number   & $1.8e^5$    &  $1.2e^5$   & $1.4e^8$   & $3.7e^7$ & $6.6e^7$    & $3.6e^5$    & $4.1e^4$  &   $4.1e^6$    & $2.1e^6$    & $1.3e^7$    &$3.4e^5$   &$2.7e^6$ & $1.7e^8$    &  $1.2e^8$    & $7.8e^7$    & $6.6e^5$    & $1.7e^6$    & $5.9e^6$    & $2.7e^8$    & -        & - & - \\ \midrule
    Full   & 2.9    & 55.4    & 89.5    & 93.5    & 27.9    & 27.4    & 0.0     & 0.9    & 19.9    & 35.8    & 31.2    & 60.0    & 93.5    & 77.8    & 62.0    & 39.8    & 50.8    & 53.9    & 87.0    & -        & - & 47.9 \\ \midrule
    NOPS   & 5.6    & 47.8    & \cellcolor{gray!30}{52.7} & 82.6    & 13.8    & 25.6    & 1.4     & 1.7    & 14.5    & 19.8    & 25.9    & 32.1    & \cellcolor{gray!30}{\textbf{56.7}} & \cellcolor{gray!30}{8.1}  & \cellcolor{gray!30}{\textbf{23.8}} & 14.3    & 49.4    & 36.2    & \cellcolor{gray!30}{44.2} & \cellcolor{gray!30}{37.1} & 26.5  & 29.3 \\
    NOPS*   & 7.9    & 55.9    & \cellcolor{gray!30}{46.7} & 89.3    & 24.7     & 27.7    & 0.0     & 1.1    & 22.7    & 25.5    & 33.8    & 57.0    & \cellcolor{gray!30}{43.2} & \cellcolor{gray!30}{17.9} & \cellcolor{gray!30}{21.7}  & 39.3    & 61.5    & 50.8    & \cellcolor{gray!30}{23.9} & \cellcolor{gray!30}{30.7} & 35.5  & 34.2 \\
    Ours   & 5.5    & 51.1    & \cellcolor{gray!30}{\textbf{74.6}} & 92.3    & 29.8    & 22.8    & 0.0     & 0.0    & 23.3    & 24.8    & 27.7    & 59.7    & \cellcolor{gray!30}{41.4} & \cellcolor{gray!30}{\textbf{22.5}} & \cellcolor{gray!30}{23.6} & 39.3    & 43.6    & 51.1    & \cellcolor{gray!30}{\textbf{66.4}} & \cellcolor{gray!30}{\textbf{45.7}} & 33.7  & 36.8 \\ \midrule
    NOPS   & 7.4    & 51.2    & 84.5    & \cellcolor{gray!30}{50.9} & \cellcolor{gray!30}{7.3}  & 28.9    & 1.8     & \cellcolor{gray!30}{0.0} & 22.2    & \cellcolor{gray!30}{19.4} & 30.4    & 37.6    & 90.1    & 72.2    & 60.8    & 16.8    & 57.3    & \cellcolor{gray!30}{\textbf{49.3}} & 85.1    & \cellcolor{gray!30}{25.4} & 46.2  & 40.7 \\
    NOPS*   & 2.1    & 52.3    & 89.2    & \cellcolor{gray!30}{52.3} & \cellcolor{gray!30}{6.5}  & 27.1    & 0.0     & \cellcolor{gray!30}{0.0} & 18.4    & \cellcolor{gray!30}{17.5} & 33.1    & 59.3    & 90.2    & 77.2    & 61.9    & 39.9    & 53.1    & \cellcolor{gray!30}{19.5} & 86.7    & \cellcolor{gray!30}{19.2} & 49.8  & 41.9 \\ 
    Ours   & 3.7    & 57.4    & 89.2    & \cellcolor{gray!30}{\textbf{56.5}} & \cellcolor{gray!30}{\textbf{17.3}} & 20.3    & 0.0     & \cellcolor{gray!30}{0.0} & 20.0    & \cellcolor{gray!30}{\textbf{30.6}} & 34.8    & 60.6    & 93.2    & 77.6    & 62.0    & 38.7    & 56.9    & \cellcolor{gray!30}{39.2} & 86.7    & \cellcolor{gray!30}{\textbf{28.7}} & 50.1  & 44.5 \\ \midrule
    NOPS   & 6.7    & 49.2    & 86.4    & 90.8    & 23.7    & \cellcolor{gray!30}{2.7}  & 0.6     & 1.9    & \cellcolor{gray!30}{\textbf{15.5}} & 29.5    & 27.9    & \cellcolor{gray!30}{\textbf{36.4}} & 90.3    & 73.4    & 61.2    & \cellcolor{gray!30}{\textbf{17.8}} & \cellcolor{gray!30}{10.3} & 46.2    & 84.3    & \cellcolor{gray!30}{16.5} & 48.0  & 39.7 \\
    NOPS*   & 4.1    & 55.2    & 89.1    & 93.4    & 29.1    & \cellcolor{gray!30}{0.6}  & 0.0     & 0.5    & \cellcolor{gray!30}{2.8}  & 33.9    & 30.9    & \cellcolor{gray!30}{32.7} & 93.1    & 77.7    & 60.9    & \cellcolor{gray!30}{0.1}  & \cellcolor{gray!30}{32.9} & 52.2    & 86.4    & \cellcolor{gray!30}{13.8} & 50.3  & 40.8 \\
    Ours   & 3.6    & 54.2    & 88.9    & 93.3    & 28.4    & \cellcolor{gray!30}{\textbf{10.2}} & 0.0     & 0.9    & \cellcolor{gray!30}{9.6}  & 33.4    & 32.2    & \cellcolor{gray!30}{36.1} & 92.7    & 77.4    & 62.2    & \cellcolor{gray!30}{10.7}  & \cellcolor{gray!30}{\textbf{34.2}} & 51.7    & 86.9    & \cellcolor{gray!30}{\textbf{20.1}} & 50.4  & 42.5 \\ \midrule
    NOPS   & \cellcolor{gray!30}{2.3} & \cellcolor{gray!30}{27.8} & 86.0    & 89.9    & 23.1    & 24.5    & \cellcolor{gray!30}{\textbf{2.9}}  & 3.1    & 18.2    & 30.1    & \cellcolor{gray!30}{\textbf{16.3}} & 39.9    & 90.7    & 73.5    & 61.0    & 17.4    & 49.8    & 44.0    & 83.2    & \cellcolor{gray!30}{12.4} & 49.0  & 41.2 \\
    NOPS*   & \cellcolor{gray!30}{2.3} & \cellcolor{gray!30}{16.9}  & 89.5    & 93.9    & 28.2    & 27.5    & \cellcolor{gray!30}{0.0} & 0.6    & 25.3    & 34.2    & \cellcolor{gray!30}{3.1}  & 60.4    & 93.2    & 77.7    & 61.3    & 38.9    & 67.0    & 54.4    & 86.6    & \cellcolor{gray!30}{5.6}  & 55.9  & 45.3 \\
    Ours   & \cellcolor{gray!30}{\textbf{2.6}} & \cellcolor{gray!30}{\textbf{32.5}} & 88.7    & 93.3    & 28.1    & 24    & \cellcolor{gray!30}{0.1}  & 1.0    & 23.7    & 35.6    & \cellcolor{gray!30}{15.3} & 59.8    & 93.2    & 77.6    & 61.4    & 37.8    & 56.6    & 52.1    & 86.7    & \cellcolor{gray!30}{\textbf{12.6}}  & 54.6  & 45.8 \\ \bottomrule
    \end{tabular}
    \end{adjustbox}
     \vspace{-0.5em}
    \end{table}

\section{Analysis of Adaptive Regularization} \label{appendix:adaptive_reg}
\begin{figure}[!t]
    \begin{subfigure}{0.5\textwidth}
        \centering
        \includegraphics[width=0.85\linewidth]{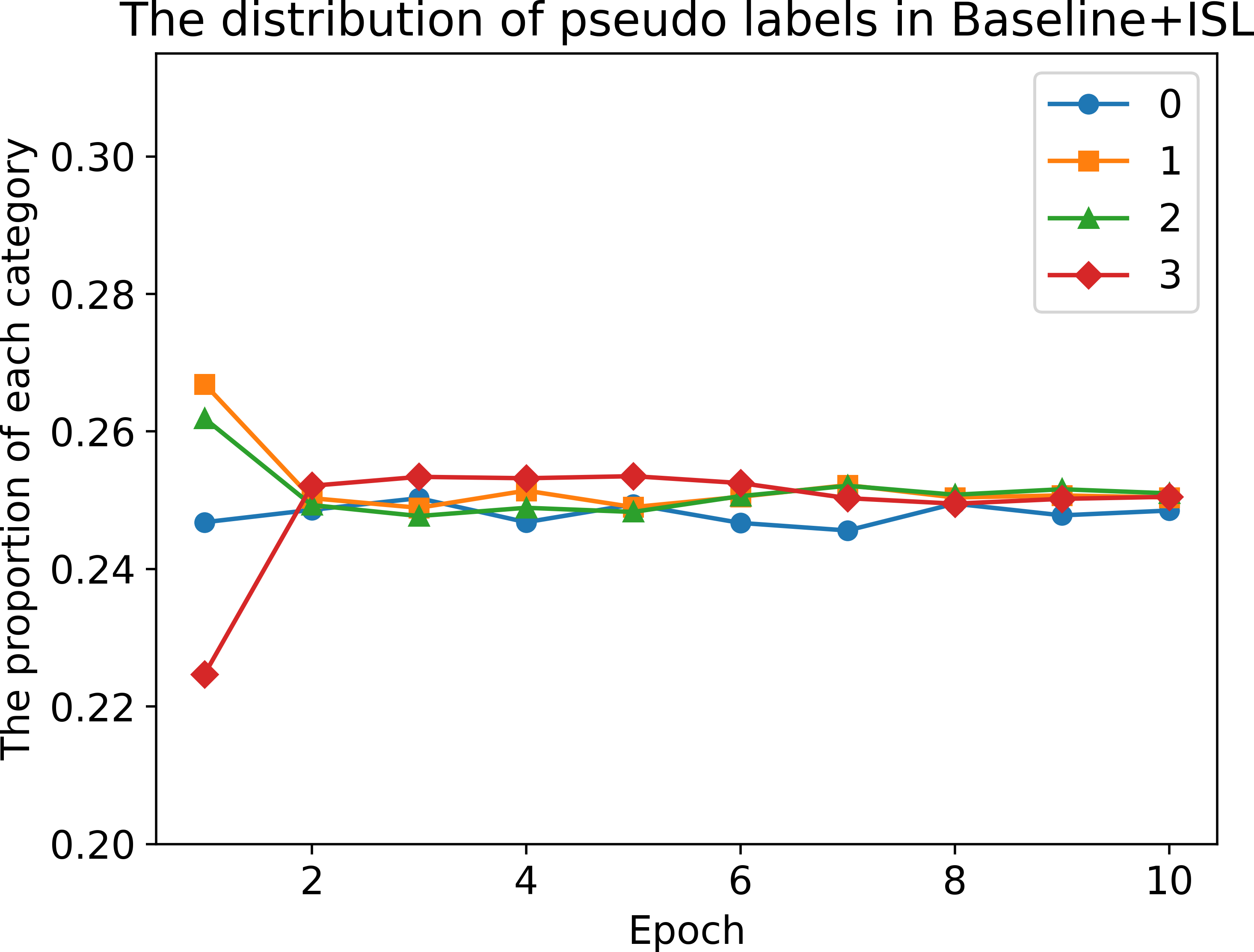}
    \end{subfigure}%
    \begin{subfigure}{0.5\textwidth}
        \centering
        \includegraphics[width=0.9\linewidth]{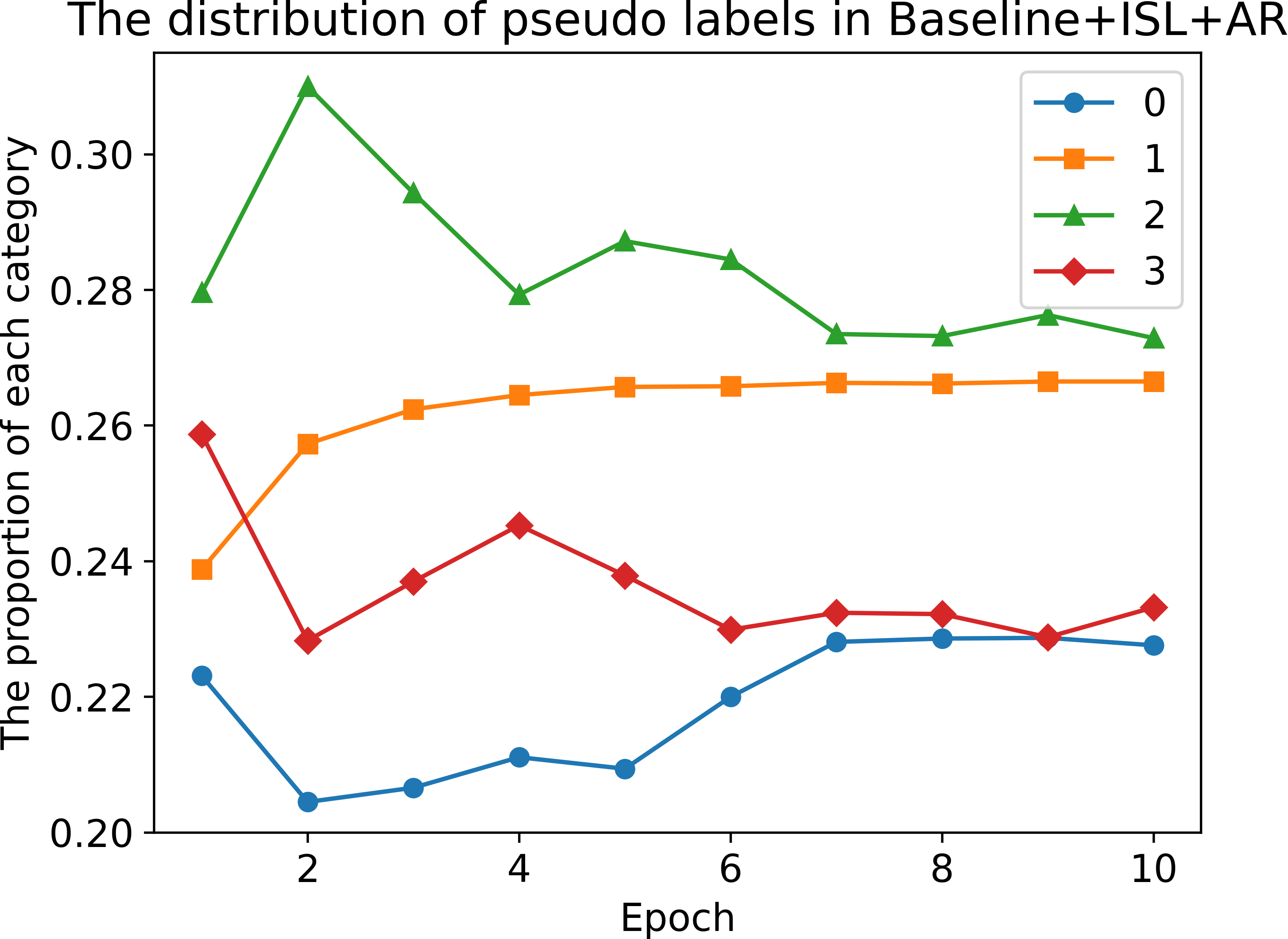}
    \end{subfigure}
    \caption{The pseudo label distribution before and after adding adaptive regularization}
    \label{fig:w}
\end{figure}

\begin{figure}[t]
    \centering
    \includegraphics[scale=0.07]{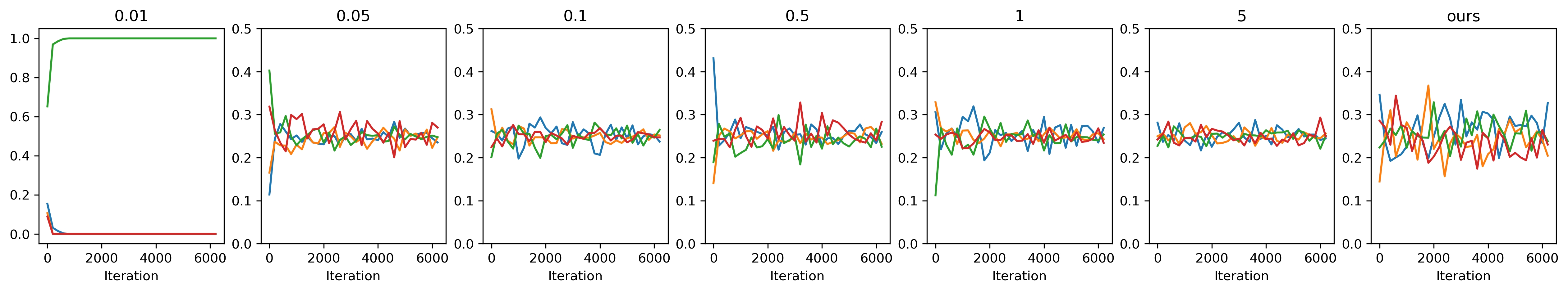}
    \caption{Class distribution for different fixed $\gamma$ during the training.}
    \label{fig:fix_w}
\end{figure}

We visualize the curve depicting changes in the pseudo-label distribution before and after adding adaptive regularization, as shown in \cref{fig:w}. Initially, the pseudo-label distribution in Baseline+ISL remains consistently uniform, with each of the four classes occupying 25\%. This indicates that the uniform constraint is too strong during later stages of training, leading to a pseudo-label distribution that tends towards uniformity, not reflecting the actual imbalanced point cloud data and resulting in suboptimal outcomes.After applying adaptive regularization, we dynamically adjust the uniform constraint based on the KL distance between the pseudo-label distribution and a uniform distribution. As depicted in the right plot of \cref{fig:w}, the curves representing changes in the pseudo-label distribution for each class do not converge towards uniformity. This demonstrates that our adaptive regularization, compared to a fixed $\gamma$, provides more flexibility in learning a pseudo-label distribution that better aligns with imbalanced point cloud data.

To illustrate how $\gamma$ affects the pseudo-label distribution, we present the class distribution of four classes for different fixed $\gamma$ values during training. As shown in \cref{fig:fix_w}, we observe the following:
1. A small $\gamma$ leads to a degenerate solution.
2. Increasing $\gamma$ gradually pushes the distribution towards uniformity.
3. Our adaptive $\gamma$ approach maintains flexibility, resulting in an imbalanced class distribution.

\section{ More Analysis of Prototype}\label{appendix:pro}
    \subsection{ {Initialization and updating process}} \label{appendix:pro1}
    In the beginning, the representation and prototypes are randomly initialized, which is very noisy. However, there are three key factors that guarantee us to gradually improve the representation and prototype. The first one is the learning of seen classes, which improves the representation ability of our model, thus improving the representation of novel classes implicitly.  To prove that known classes can help the representation of novel classes, we cluster the representation of novel classes obtained from a known-class supervised pre-trained model and a randomly initialized model on SemanticPOSS split 0. 

    The results indicate that features extracted from a known-class pre-trained model exhibit better clustering performance compared to features extracted from a randomly initialized model. The former outperforms the latter by nearly 7\% in mIoU for novel classes, demonstrating that known classes can indeed enhance the representation of novel classes.
    The second one is the view-invariant training, which learns invariant representation for different transformations and promotes the representation directly. Some studies \cite{long2023pointclustering, zhang2021self} have advanced unsupervised representation learning for point clouds by incorporating transformation invariance. 
    The third one is the utilization of spatial prior, which enforces the point in the same region to be coherent, which may be validated by Fig.3 and 4, and unsupervised clustering \cite{long2023pointclustering, zhang2023growsp}. 

\begin{figure}[t]
        \centering
        \begin{subfigure}[b]{0.18\textwidth}
            \centering
            \includegraphics[scale=0.12]{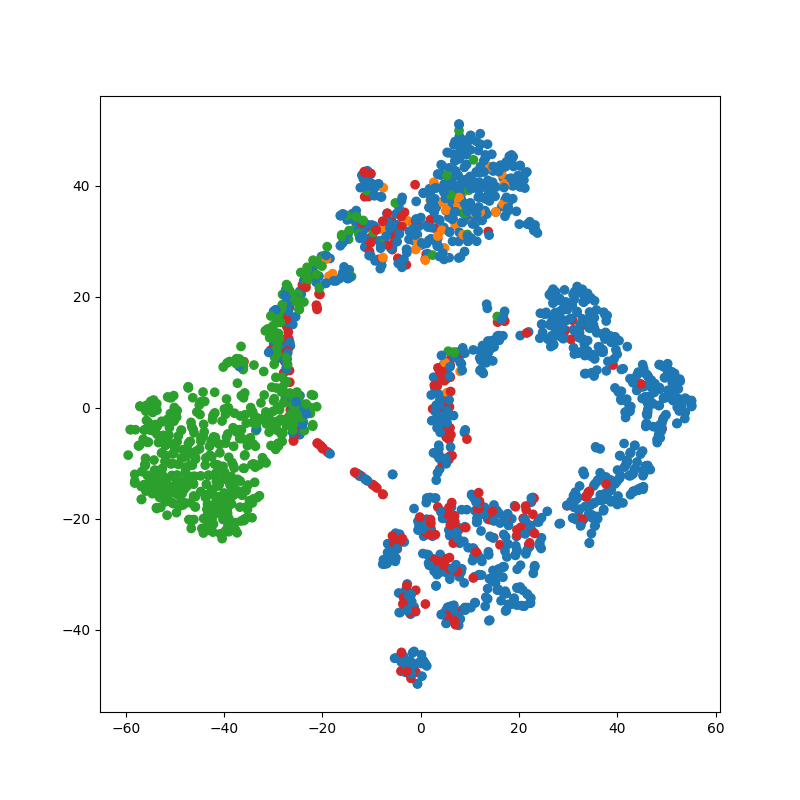}
            \caption{After 1 epoch}
            \label{fig:subfig1}
        \end{subfigure}
        \hfill
        \begin{subfigure}[b]{0.18\textwidth}
            \centering
            \includegraphics[scale=0.12]{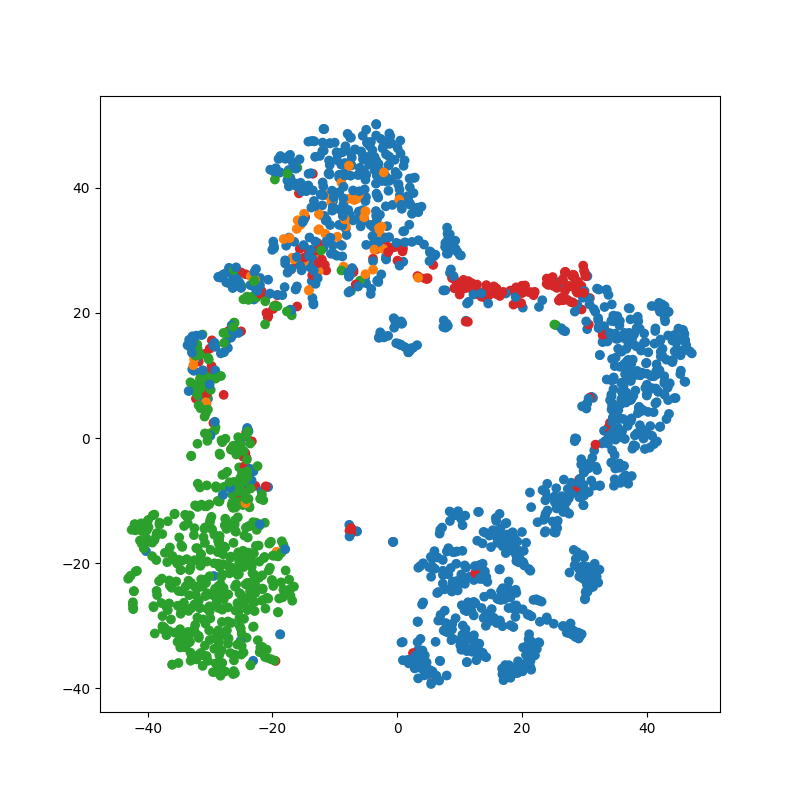}
            \caption{After 2 epochs}
            \label{fig:subfig2}
        \end{subfigure}
        \hfill
        \begin{subfigure}[b]{0.18\textwidth}
            \centering
            \includegraphics[scale=0.12]{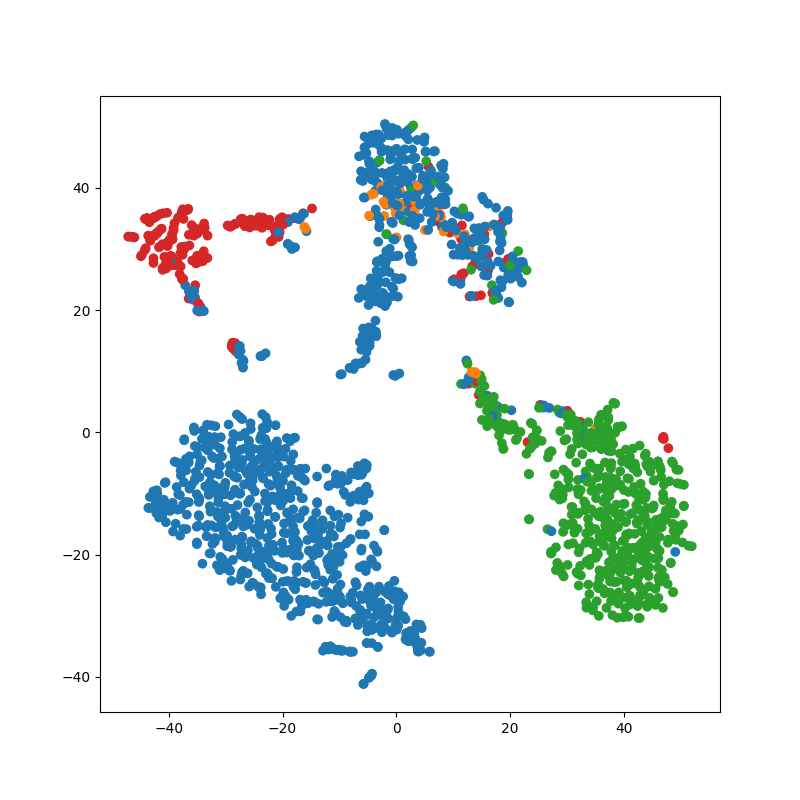}
            \caption{After 3 epochs}
            \label{fig:subfig3}
        \end{subfigure}
        \hfill
        \begin{subfigure}[b]{0.18\textwidth}
            \centering
            \includegraphics[scale=0.12]{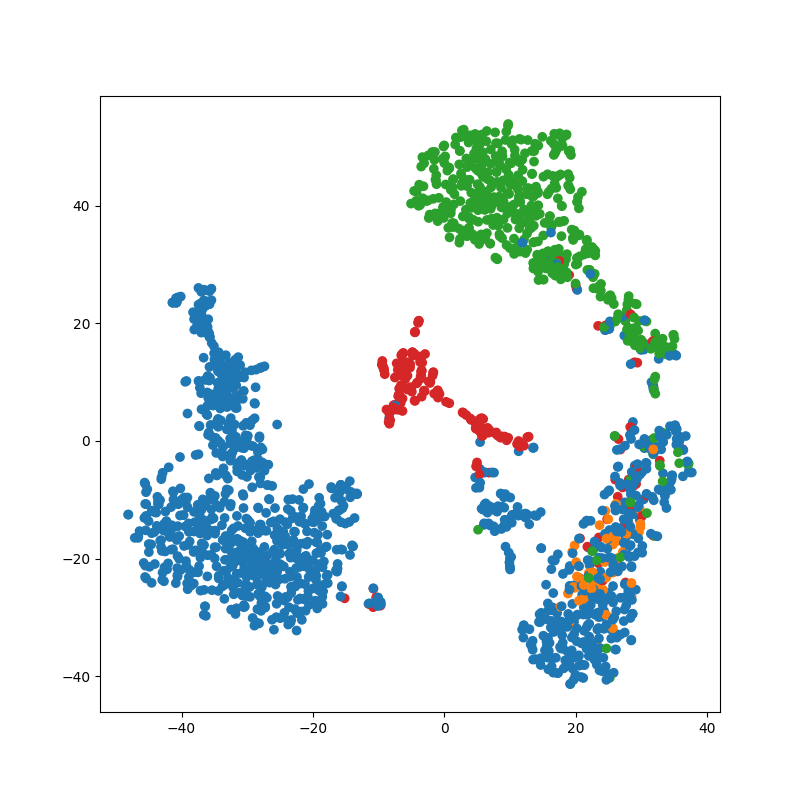}
            \caption{After 4 epoch}
            \label{fig:subfig4}
        \end{subfigure}
        \hfill
        \begin{subfigure}[b]{0.18\textwidth}
            \centering
            \includegraphics[scale=0.12]{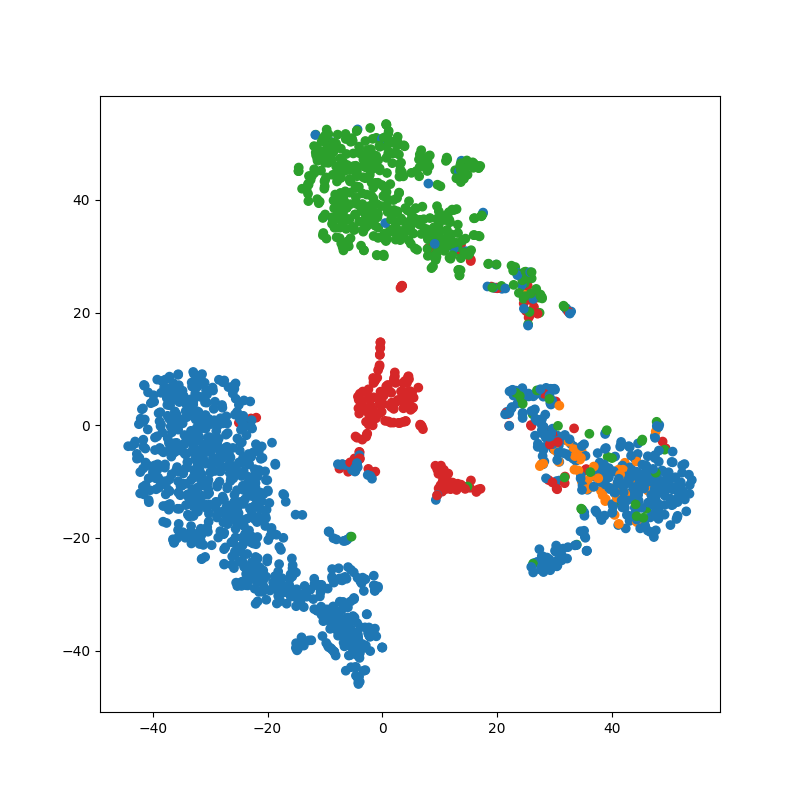}
            \caption{After 5 epochs}
            \label{fig:subfig5}
        \end{subfigure}
    
        \medskip % Add some space between the rows
    
        \begin{subfigure}[b]{0.18\textwidth}
            \centering
            \includegraphics[scale=0.12]{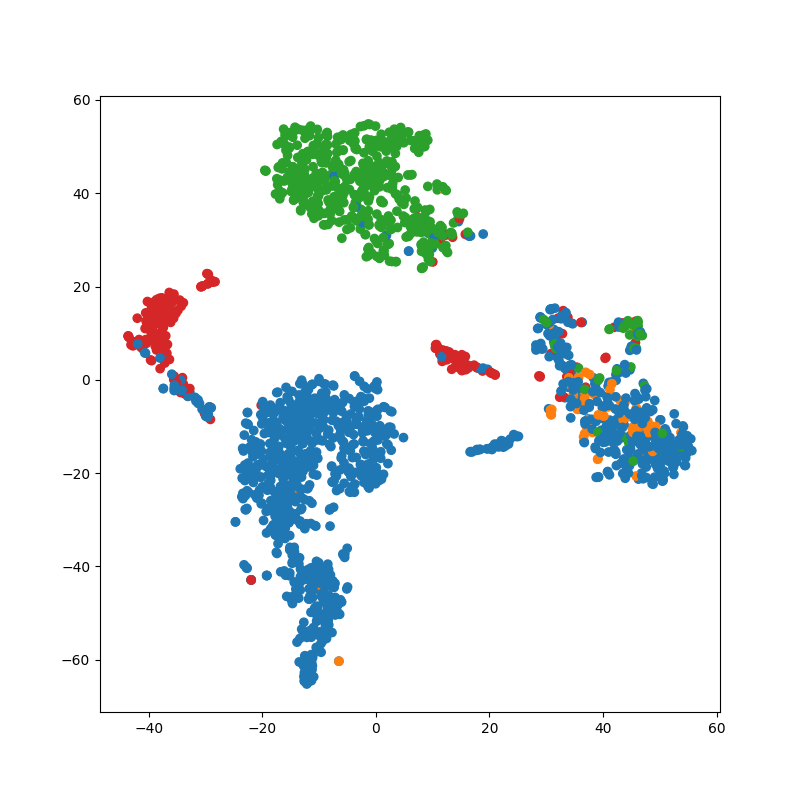}
            \caption{After 6 epochs}
            \label{fig:subfig6}
        \end{subfigure}
        \hfill
        \begin{subfigure}[b]{0.18\textwidth}
            \centering
            \includegraphics[scale=0.12]{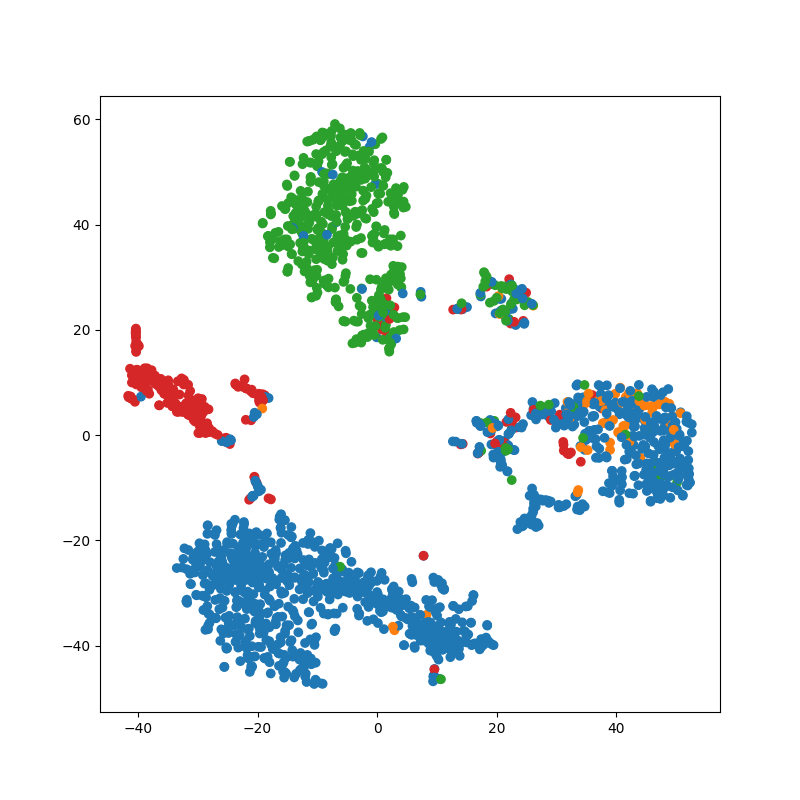}
            \caption{After 7 epoch}
            \label{fig:subfig7}
        \end{subfigure}
        \hfill
        \begin{subfigure}[b]{0.18\textwidth}
            \centering
            \includegraphics[scale=0.12]{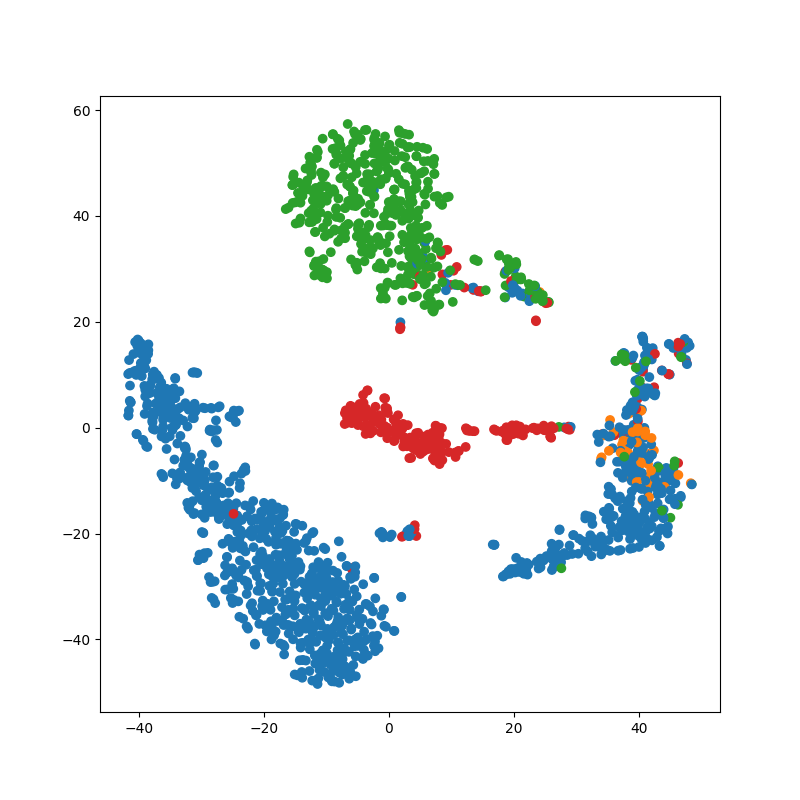}
            \caption{After 8 epochs}
            \label{fig:subfig8}
        \end{subfigure}
        \hfill
        \begin{subfigure}[b]{0.18\textwidth}
            \centering
            \includegraphics[scale=0.12]{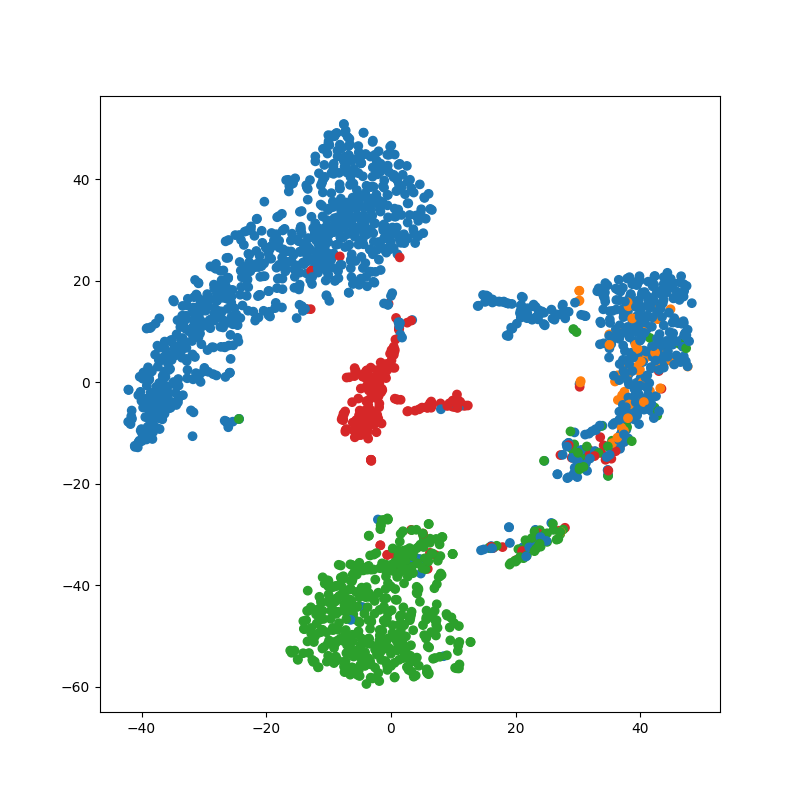}
            \caption{After 9 epochs}
            \label{fig:subfig9}
        \end{subfigure}
        \hfill
        \begin{subfigure}[b]{0.18\textwidth}
            \centering
            \includegraphics[scale=0.12]{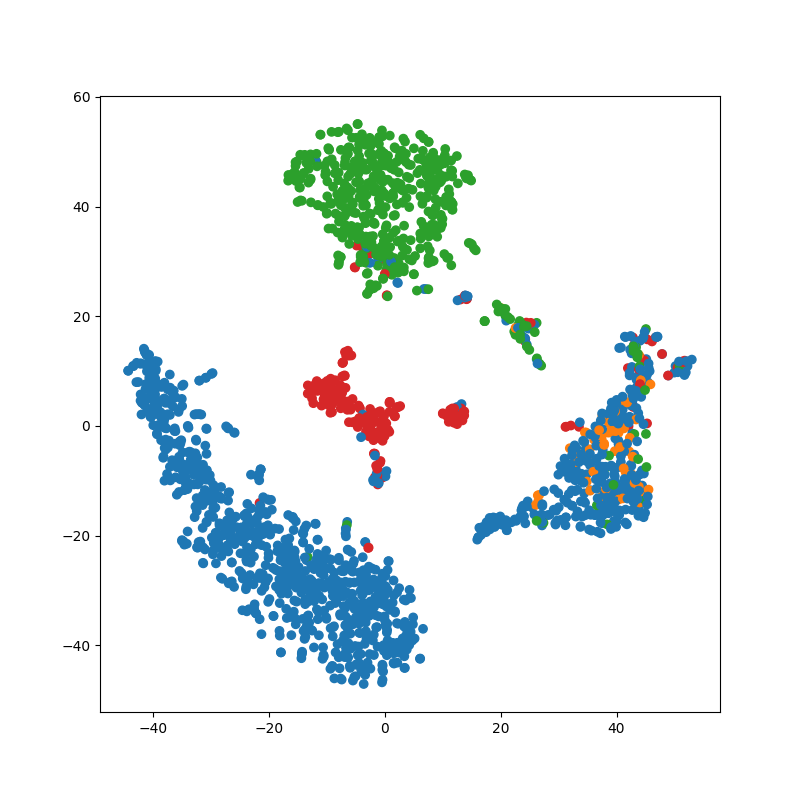}
            \caption{After 10 epoch}
            \label{fig:subfig10}
        \end{subfigure}

        \caption{{The quality of the representation of unseen classes during training in SemanticPOSS split 0. Blue dots are plants, green dots are the ground, orange are cars, and red are buildings }}
        \label{fig:pre}
    \end{figure}

    Those factors gradually improve the representation and prototype, leading to an informative prediction $P$. Then, our adaptive self-labeling algorithm utilizes $P$ and several marginal distribution constraints to generate pseudo-label $Q$. Finally, the $Q$ guides the learning of representation and prototype. In conclusion,  the above three factors and our self-labeling learning process ensure our method learns meaningful representation and prototypes gradually. Furthermore, we visualize the representation of novel classes during training in \cref*{fig:pre}, showing that as the training time increases, the learned representation gradually becomes better, validating our analysis.
    
\begin{table}[!t]
\footnotesize
\centering
\caption{Ablation alternate design of region-level prototype on split 0 of SemanticPOSS dataset. The results are on novel classes.}
 
\label{tab:prototype}
\begin{tabular}{c|cccc|c}
\toprule
Prototype Sharing & Building & Car  & Ground & Plants & Avg  \\ \midrule
$\times$  & 25.4     & 9.5 & 81.6   & 31.0   & 36.9 \\
 $\checkmark$       & 51.5  & 6.0 & 83.0   & 53.1   & 48.4 \\ \bottomrule
\end{tabular}
 
\end{table}
\subsection{Prototype sharing of region-level learning}
% 说明region-level的design为什么要share prototype。
%prototype；（生成super的大小，离群点的角度（appendix）\paragraph{Parameter analysis}%
% \begin{figure}
%     \centering
%     \includegraphics[scale=0.43]{imgs/pro.png}
%     \caption{Caption}
%     \label{fig:prototype}
% \end{figure}

We conduct experiments without sharing prototypes, and the results are depicted in \cref{tab:prototype}. It is noteworthy that utilizing two isolated prototypes results in a significant drop of nearly 10\% in performance in the novel class. 

%The analysis of similarity between point and region prototypes in Appendix H %\ref{appendix:pro} further illustrates that having two isolated prototypes leads to disparities in point-wise and region-wise learning directions. Conversely, sharing prototypes promotes the model to learn more compact and coherent representations.

% \subsection{More analysis of prototype sharing of region-level learning} \label{appendix:pro}

\begin{figure}[t]
     \centering
     \includegraphics[scale=0.43]{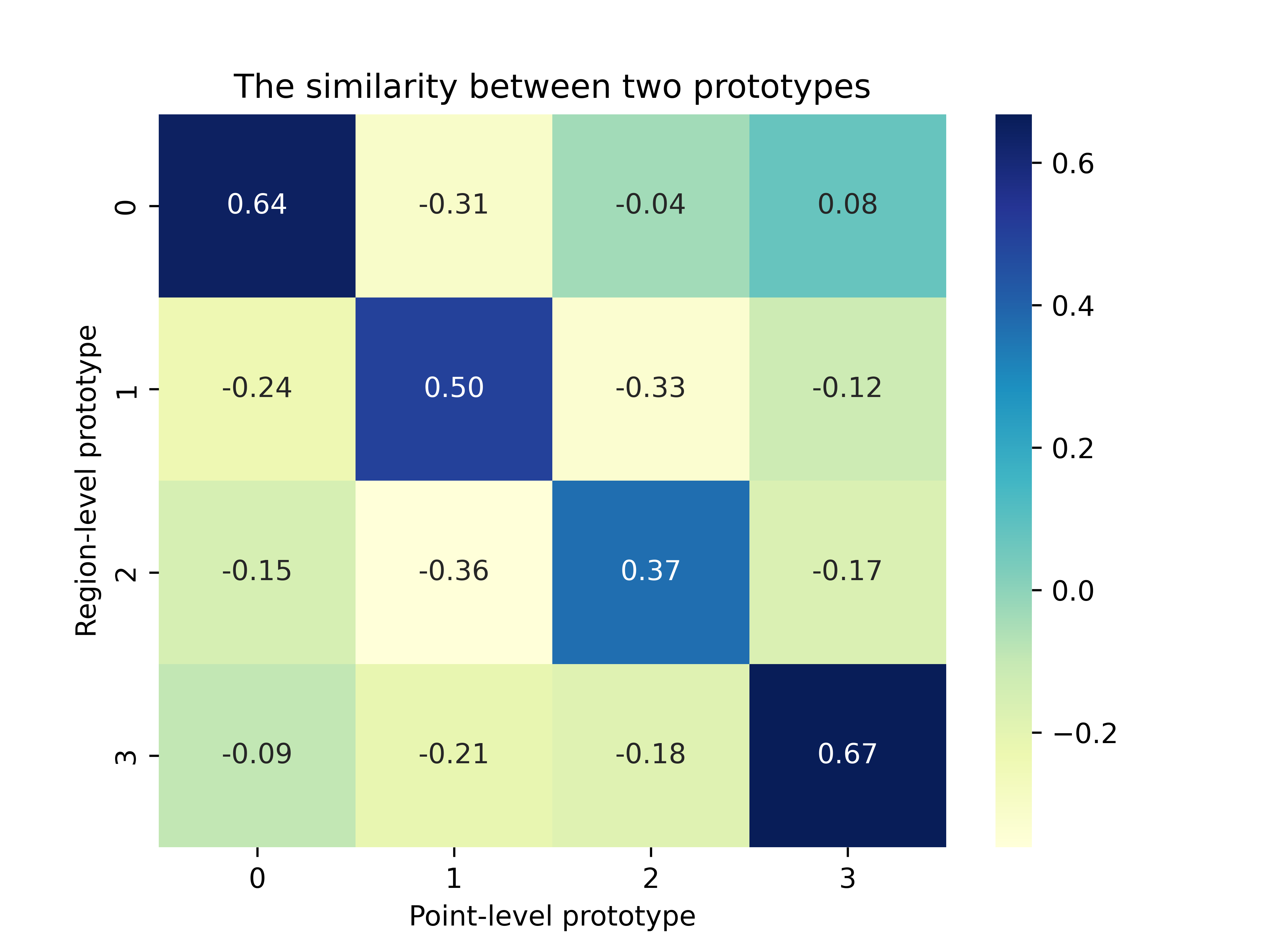}
     \caption{The similarity of two prototypes}
     \label{fig:prototype}
 \end{figure}

In addition, when prototypes are not shared, we visualize the similarity matrix between the two prototypes. As shown in \cref{fig:prototype}, we observe that the similarity between the two prototypes is low, which validates that having two prototypes leads to disparities in point-wise and region-wise learning directions. In contrast, sharing a prototype avoids this issue.

 \section{{More details on augmentation}} \label{appendix:augmentation}
Using augmentation to create two views is a well-established technique for learning a transformation-invariant representation, widely employed in novel class discovery literature \cite{fini2021unified,han2021autonovel}, and more recently applied in point clouds \cite{riz2023novel,long2023pointclustering,zhang2023growsp}. In our comparison, the previous sota method\cite{riz2023novel} also adopts the same augmentation as ours to generate two views. Therefore, our comparison ensures a fair assessment.
%Employing augmentation to generate two views is a well-known technique to learn transformation invariant representation and has been widely used in novel class discovery literature and point clouds. In our method, we first generate two views of the same point cloud by different augmentation, including naive rotation and scale, which are inherent from previous work [1] and have been illustrated in Implementation Details. Therefore, our comparison with existing work is fair.  Then, we generate point-level and region-level imbalanced pseudo-labels through adaptive self-labeling on two views in parallel. Subsequently, we swap the pseudo-labels generated from the two views to learn transformation invariant representation.

To show the effect of augmentation, we conduct ablation on the two views and augmentation.
\begin{table}[!t]
    \centering
    \caption{{More ablation experiments on SemanticPOSS}}
\label{tab:aug}
\begin{tabular}{@{}ccc|ccccc@{}}
    \toprule
    Aug & Two views & ISL+AR+Region & Building & Car & Ground & Plants & mIoU \\ \midrule
                            &                               & $\checkmark$                       & 22.1     & 2.1 & 34.4   & 24.8   & 20.9 \\
    $\checkmark$            &                               & $\checkmark$                       & 46.3     & 1.9 & 34.5   & 18.6   & 25.3 \\
    $\checkmark$            & $\checkmark$                  &                                    & 21.6     & 2.7 & 76.6   & 26.1   & 31.8 \\
    $\checkmark$            & $\checkmark$                  & $\checkmark$                       & 51.5     & 6.0 & 83.0   & 53.1   & 48.4 \\ \bottomrule
    \end{tabular}
\end{table}

As \cref{tab:aug}, From the results above, we draw the following conclusion: 1) Compare columns 1 and 2, augmentation on one view has improved by 4.4\% in novel classes compared with no augmentation; 2) Compare columns 2 and 4, employing two views has improved by 23.1\% in novel classes; 3) Compare column 3 and 4, our proposed techniques result in a significant improvement of 17\%.
%augmentation on one view has improved by 5\% in novel classes compared with no augmentation, and learning transformation invariant representation through two views has improved by 23\% in novel classes. However, the simple two views framework only reaches a lower baseline, such as the third row of the table. On this basis, adding the method we proposed will result in a significant improvement of 17\%.

\section{{More ablation study}} \label{appendix:aba}

We conduct additional ablation on split 0 of SemanticKITTI. The results are shown in \cref{tab:moreablationkitti}:
Similar to SemanticPOSS, each component enhances performance. Specifically, adaptive regularization and region-level learning individually contribute to a 6.6\% and 5.0\% improvement in mIoU for the model.

 \begin{table}[t]
    \centering
    \caption{{More ablation experiments on SemanticKITTI split 0}}
\label{tab:moreablationkitti}
\begin{tabular}{@{}ccc|cccccc@{}}
    \toprule
    ISL          & AR           & Region       & Building & Road & Sidewalk & Terrain & Vegetation & Avg  \\ \midrule
                 &              &              & 46.7     & 43.2 & 17.9     & 21.7    & 23.9       & 30.7 \\
    $\checkmark$ &              &              & 57.4     & 32.1 & 25.2     & 18.9    & 37.2       & 34.1 \\
    $\checkmark$ & $\checkmark$ &              & 70.8     & 34.7 & 23.2     & 16.8    & 57.9       & 40.7 \\
    $\checkmark$ & $\checkmark$ & $\checkmark$ & 74.6     & 41.4 & 22.5     & 23.6    & 66.4       & 45.7 \\ \bottomrule
    \end{tabular}
\end{table}

\section{{More details on DBSCAN}} \label{appendix:dbscan}
\subsection{Parameters for the DBSCAN algorithm}
\vspace{-1mm}
DBSCAN is a density-based clustering algorithm: given a set of points in some space, it groups points close to each other, marking as outliers points that lie alone in low-density regions. DBSCAN has two key parameters: epsilon and min-
samples. epsilon represents the maximum distance between two samples for one to be considered as in the neighborhood of the other, while min-samples denote the minimal number of samples in a region. In our experiments, we set min-samples to be reasonable minimal 2, indicating that there must be at least two points in a region. For epsilon, we determine a value of 0.5 based on the proportion of outliers, ensuring that 95\% of the point clouds participate in region branch learning. In the following part, we conduct experiments with different epsilon values and analyze the results.

\subsection{Visual examples of the resultant regions}
\begin{figure}[H]
        \centering
        \includegraphics[scale=0.53]{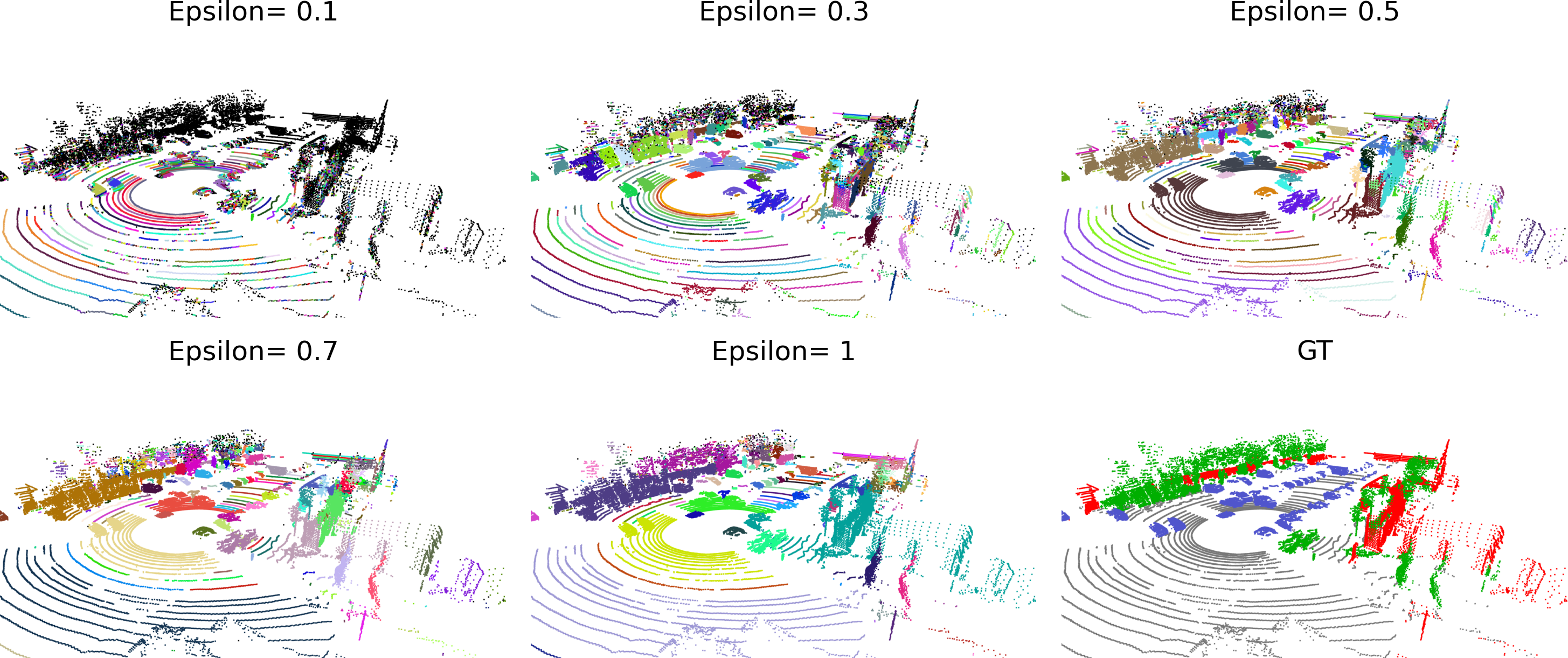}
        \caption{{Visualization of regions under different epsilon. For the first five pictures, the black point clouds are outliers, which means that the point does not belong to any region. Other random colors represent a region.}}
       \label{fig:dbscan}
\end{figure}

\vspace{-0cm}
We present visualizations of regions under different epsilon values in \cref{fig:dbscan}. As shown in the visualizations, a smaller epsilon results in more outliers and smaller generated regions. Conversely, a higher epsilon leads to fewer outliers and larger generated regions.
\vspace{-0mm}

\subsection{Sensitivity analysis of DBSCAN parameters}
\vspace{-0cm}
As shown in the \cref{tab:epsilon}, we supplement the proportion of outlier points in the 7th column and model training results under different epsilon in the 6th column. To assess the quality of regions, we assign a category label to each region based on the category with the highest point count within the region, with outliers being disregarded, and then calculate the mIoU in the 8th column. The results indicate that selecting 0.5 based on the outlier ratio yields satisfactory outcomes. Moreover, fine-tuning epsilon, for instance, setting it to 0.7, leads to improved performance. It is worth noting that the results first increased and then decreased with the increase of epsilon. This is because when epsilon is low, as shown in the visualization, there are more outliers, the generated region is smaller, and less spatial context information is used. When epsilon is higher, the generated region is larger and the Regions mIoU is lower, resulting in noisy region-level representation.

\begin{table}[t]
        \centering
        \caption{{Model training results, proportion of outlier points, and region mIoU under different epsilon. 
    Region mIoU is the mIoU between regions label and ground true. The region label is composed of 
    each region label, which is the category with the most points in each region. Region mIoU ignores outliers.}}
    \label{tab:epsilon}
        \begin{tabular}{@{}lccccccc@{}}
            \toprule
            epsilon                  & Building & Car  & Plants & Ground & mIoU  & Outlier & Region mIoU \\ \midrule
            \multicolumn{1}{l|}{0.1} & 41.5    & 1.7 & 45.6  & 80.7  & 42.4 & 45.6\%  & 97.0        \\
            \multicolumn{1}{l|}{0.3} & 49.2    & 8.3 & 49.2  & 83.8  & 47.6 & 7.5\%   & 84.5        \\
            \multicolumn{1}{l|}{0.5} & 51.5    & 6.0 & 53.1  & 83.0  & 48.4 & 2.5\%   & 74.8        \\
            \multicolumn{1}{l|}{0.7} & 65.5    & 9.0 & 61.3  & 78.2  & 53.5 & 1.3\%   & 64.5        \\
            \multicolumn{1}{l|}{1}   & 49.2    & 8.9 & 55.3  & 82.9  & 49.1 & 0.5\%   & 44.3        \\ \bottomrule
            \end{tabular}
\end{table}

 \section{More Visualization}\label{suppl:more_vis}

 To demonstrate the effectiveness of our method, we created a video to compare NOPS with our prediction results, and our method shows a significant improvement over NOPS.

 \begin{figure}[t]
    \centering
    \includegraphics[scale=0.06]{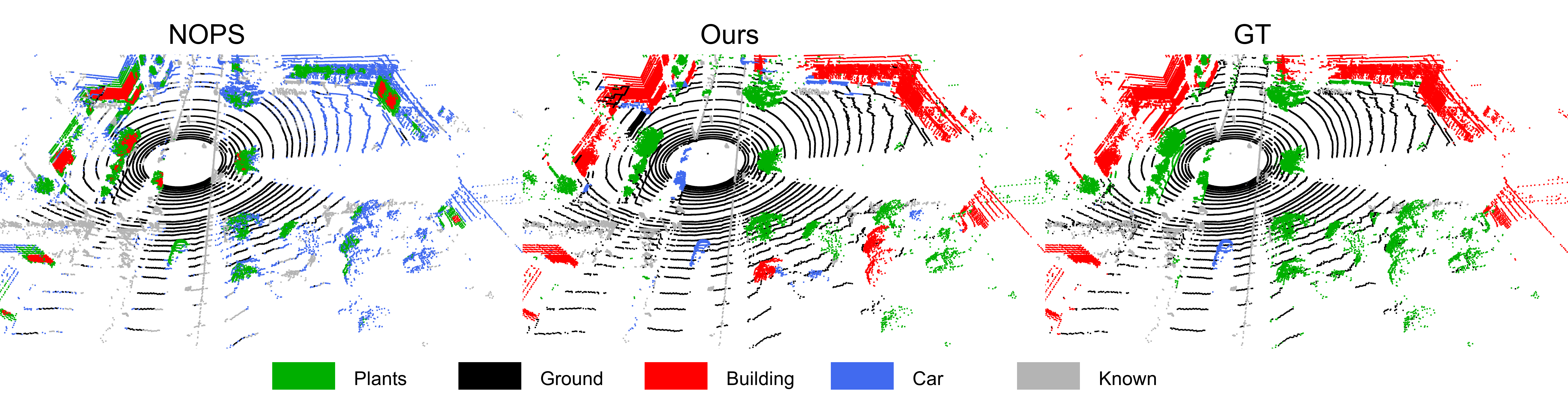}
    \caption{One frame from our video, the dataset being SemanticPOSS split 0.}
    \label{fig:video}
\end{figure}